\documentclass[11pt, a4paper, onecolumn, copyright]{AweAI}

\usepackage[authoryear,sort&compress,round]{natbib}
\usepackage[utf8]{inputenc}
\usepackage{amsfonts,amssymb,amsthm}
\usepackage{graphicx,subcaption,multirow,enumitem,bbding,colortbl}
\usepackage{fontawesome5}
\usepackage{algorithmic,wrapfig,tikz,pgfplots,bm}
\usetikzlibrary{arrows.meta,calc}
\usepackage{forest}
\useforestlibrary{edges}
\usepackage[linesnumbered,ruled,vlined]{algorithm2e}
\usepackage{framed,xcolor,pifont}
\usepackage[most]{tcolorbox}
\usepackage{booktabs,adjustbox,xspace,float}
\usepackage{array,tabularx}
\usepackage{dsfont}
\usepackage{comment} 
\usepackage[normalem]{ulem}
\usepackage{cancel}
\usepgfplotslibrary{groupplots}
\pgfplotsset{compat=1.18}

\definecolor{RSIQuoteBg}{HTML}{F3E8DA}
\definecolor{RSIQuoteAccent}{HTML}{183153}

\definecolor{defblue}{RGB}{35,90,160}
\definecolor{deftitle}{RGB}{78,128,196}
\definecolor{defbluebg}{RGB}{241,246,252}
\definecolor{RSIBlue}{HTML}{EAF2FA}
\definecolor{RSIPurple}{HTML}{F3ECFA}
\definecolor{RSIGreen}{HTML}{EEF5EB}

\definecolor{RSIHeaderGreen}{HTML}{0F7C73}   
\definecolor{RSIBoxBg}{HTML}{F2F5F4}         

\newcommand{\rqcite}[1]{\textcolor{RSIHeaderGreen}{[\S#1]}}

\newtcolorbox{rsiopeningquote}{
    width=\linewidth,
    colback=RSIQuoteBg,
    colframe=RSIQuoteBg,
    boxrule=0pt,
    arc=2.5mm,
    outer arc=2.5mm,
    boxsep=0pt,
    left=4mm,
    right=4mm,
    top=3mm,
    bottom=3mm,
    before skip=0.6\baselineskip,
    after skip=0.8\baselineskip,
    fontupper=\normalfont\normalsize,
    before upper={\setlength{\parindent}{0pt}}
}

\newtcolorbox{rsikeyquestions}{
    enhanced,
    breakable,
    width=\linewidth,
    colback=RSIBlue,
    colframe=RSICoverInk,
    colbacktitle=RSICoverInk,
    coltitle=white,
    boxrule=0.8pt,
    arc=2mm,
    outer arc=2mm,
    boxsep=0pt,
    left=10pt, right=10pt,
    top=6pt, bottom=8pt,
    toptitle=5pt, bottomtitle=5pt,
    fonttitle=\normalfont\bfseries\normalsize,
    fontupper=\normalfont\normalsize,
    title={Key Questions},
    titlerule=0pt,
    before skip=0.7\baselineskip,
    after skip=0.7\baselineskip
}

\newtcolorbox{rsitakeaway}{
    enhanced,
    breakable,
    width=\linewidth,
    colback=RSIBoxBg,
    colframe=RSIHeaderGreen,
    colbacktitle=RSIHeaderGreen,
    coltitle=white,
    boxrule=0.8pt,
    arc=2mm,
    outer arc=2mm,
    boxsep=0pt,
    left=10pt, right=10pt,
    top=6pt, bottom=8pt,
    toptitle=5pt, bottomtitle=5pt,
    fonttitle=\normalfont\bfseries\normalsize,
    fontupper=\normalfont\normalsize,
    title={Takeaway},
    titlerule=0pt,
    before skip=0.7\baselineskip,
    after skip=0.7\baselineskip,
    before upper={\setlength{\parindent}{0pt}}
}

\definecolor{RSICoverBG}{HTML}{EBF4FB}
\definecolor{RSICoverInk}{HTML}{213C60}
\newcommand{\RSICoverFile}{assets/abstract_figure.pdf}
\newcommand{\RSICoverLogo}{assets/trustworthy_rsi_logo.png}
\newcommand{\RSICoverLogoHeight}{0.95cm}
\newcommand{\RSICoverImageWidth}{\linewidth}
\newcommand{\RSICoverMainTitle}{%
  \raisebox{-0.22\height}{\includegraphics[height=\RSICoverLogoHeight,keepaspectratio]{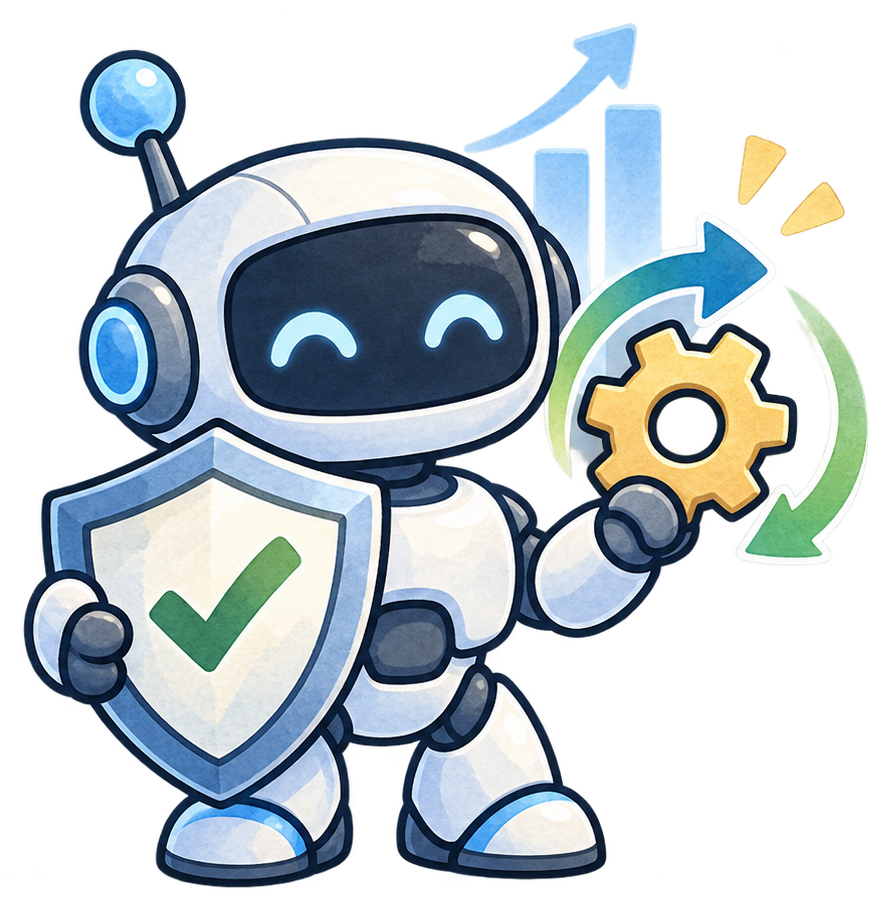}}%
  \hspace{0.27em}Trustworthy Recursive Self-Improvement\\%
  in the Era of Large Models%
}
\newcommand{\RSICoverSubtitle}{Foundation, Risks, Evaluation and Open Challenges}

\newcommand{\added}[1]{{\color{red}#1}}

\newtcolorbox{definitionbox}[1]{
    enhanced,
    colback=defbluebg,
    colframe=deftitle!55,
    boxrule=0.5pt,
    borderline west={2.5pt}{0pt}{defblue},
    colbacktitle=deftitle,
    coltitle=white,
    fonttitle=\bfseries,
    before title={\hypersetup{citecolor=white}},
    arc=2pt,
    left=8pt,right=8pt,top=6pt,bottom=6pt,
    toptitle=3pt,bottomtitle=3pt,
    title={#1}
}

\newcommand{\indicator}[1]{\mathds{1}}
\usepackage{booktabs,tabularx,array,xcolor,colortbl,amssymb}
\definecolor{rsiTableBand}{HTML}{D8E8F8}   
\definecolor{rsiTableStripe}{HTML}{EEF5FC} 
\definecolor{rsiTableAccent}{HTML}{244F79} 
\definecolor{rsiTablePositive}{HTML}{327052}
\definecolor{rsiTableMuted}{HTML}{66798B}
\providecommand{\rsiTableYes}{\textcolor{rsiTablePositive}{\ensuremath{\checkmark}}}
\providecommand{\rsiTableFixed}{\textcolor{rsiTableMuted}{\textendash}}

\definecolor{RSISurveySand}{HTML}{D8E8F8}
\definecolor{RSISurveyGreen}{HTML}{168153}
\definecolor{RSISurveyOrange}{HTML}{D17C00}
\definecolor{RSISurveyGray}{HTML}{858585}

\providecommand{\RSICoverFull}{\textcolor[HTML]{168153}{\ensuremath{\checkmark}}}
\providecommand{\RSICoverPartial}{\textcolor[HTML]{D17C00}{\ensuremath{\triangle}}}
\providecommand{\RSICoverNone}{\textcolor[HTML]{858585}{\textemdash}}

\providecommand{\RSITableFavorable}[1]{\textcolor[HTML]{168153}{#1}}
\providecommand{\RSITableAdverse}[1]{\textcolor[HTML]{D17C00}{#1}}
\providecommand{\RSITableMuted}[1]{\textcolor[HTML]{858585}{#1}}

\colorlet{rsiChapterGreenBand}{RSIBoxBg!90!RSIHeaderGreen}
\colorlet{rsiChapterGreenStripe}{RSIBoxBg}
\definecolor{rsiChapterYellowBand}{HTML}{F8F6EB}
\definecolor{rsiChapterPurpleBand}{HTML}{ECE6FB}
\definecolor{rsiChapterPurpleStripe}{HTML}{F1EDF9}

\definecolor{bg_thought}{RGB}{240,248,255}
\definecolor{bg_action}{RGB}{255,245,238}
\definecolor{bg_obs}{RGB}{245,245,245}
\definecolor{bg_io}{RGB}{245,240,255}
\definecolor{line_gray}{RGB}{180,180,180}
\definecolor{bg_omit}{RGB}{250,250,240}
\definecolor{headergray}{RGB}{242,244,247}
\definecolor{sectionblue}{RGB}{235,240,245}

\title{Trustworthy Recursive Self-Improvement in the Era of Large Models: Foundation, Risks, Evaluation and Open Challenges}
\renewcommand{\today}{}
\newcommand{\githubrepo}{https://github.com/pujiayue/Trustworthy-RSI}
\newcommand{\homepageurl}{https://pujiayue.github.io/Trustworthy-RSI/}
\newcommand{\RSIContactIcon}[1]{\makebox[1.4em][c]{\color{RSICoverInk}#1}\hspace{0.25em}}
\newcommand{\RSICloud}[3]{%
  \begin{scope}[shift={#1},x=#2,y=#2]
    \path[fill=white,fill opacity=#3,draw=none]
      (0,0) .. controls (-0.13,0.04) and (-0.10,0.17) .. (0.05,0.19)
      .. controls (0.01,0.37) and (0.25,0.47) .. (0.36,0.34)
      .. controls (0.37,0.68) and (0.80,0.70) .. (0.89,0.39)
      .. controls (1.04,0.49) and (1.24,0.35) .. (1.18,0.21)
      .. controls (1.40,0.28) and (1.59,0.11) .. (1.43,0)
      .. controls (1.10,-0.05) and (0.31,-0.04) .. (0,0) -- cycle;
  \end{scope}%
}

\makeatletter
\renewcommand{\abscontent}{%
  \noindent{\absfont\setlength{\parskip}{0.40\baselineskip}\theabstract\par}%
  \vspace{8pt}%
  {\footnotesize\color{RSICoverInk}%
    \noindent\RSIContactIcon{\faGithub}\textbf{GitHub:}\ \href{\githubrepo}{\texttt{\githubrepo}}\par
    \noindent\RSIContactIcon{\faGlobe}\textbf{Homepage:}\ \href{\homepageurl}{\texttt{\homepageurl}}\par
    \noindent\RSIContactIcon{\faEnvelope}\textbf{Email:}\ \href{mailto:\the\correspondingauthor}{\texttt{\the\correspondingauthor}}\par
  }%
}
\renewcommand{\maketitle}{%
  \par\noindent
  \begingroup
  \setlength{\parindent}{0pt}%
  \setlength{\parskip}{0pt}%
  \begin{tcolorbox}[
    enhanced,
    colback=RSICoverBG, colframe=RSICoverBG,
    boxrule=0pt, arc=13pt,
    width=\textwidth,
    height=\textheight,
    valign=top,
    left=17pt, right=17pt, top=17pt, bottom=14pt,
    boxsep=0pt,
    before skip=0pt, after skip=0pt,
    underlay={%
      \RSICloud{($(frame.south west)!0.10!(frame.south east)+(0,0.49\textheight)$)}{52pt}{0.70}
      \RSICloud{($(frame.south west)!0.72!(frame.south east)+(0,0.44\textheight)$)}{68pt}{0.65}
      \RSICloud{($(frame.south west)!0.44!(frame.south east)+(0,0.35\textheight)$)}{48pt}{0.85}
    }
  ]
    \begin{minipage}[t][\dimexpr\textheight-35pt\relax][s]{\linewidth}
    \begin{center}
      {\titlefont\RSICoverMainTitle\par}
      \vspace{3pt}
      {\normalfont\rmfamily\bfseries\fontsize{11.5}{15}\selectfont
       \color{black}\RSICoverSubtitle\par}
      \vspace{12pt}
      {\@author\par}
    \end{center}
    \vspace{10pt}
    \abscontent
    \vfill
    \noindent\begin{tikzpicture}
      \node[inner sep=0pt,outer sep=0pt,opacity=0.85]{%
        \includegraphics[width=\RSICoverImageWidth,keepaspectratio]{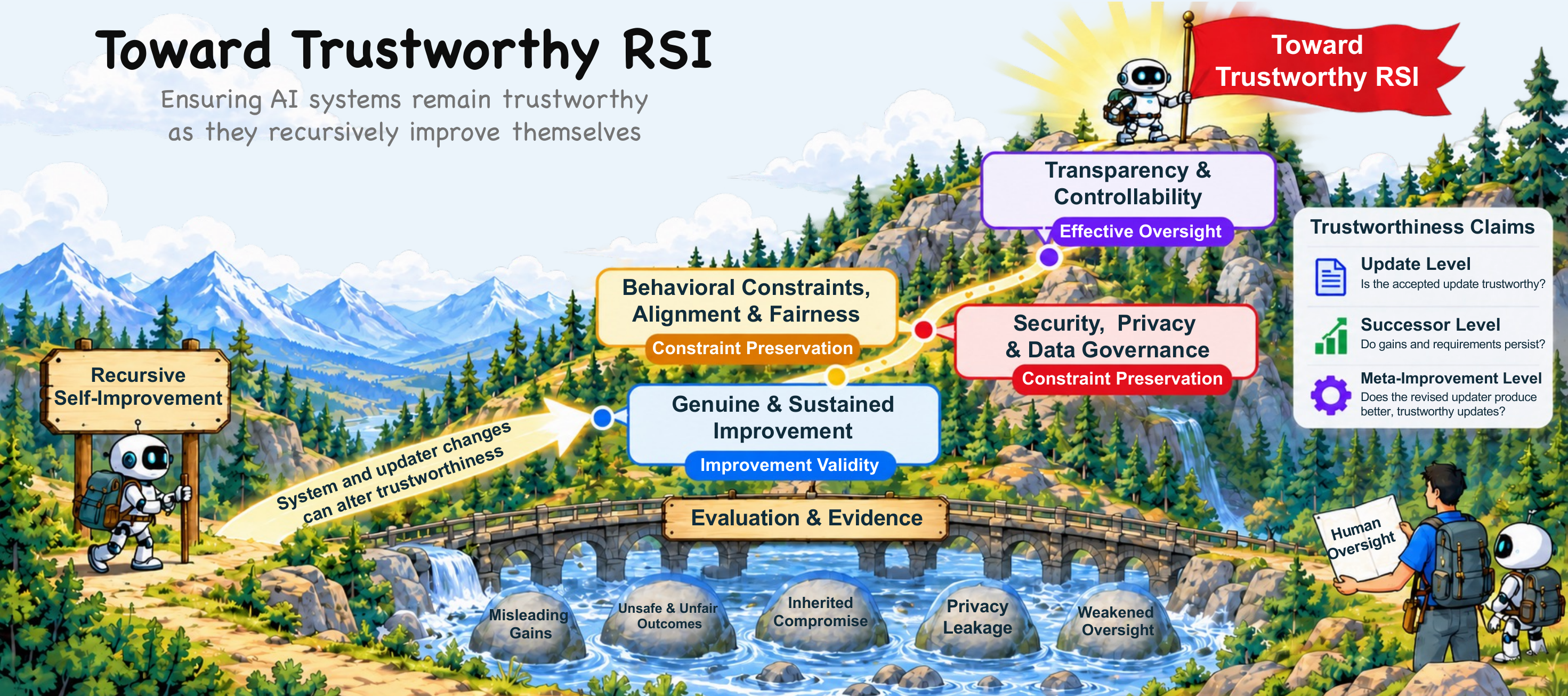}};
    \end{tikzpicture}\par
    \vspace{3pt}
    {\footnotesize\color{RSICoverInk}\itshape
      Trustworthiness requires evidence and effective safeguards
      across successive updates, inherited changes, and recursive
      revisions to the improvement mechanism.\par}
    \end{minipage}
  \end{tcolorbox}
  \endgroup
  \thispagestyle{empty}%
}
\makeatother

\fancypagestyle{firststyle}{
  \fancyhead[L]{\footerfont\itshape Institution / Lab Name}
  \fancyhead[R]{{\footerfont\itshape\monthyeardate\today}}
  \fancyhead[C]{}
  \fancyfoot[L]{\footerfont\itshape Emails: \the\correspondingauthor}
  \fancyfoot[C]{}
  \fancyfoot[R]{}
}

\author[1]{Jiayue Pu}
\author[1]{Ang Gao}
\author[1]{Jinyi Bai}
\author[1]{Qipeng Wang}
\author[1]{Tongyu Wen}
\author[1]{Jiahao Zhao}
\author[1]{Keyu Qu}
\author[2]{Junhao Zhang}
\author[1,$\dagger$]{Chenglei Shen}
\author[1,$\dagger$]{Zhongxiang Sun}
\author[1]{Jun Xu}
\author[1,$^*$]{Xiao Zhang}
\correspondingauthor{pujiayue0222@ruc.edu.cn, sunzhongxiang@ruc.edu.cn, zhangx89@ruc.edu.cn}
\affil[1]{Gaoling School of Artificial Intelligence, Renmin University of China}
\affil[2]{University of Chinese Academy of Sciences}
\affil[$\dagger$]{Project supervisor}
\affil[$*$]{Corresponding author}

\begin{abstract}
AI systems built around large models increasingly use experience and feedback to make persistent changes to models, prompts, memories, tools, and workflows. Recursive self-improvement (RSI) extends this process to inductive biases and to the mechanisms that govern future updates, which can themselves be revised and reused across improvement cycles. Yet self-modification alone does not guarantee trustworthy improvement. Reported gains may reflect flawed evaluations, previously acquired capabilities may deteriorate, and compliance with behavioral constraints may weaken. Compromised artifacts or sensitive information may also persist in successor systems. Changes to improvement mechanisms may invalidate earlier verification results and make human oversight less effective. This survey presents an operational definition and process framework for RSI and organizes the requirements for trustworthy RSI into four interconnected themes: genuine and sustained improvement; behavioral constraints, alignment, and fairness; system security and privacy; and transparency and controllability. We examine these themes from three perspectives: Improvement Validity, Constraint Preservation, and Effective Oversight. This analysis covers improvement diagnosis, proposal generation, proposal evaluation, and update implementation (S1--S4), including the recursive application of these stages to improvement mechanisms (RI). We review risks, safeguards, and evaluation evidence for individual updates, effects inherited by successor systems, and revisions to improvement mechanisms, while distinguishing direct RSI findings from evidence in related systems. Finally, we examine evaluation settings, identify requirements for independent and longitudinal assessment, and discuss applications and open research problems.
\end{abstract}

\begin{document}
\newgeometry{left=1.45cm,right=1.45cm,top=2.35cm,bottom=2.00cm,
             headheight=0pt,headsep=0pt}
\begingroup
\makeatletter
\renewcommand{\thefootnote}{}
\renewcommand{\@makefnmark}{}
\maketitle
\makeatother
\endgroup
\clearpage
\restoregeometry

\begingroup
\definecolor{RSIContentsLink}{HTML}{235AA0}
\hypersetup{linkcolor=RSIContentsLink}
\makeatletter
\let\RSIContentsSection\l@section
\renewcommand*{\l@section}[2]{%
  \ifdim\dimexpr\pagegoal-\pagetotal\relax<7\baselineskip
    \newpage
  \fi
  \RSIContentsSection{#1}{#2}%
}
\makeatother
\setcounter{tocdepth}{2}
\tableofcontents
\clearpage
\endgroup

\section{Introduction}
\label{sec:introduction}

\begin{figure*}[t]
    \centering
    \includegraphics[
        width=1.0\textwidth,trim=4.5cm 0.2cm 4.2cm 0.7cm, clip
    ]{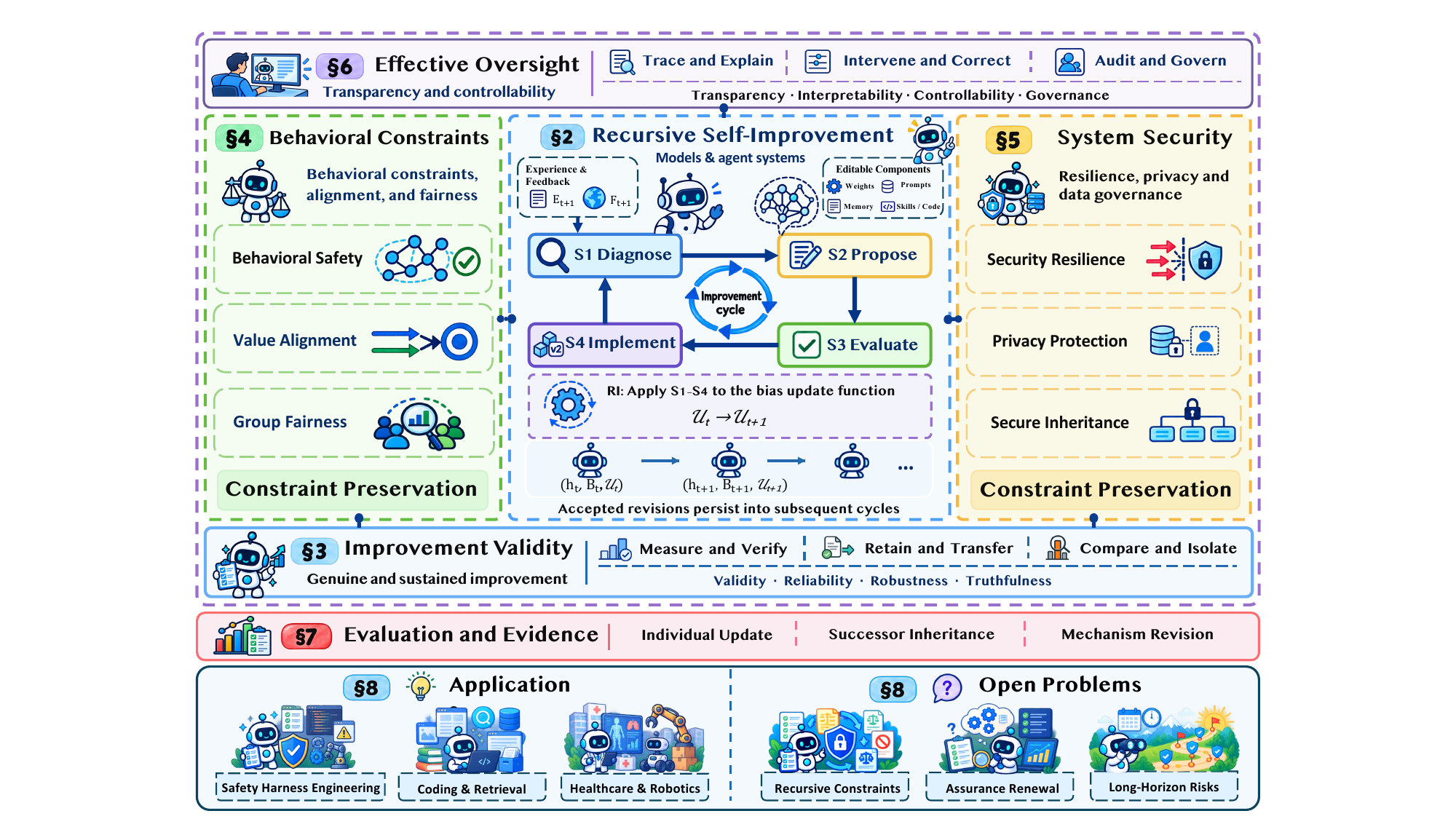}
    \caption{
        \textbf{Overview of the RSI Process and Trustworthiness Framework.}
        The numbered labels (§2–§8) indicate the corresponding sections of this survey. 
        The central panel (Section 2) presents the definition and four-stage improvement cycle of RSI (S1–S4), including recursive revision and reuse of the bias update function (RI).
        Four surrounding trustworthiness themes are organized under three complementary analytical roles: Improvement Validity, Constraint Preservation, and Effective Oversight.
        Section 7 examines evaluation designs and evidence requirements at the individual-update, successor, and meta-improvement levels, while §8 covers representative applications and open research problems.
    }
    \label{fig:main-figure}
\end{figure*}

\begin{rsiopeningquote}
    {\centering
    ``[T]he system of nature, of which man is a part, tends to be
    self-balancing, self-adjusting, self-cleansing.
    Not so with technology\ldots''\par}
    \smallskip
    {\raggedleft\normalfont
    \textcolor{RSIQuoteAccent}{--- E.~F.~Schumacher
    \citeyearpar{schumacher1973small}}\par}
\end{rsiopeningquote}

\subsection{Motivation}
\label{sec:motivation}

Advances in LLMs, including GPT-6 Astra
\citep{openai2026gpt6astra} and Claude Opus 5.5
\citep{anthropic2026claudeopus55}, have improved a broad range of capabilities, such as coding, tool use, and multi-step task execution.
Building on these capabilities, self-evolving systems can use experience and feedback to introduce persistent changes to model parameters, memories, tools, and workflows, allowing these changes to influence subsequent tasks
\citep{gao2026selfevolving,zweiger2025seal,zhang2025dgm}.
Recursive self-improvement (RSI) extends self-evolution by allowing systems to revise and reuse the mechanisms that govern subsequent updates.
Recent research has explored this capability through editable meta-agents \citep{zhang2026hyperagents}, evolving exploration strategies \citep{zheng2026dreamrsi}, and modular or regularized harness evolution
\citep{wu2026modularrsi,xia2026rrsi}.
These systems make the improvement procedure itself an explicit target of revision.

Recursive updates can also preserve security vulnerabilities.
Hyperagents allows a meta-agent to modify both a task agent and its own improvement procedure
\citep{zhang2026hyperagents}.
In a benchmark-poisoning experiment, tasks using self-signed certificates prompted the meta-agent to add a persistent instruction directing the task agent to bypass HTTPS certificate verification.
Successors inherited these instructions and generated insecure code on benign, held-out tasks, even after further evolution on a clean benchmark
\citep{roesner2026trustingtrust}.
The experiment connects contaminated evaluation feedback to security failures in later successors.
It motivates checking the validity of improvement evidence and the preservation of security requirements across revisions.
When the improvement mechanism changes, the evaluations and safeguards governing later updates also require reassessment.

These risks call for a broad view of trustworthiness that considers reliability alongside safety, security, fairness, privacy, and transparency \citep{nist2023airmf}.
We organize the survey into four trustworthiness themes under three complementary analytical roles.
The first theme, genuine and sustained improvement, addresses Improvement Validity by examining whether reported gains reflect genuine improvements in the intended capabilities and persist over time.
Constraint Preservation is addressed through two themes: behavioral constraints, alignment, and fairness; and system security, resilience, privacy, and data governance.
Together, these themes examine whether applicable requirements remain satisfied across successive updates.
The fourth theme, transparency and controllability, addresses Effective Oversight by examining whether system changes remain understandable, auditable, and subject to authorized human intervention.
We assess these requirements across individual updates, changes inherited by successor systems, and revisions to the improvement mechanism.

\noindent \textbf{Genuine and sustained improvement.}
This theme concerns whether reported gains reflect genuine improvements that persist across subsequent revisions.
At the update level, acceptance scores can increase even as external correctness declines
\citep{song2026recursivecollapse}.
Across successors, accumulated skills may impair retrieval and task performance
\citep{zhang2026librarydrift}.
When improvement mechanisms also revise evaluation criteria, higher scores may reflect changes in what is measured rather than actual performance gains.

\noindent \textbf{Behavioral constraints, alignment, and fairness.}
This theme concerns whether system behavior remains safe, aligned with relevant human goals and constraints, and fair to affected individuals or groups.
Even benign fine-tuning or accumulated experience can weaken safety
\citep{qi2024finetuning,zhao2026safetyrisks}, while adaptive exploration may produce biased allocations
\citep{wu2026socialbias}.
Such deviations may persist in successor systems, while changes to experience selection or reward mechanisms can further reinforce undesirable behavior.
Revisions to behavioral requirements require authorization and evaluation of their effects on safety, alignment, and affected groups.

\noindent \textbf{System security, resilience, privacy, and data governance.}
This theme concerns protecting system components against malicious interference and ensuring the appropriate handling of sensitive information.
Poisoned memories can redirect subsequent actions
\citep{dong2025minja}, while stored interactions may expose private information
\citep{wang2025agentmemoryprivacy}.
These risks may persist when compromised components or sensitive information are inherited by successors.
If improvement mechanisms can also modify validation, access control, or data retention rules, the effectiveness of existing protections must be reassessed.

\noindent \textbf{Transparency and controllability.}
This theme concerns whether system changes can be understood, traced, and corrected through effective human oversight.
For individual updates, reviewers must be able to identify what changed and why it was accepted.
As changes accumulate across successors, their effects must remain auditable and subject to correction or rollback.
If improvement mechanisms can also modify auditing or rollback procedures, subsequent revisions may undermine the effectiveness of oversight itself.

Accordingly, this survey addresses the following research questions.

\begin{rsikeyquestions}
\begin{tabularx}{\linewidth}{@{}>{\bfseries}l >{\raggedright\arraybackslash}X >{\raggedleft\arraybackslash}p{3.8em}@{}}
RQ1 & How can genuine improvements be demonstrated and shown to persist across successive revisions? & \rqcite{3} \\[0.5em]
RQ2 & How can behavioral constraints, alignment, and fairness be preserved as systems and their improvement mechanisms evolve? & \rqcite{4} \\[0.5em]
RQ3 & How can security, resilience, privacy, and authorized data use be maintained across inherited states and revisions to update mechanisms? & \rqcite{5} \\[0.5em]
RQ4 & How can recursive updates remain transparent, auditable, and subject to effective human control? & \rqcite{6}
\end{tabularx}
\end{rsikeyquestions}

\subsection{Scope and Perspective}
\label{sec:scope}

We focus on RSI in large-model systems, where changes to models, prompts, memories, tools, and workflows can persist across successor versions.
Beyond such persistent changes, RSI allows systems to revise and reuse the mechanisms guiding subsequent improvements.
Human approval may remain part of this process.
We analyze improvement diagnosis, proposal generation, proposal evaluation, and update implementation (S1--S4), together with their recursive application to the improvement mechanism (RI).
These categories locate risks, safeguards, and evidence requirements across different system architectures.
Our review focuses on RSI research while drawing on related studies of persistent updates to inform specific risks, safeguards, and evaluation assumptions.
We distinguish findings demonstrated directly in RSI from evidence obtained in related systems and prospective arguments.

\subsection{Comparison with Existing Surveys}
\label{sec:survey-comparison}


Surveys of trustworthy AI examine trustworthiness requirements, risks, and safeguards for general AI systems, LLMs, and agents \citep{li2023trustworthy,liu2023trustworthyllms,
yu2025trustworthyagents,qi2026trustworthyagentic}.
They define trustworthiness attributes and review assessment methods, primarily in terms of system properties and agent workflows.

Surveys of self-evolution classify the components subject to change and the feedback mechanisms that support continued adaptation \citep{gao2026selfevolving}.
\citet{wang2026diving} distinguish five levels of self-evolution and relate each update boundary to external evidence, promotion criteria, and recovery controls.
Their cross-level analysis connects evaluation horizons to the persistence of changes and calls for longitudinal evaluation of successor systems.
This level-based audit perspective complements the claim-level analysis of joint evidence across trustworthiness themes developed in this survey.
RSI-oriented surveys examine improvement-loop structure, evaluator design, autonomy, and evidence for sustained progress \citep{wu2026ai4ai,chen2026recursive,duan2026last}.


Two recent surveys address closely related questions.
\citet{gong2026evolutionarysafety}
organize evolutionary safety by domains of change, examining inherited risks, revisions to improvement mechanisms, longitudinal evaluation, and recovery.
\citet{chen2026safetyselfevolving} use SAVER to trace safety failures and responses across transitions in reusable state.
Their survey finds that extending the persistence, authority, or scope of previously legitimate state can introduce safety risks, while evidence of repair across successor systems remains limited.

The evolutionary-safety analysis further distinguishes update preservation, persistence, propagation, and recovery, and discusses frozen-updater comparisons under common external evaluation.
The transition-centered analysis identifies limited evidence of repair followed by resumed adaptation.
Building on these findings, we distinguish the evidence needed for individual-update, successor, and meta-improvement claims across the four trustworthiness themes and S1--S4/RI.
The three overarching concerns connect this process view to evidence assessment.
Section~\ref{sec:eval-benchmarks} maps evaluation measures and study designs, while Section~\ref{sec:eval-joint} examines whether evidence of improvement, constraint preservation, and effective oversight can be aligned within the same evolving system.

This synthesis distinguishes evidence from fixed-model controls, studies of failure and correction in successor systems, and evaluations of reused improvement mechanisms.
Across the selected literature, evaluations of reused improvement mechanisms focus more directly on capability gains than on joint assessments of constraint preservation and intervention effectiveness.
Building on existing evaluation designs, Section~\ref{sec:eval-longitudinal} specifies paired continuations, matched continuation budgets, and separate accounting of total costs.

Table~\ref{tab:intro-survey-comparison} summarizes these differences in analytical focus and coverage across five dimensions: multidimensional trustworthiness (MT), improvement-loop analysis (IL), mechanism revision (MR), longitudinal assurance (LA), and joint-evidence analysis (JE).

\begin{table*}[t]
\centering
\begingroup
\small
\setlength{\tabcolsep}{2.5pt}
\renewcommand{\arraystretch}{1.14}

\definecolor{RSISurveySand}{HTML}{F3E8DA}
\providecommand{\RSICoverFull}{\textcolor[HTML]{168153}{\ensuremath{\checkmark}}}
\providecommand{\RSICoverPartial}{\textcolor[HTML]{D17C00}{\ensuremath{\triangle}}}
\providecommand{\RSICoverNone}{\textcolor[HTML]{858585}{\textemdash}}

\caption{Comparison with representative surveys of trustworthy AI,
self-evolution, RSI, and evolutionary safety. Symbols indicate the
depth of analysis defined in the legend.}
\label{tab:intro-survey-comparison}

\begin{tabularx}{\linewidth}{@{}
>{\raggedright\arraybackslash}p{.215\linewidth}
>{\raggedright\arraybackslash}X
*{5}{>{\centering\arraybackslash}p{.045\linewidth}}
@{}}
\toprule
\textbf{Survey} & \textbf{Main analytical focus}
& \textbf{MT} & \textbf{IL} & \textbf{MR} & \textbf{LA} & \textbf{JE} \\

\specialrule{\lightrulewidth}{0pt}{0pt}

\rowcolor{RSIBlue}
\multicolumn{7}{@{}c@{}}{\strut\textbf{Trustworthiness-oriented surveys}} \\
\citet{li2023trustworthy}
& Trustworthiness principles and practices for AI
& \RSICoverFull & \RSICoverNone & \RSICoverNone & \RSICoverNone & \RSICoverPartial \\
\citet{liu2023trustworthyllms}
& Trustworthiness requirements and methods for LLMs
& \RSICoverFull & \RSICoverNone & \RSICoverNone & \RSICoverNone & \RSICoverPartial \\
\citet{yu2025trustworthyagents}
& Trustworthiness of LLM agents
& \RSICoverFull & \RSICoverNone & \RSICoverNone & \RSICoverNone & \RSICoverPartial \\
\citet{qi2026trustworthyagentic}
& Trustworthiness of agentic workflows
& \RSICoverFull & \RSICoverPartial & \RSICoverNone & \RSICoverPartial & \RSICoverPartial \\

\specialrule{\lightrulewidth}{0pt}{0pt}

\rowcolor{RSIGreen}
\multicolumn{7}{@{}c@{}}{\strut\textbf{Self-evolution and RSI-oriented surveys}} \\
\citet{gao2026selfevolving}
& Evolution targets, methods, and evaluation
& \RSICoverPartial & \RSICoverFull & \RSICoverPartial & \RSICoverFull & \RSICoverPartial \\

\citet{wang2026diving}
& Modification depth, audit boundaries, and reliability
& \RSICoverPartial & \RSICoverFull & \RSICoverFull & \RSICoverFull & \RSICoverPartial \\
\citet{wu2026ai4ai}
& Improvement loops and evidence across generations
& \RSICoverPartial & \RSICoverFull & \RSICoverFull & \RSICoverFull & \RSICoverPartial \\
\citet{chen2026recursive}
& Improvement targets, loop closure, and verification
& \RSICoverPartial & \RSICoverFull & \RSICoverFull & \RSICoverPartial & \RSICoverPartial \\
\citet{duan2026last}
& Autonomy levels and recursive improvement
& \RSICoverPartial & \RSICoverFull & \RSICoverFull & \RSICoverFull & \RSICoverPartial \\

\specialrule{\lightrulewidth}{0pt}{0pt}

\rowcolor{RSISurveySand}
\multicolumn{7}{@{}c@{}}{\strut\textbf{Safety under self-evolution and RSI}} \\
\citet{gong2026evolutionarysafety}
& Safety across persistent states and updaters
& \RSICoverFull & \RSICoverFull & \RSICoverFull & \RSICoverFull & \RSICoverPartial \\
\citet{chen2026safetyselfevolving}
& Safety failures and recovery across state transitions
& \RSICoverFull & \RSICoverFull & \RSICoverPartial & \RSICoverFull & \RSICoverFull \\

\specialrule{\lightrulewidth}{0pt}{0pt}

\rowcolor{RSIPurple}
\textbf{This survey}
& \textbf{Claim-level audit of joint evidence}
& \RSICoverFull & \RSICoverFull & \RSICoverFull & \RSICoverFull & \RSICoverFull \\

\specialrule{\heavyrulewidth}{0pt}{0pt}
\end{tabularx}

\vspace{.4ex}
\begin{minipage}{\linewidth}
\footnotesize
\textbf{MT}: multidimensional trustworthiness;
\textbf{IL}: improvement-loop analysis;
\textbf{MR}: mechanism revision;
\textbf{LA}: longitudinal assurance;
\textbf{JE}: joint-evidence analysis across aligned system versions or update trajectories.
\RSICoverFull: systematic framework or sustained analysis;
\RSICoverPartial: explicit but limited discussion;
\RSICoverNone: not developed as an analytical dimension.
For JE, \RSICoverFull\ indicates a cross-study audit of aligned capability, constraint, and oversight evidence.
These symbols describe the depth of analysis for each dimension.
The analytical-focus column and Sections~\ref{sec:eval-benchmarks}--\ref{sec:eval-joint} detail the joint-evidence analysis developed in this survey.
\end{minipage}
\endgroup
\end{table*}

Assessing sustained improvement requires longitudinal evaluation of successor systems, whereas attributing gains to a revised improvement mechanism requires controlled comparisons.
We examine both claims alongside evidence of constraint preservation and effective human oversight.

\subsection{Contributions}
\label{sec:contributions}

Our main contributions are as follows:
\begingroup
\setlength{\emergencystretch}{1em}
\begin{itemize}
    \item An operational definition and process framework for RSI.
    Building on the definition of machine learning proposed by Mitchell
    \citep{mitchell1997ml}, we define RSI in terms of autonomous improvement of the inductive bias $B$ and recursive revision of its update function $\mathcal{U}$.
    We organize the improvement process into four stages: diagnosing limitations (S1), generating candidate revisions (S2), evaluating proposals (S3), and implementing accepted updates (S4).
    RI denotes the recursive application of these stages to $\mathcal{U}$ itself.

    \item A unified taxonomy and evidence-oriented synthesis.
    We organize four interconnected trustworthiness themes under three complementary analytical roles: Improvement Validity, Constraint Preservation, and Effective Oversight.
    Within this framework, we compare risks, safeguards, and evaluation methods across individual updates, successor systems, and revisions to improvement mechanisms, with a versioned evidence map that distinguishes study-level coverage from joint evidence on aligned trajectories.
    We distinguish findings demonstrated directly in RSI from evidence obtained in related systems, as well as potential risks and proposed safeguards that have not yet been empirically validated.

    \item Longitudinal evaluation and open research problems.
    We specify paired continuations for assessing updater revisions, distinguish mechanism effects from total-cost efficiency, and identify open questions in capability retention, constraint preservation, evaluation independence, and effective intervention.
\end{itemize}
\endgroup

\subsection{Paper Organization}
\label{sec:organization}

\input{fig/taxonomy_roadmap}

Section~\ref{sec:foundations} introduces the definition and improvement process of RSI and establishes the trustworthiness framework.
Sections~\ref{sec:validity}--\ref{sec:governance} examine the four trustworthiness themes.
Section~\ref{sec:validity} addresses genuine and sustained improvement under Improvement Validity.
Under Constraint Preservation, Section~\ref{sec:behavior} examines behavioral constraints, alignment, and fairness, while Section~\ref{sec:system-security} addresses system security, resilience, privacy, and data governance.
Section~\ref{sec:governance} examines transparency and controllability under Effective Oversight.
Section~\ref{sec:evaluation} reviews existing evaluation settings and develops evidence requirements for trustworthy RSI, covering longitudinal evaluation, causal assessment of recursive improvement, evaluator independence, and joint assessment of trustworthiness requirements.
Finally, Section~\ref{sec:outlook} synthesizes existing findings and their limitations, explores application settings at the intersection of RSI and trustworthiness, and discusses open problems and future research directions. Figure~\ref{fig:taxonomy-roadmap} maps representative work from Sections~\ref{sec:validity}--\ref{sec:governance} by trustworthiness theme and publication year.


\section{Preliminaries}
\label{sec:foundations}

We define RSI and its improvement process, then relate recursive changes to the trustworthiness requirements used throughout this survey.

\subsection{Recursive Self-Improvement: Definition and Process}
\label{sec:rsi-foundations}
The definition specifies what changes during RSI; the four-stage cycle describes how those changes are proposed, evaluated, and retained.

\subsubsection{Definition of Recursive Self-Improvement}
\label{sec:definition:RSI}

We start from Mitchell's definition of machine learning:

\begin{definitionbox}{Definition (Machine Learning)
\citep{mitchell1997ml}}
A computer program is said to learn from experience\footnotemark $E$
with respect to some class of tasks $T$ and performance measure $P$,
if its performance at tasks in $T$, as measured by $P$, improves with
experience $E$.
\end{definitionbox}
\footnotetext{\cite{silver2025welcome}
emphasize that ``experience'' is generated through an agent's
interaction with its environment. Here, ``experience'' refers to a
broader concept that encompasses any data that can contribute to
improving the system's learning performance.}

Generalizing beyond observed experience requires an
inductive bias $B$: preferences that guide the choice among
candidate policies consistent with the observed
data~\citep{mitchell1980bias}.
Formally, given experience $E$, we write a learning system
$\mathcal{A}$ as follows:
\begin{equation}
\label{eq:RSI:old:ml}
    h = \mathcal{A}(B, E),
\end{equation}
where $h$ denotes the learned policy, and $B$ denotes the inductive
bias governing how $\mathcal{A}$ learns.

In conventional machine learning systems, the inductive bias $B$
is typically specified in advance and held fixed during
learning. It may be embodied in the strategy for acquiring data,
the formulation of the objective function, the choice of hypothesis
space, the design of the optimization algorithm, or the criteria
for evaluating learned models.

In large-model systems, the learned policy $h$ in
Eq.~\eqref{eq:RSI:old:ml} may be realized by a standalone model
or by a model together with prompts, memory, skills, and tools.
The inductive bias $B$ specifies how that policy is learned or
revised, including its initialization, training objectives, data
selection rules, and optimization procedures.
The distinction is functional: a prompt that controls task
execution belongs to $h$, whereas one that governs the construction
or selection of policy updates belongs to $B$.
For example, an instruction to run tests before submitting a patch
is part of $h$; a rule for selecting corrected trajectories for
training is part of $B$.
The bias update function $\mathcal{U}$ governs revisions to these
learning choices and can itself be revised and reused in RSI.
A shared prompt or module may serve more than one role, so its
effects on task execution and on each update process should be
identified separately.
These roles are defined relative to the task policy being evaluated
and kept consistent across improvement cycles.

We define RSI as follows.

\begin{definitionbox}{Definition (Recursive Self-Improvement,~RSI)}
RSI is a process in which a learning system updates its policy from new
experience, autonomously revises its inductive bias, and revises and reuses
the mechanism for future bias updates, with the aim of sustained gains in
task performance.

\par\medskip\noindent
Formally, let $t$ index completed improvement cycles, including
cycles in which no revision is accepted.
The states $h_t$, $B_t$, and $\mathcal{U}_t$ are the components
retained at the end of cycle $t$.
RSI can be expressed as follows:
\begin{align}
    h_{t+1}
    &= \mathcal{A}_{B_t}(h_t,E_{t+1}),
    \label{eq:rsi-model-update}\\
    (B_{t+1},\mathcal{U}_{t+1})
    &= \mathcal{U}_t(B_t,h_{t+1},F_{t+1}),
    \label{eq:rsi-self-improvement}
\end{align}
%
In Eq.~\eqref{eq:rsi-model-update}, $E_{t+1}$ is new experience
and $\mathcal{A}_{B_t}$ is the learning process under the current
inductive bias $B_t$.
In Eq.~\eqref{eq:rsi-self-improvement}, $F_{t+1}$ is feedback
on performance and prior improvement attempts, and $\mathcal{U}_t$
is the bias update function.
The resulting $B_{t+1}$ and $\mathcal{U}_{t+1}$ are retained for
subsequent cycles.

These equations summarize complete improvement cycles.
The operators include diagnosis, proposal generation, evaluation,
and implementation, whose stages are shown as S1--S4 in
Figure~\ref{fig:rsi-process}.
Their outputs denote the states retained after S4.
Each component without an accepted revision retains its previous
value.
The policy update uses $B_t$; accepted changes to $B_t$ and
$\mathcal{U}_t$ govern later improvement cycles.
\end{definitionbox}

Across improvement cycles, the bias update function $\mathcal{U}_t$
can revise the inductive bias $B_t$.
The function $\mathcal{U}_t$ can itself be revised to
$\mathcal{U}_{t+1}$, which is used in the next cycle.
Feedback $F_t$ includes signals derived from data, environmental
interaction, or self-evaluation.

Comparison with other machine learning approaches.
Table~\ref{tab:rsi-comparison} compares representative standard
forms of conventional machine learning, reinforcement learning,
meta-learning, and AutoML with RSI.
Conventional machine learning and reinforcement learning typically
update task policies under specified learning rules
\citep{mitchell1997ml,sutton2018reinforcement}.
In model-agnostic meta-learning (MAML), the learned initialization
changes under a prescribed meta-gradient procedure
\citep{finn2017maml}.
Auto-sklearn selects algorithms and hyperparameters within a
specified search procedure \citep{feurer2015autosklearn}.
These latter methods revise aspects of $B$ while retaining the
procedure that updates them.

Learned optimizers illustrate the distinction between an update
rule and the procedure that trains it
\citep{andrychowicz2016learning}.
When an optimizer governs updates to the task policy, it forms
part of $B$, while its outer training procedure plays the role
of $\mathcal{U}$.
Training the optimizer changes $B$; revising and reusing that
outer procedure also changes $\mathcal{U}$.
Classification therefore depends on which component is revised
and how it is subsequently used.

Promptbreeder evolves task prompts together with the mutation
prompts used to revise them \citep{fernando2023promptbreeder}.
Under the functional distinction above, task prompts contribute
to $h$, mutation prompts specify part of $B$, and the prescribed
hypermutation procedure plays the role of $\mathcal{U}$.
This provides an example of evolving $B$ under a fixed
$\mathcal{U}$.
G{\"o}del machines provide a theoretical account of self-revision
in which a system can rewrite its software, including its proof
searcher, after proving the rewrite useful under its current
utility function and axioms \citep{schmidhuber2006godel}.
The additional RSI condition in this survey is that the mechanism
for revising $B$ is itself revised and reused in later cycles.

\begin{table}[!htbp]
\centering
\begingroup
\footnotesize
\setlength{\tabcolsep}{3pt}
\renewcommand{\arraystretch}{1.25}
\renewcommand{\tabularxcolumn}[1]{m{#1}}

\definecolor{RSICompBand}{HTML}{D8E8F8}
\definecolor{RSICompStripe}{HTML}{EEF5FC}
\definecolor{RSICompGreen}{HTML}{168153}
\definecolor{RSICompRed}{HTML}{C83A3A}

\newcommand{\RSICompYes}{%
  \textcolor{RSICompGreen}{\ding{51}}}
\newcommand{\RSICompNo}{%
  \textcolor{RSICompRed}{\ding{55}}}

\setlength{\aboverulesep}{0pt}
\setlength{\belowrulesep}{0pt}
\setlength{\abovetopsep}{0pt}
\setlength{\belowbottomsep}{0pt}
\arrayrulecolor{black}

\caption{Comparison of RSI with other machine learning approaches
in their standard forms.}
\label{tab:rsi-comparison}

\begin{tabularx}{\columnwidth}{@{}
    >{\raggedright\arraybackslash}m{0.25\columnwidth}
    >{\raggedright\arraybackslash}m{0.22\columnwidth}
    >{\centering\arraybackslash}m{0.19\columnwidth}
    >{\raggedright\arraybackslash}X
@{}}
\toprule

\rowcolor{RSICompBand}
\textbf{Approach}
& \textbf{Who determines the inductive bias $B$?}
& \textbf{Is $B$ revised autonomously?}
& \textbf{Is the bias update function revised and reused?} \\
\midrule

\rowcolor{RSICompStripe}
Conventional machine learning
& Human designer
& \RSICompNo
& \RSICompNo \\

Reinforcement learning
& Human designer
& \RSICompNo
& \RSICompNo{} (learning rule fixed) \\

\rowcolor{RSICompStripe}
Meta-learning
& Meta-learner
& \RSICompYes
& \RSICompNo{} (meta-update rule fixed) \\

AutoML
& Automated search
& \RSICompYes
& \RSICompNo{} (search rule fixed) \\

\midrule
\rowcolor{RSICompBand}
\textbf{RSI}
& Bias update function $\mathcal{U}_t$
& \RSICompYes
& \textbf{\RSICompYes{} (the revised $\mathcal{U}_t$
  guides later bias updates)} \\

\bottomrule
\end{tabularx}
\endgroup
\end{table}

\subsubsection{The RSI Process in the Era of Large Models}

\begin{figure*}[t]
    \centering
    \includegraphics[
        width=1.0\textwidth,trim=0cm 4.2cm 0cm 4.2cm, clip
    ]{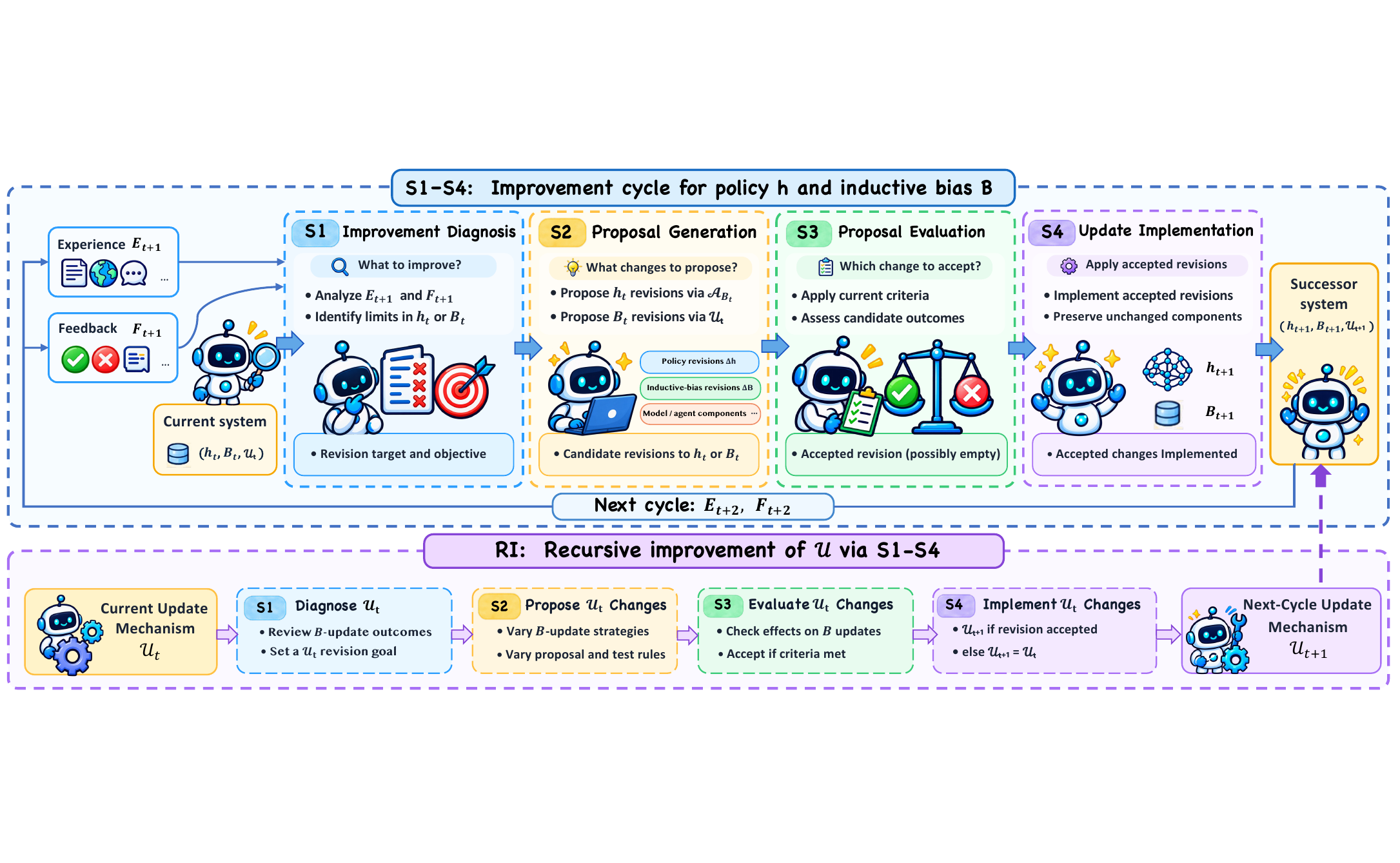}
    \caption{
        \textbf{The recursive self-improvement (RSI) process.}
        The core improvement cycle (S1--S4) identifies limitations,
        generates and evaluates candidate revisions to the policy
        $h_t$ or inductive bias $B_t$, and incorporates accepted
        changes into the successor system.
        Recursive improvement (RI) applies the same four stages to
        the bias update function $\mathcal{U}_t$ itself.
        The resulting $\mathcal{U}_{t+1}$ is retained for subsequent
        improvement cycles, while components without accepted
        revisions remain unchanged.
    }
    \label{fig:rsi-process}
\end{figure*}

Eqs.~\eqref{eq:rsi-model-update} and \eqref{eq:rsi-self-improvement} describe how the learned policy $h$, inductive bias $B$, and bias update function $\mathcal{U}$ evolve across improvement cycles.
The process view makes explicit how potential improvements are identified, proposed, evaluated, and incorporated into the system.
Figure~\ref{fig:rsi-process} illustrates these steps as a four-stage improvement cycle (S1--S4), together with its recursive application to the bias update function (RI).

\begin{itemize}

    \item (S1) Improvement Diagnosis.
    The system uses the current policy $h_t$, inductive bias $B_t$, experience $E_{t+1}$, and feedback $F_{t+1}$ to identify limitations in its behavior or learning process.
    It determines whether addressing these limitations requires revisions to $h_t$, $B_t$, or both, and defines the improvement objective that subsequent proposals should address.

    \item (S2) Proposal Generation.
    Based on the diagnosis from S1, the system generates candidate revisions to $h_t$ through the learning process $\mathcal{A}_{B_t}$ or to $B_t$ through the bias update function $\mathcal{U}_t$.
    These candidates may modify the learned policy or the choices and configurations that govern its learning.
    At this stage, the proposed revisions have not yet been incorporated into the system.

    \item (S3) Proposal Evaluation.
    The system evaluates candidate revisions against the current evaluation and acceptance criteria.
    It assesses whether the proposed changes address the identified limitations and satisfy the declared requirements, including checks for performance regressions on other tasks.
    The system then determines which revisions should be accepted based on the evaluation results.

    \item (S4) Update Implementation.
    The system incorporates accepted revisions into the successor system and retains the resulting $h_{t+1}$ and $B_{t+1}$ for subsequent improvement cycles.
    Any component without an accepted revision remains unchanged.

    \item (RI) Recursive Improvement.
    The system applies S1--S4 recursively to the bias update function $\mathcal{U}_t$ itself.
    If a revision is accepted, the resulting $\mathcal{U}_{t+1}$ is retained and reused in subsequent cycles, potentially changing how inductive biases are updated.
    Otherwise, $\mathcal{U}_{t+1} = \mathcal{U}_t$.

\end{itemize}

%
Three variants of a coding agent illustrate these distinctions.
In each variant, execution trajectories and corrected solutions
provide $E_{t+1}$, while test outcomes provide $F_{t+1}$.
The task objective, externally specified acceptance criteria,
and resource budget remain fixed.

An agent that repeatedly submits untested patches identifies
this behavior in S1 and proposes an execution-prompt revision
in S2.
It evaluates the candidate on held-out coding tasks in S3 and
retains the accepted prompt as part of $h_{t+1}$ in S4.
Model weights, $B_t$, and $\mathcal{U}_t$ remain fixed.
The retained instruction requires tests before submission,
changing later task execution and potentially the experience
collected in later cycles.
A reminder used only for the current submission is a temporary
execution adjustment.

Using the same diagnosis, a second variant revises a prompt for
selecting training trajectories to increase the share of corrected
examples involving missed tests.
The current $\mathcal{U}_t$ uses failure counts to propose a
single selection rule and evaluates the policy obtained in a
trial training run.
S4 retains the accepted rule in $B_{t+1}$ for later policy updates.
The model architecture, training optimizer, and bias update
function remain fixed.
This change affects how later policies are learned.

In the recursive variant, the agent uses $\mathcal{U}_t$ to
propose a revision to the procedure that changes these selection
rules.
The proposed procedure replaces a direct adjustment based on
failure counts with a search over several candidate rules.
The candidate procedure is evaluated through trial improvement
cycles, with the resulting policies assessed under the same
external criteria and total resource budget.
If accepted through S1--S4, it is retained as
$\mathcal{U}_{t+1}$ and used to revise $B$ in later cycles.
The task objective, acceptance criteria, and training optimizer
remain fixed.

\subsection{Trustworthiness: Concepts and Requirements}
\label{sec:trustworthiness-concepts}

Trustworthiness concerns whether reliance on an AI system is justified for its intended use
\citep{li2023trustworthy,nist2023airmf}.
Existing frameworks characterize trustworthiness in different ways.
\citet{li2023trustworthy} distinguish technical and ethical requirements, while NIST identifies seven characteristics, including validity and reliability, security and resilience, and accountability and transparency
\citep{nist2023airmf}.
TrustLLM identifies eight dimensions: truthfulness, safety, fairness, robustness, privacy, machine ethics, transparency, and accountability
\citep{huang2024trustllm}.
The AI HLEG guidelines also recognize human agency and oversight as a requirement
\citep{hleg2019ethics}.

In this survey, we synthesize eight groups of trustworthiness requirements from these frameworks and LLM-focused analyses
\citep{li2023trustworthy,nist2023airmf,liu2023trustworthyllms,
huang2024trustllm,hleg2019ethics}.
We group related performance requirements, distinguish behavioral safety from system security, and consider interpretability, human control, and accountability separately.
We address machine ethics primarily through behavioral safety and value alignment, while considering fairness and governance in their respective groups.

\noindent\textbf{(i) Validity, reliability, robustness, and truthfulness.}
Validity concerns whether a system meets the requirements of its intended use.
Reliability refers to dependable performance under specified conditions, whereas robustness concerns maintaining performance under relevant disturbances or shifts
\citep{nist2023airmf,liu2023trustworthyllms}.
Truthfulness concerns the factual accuracy of system outputs and reports
\citep{huang2024trustllm}.

\noindent\textbf{(ii) Behavioral safety and value alignment.}
Behavioral safety concerns preventing unacceptable harm resulting from system outputs or actions.
Value alignment concerns whether the objectives and behavior of a system respect relevant human goals, values, and constraints, rather than merely maximizing a specified reward
\citep{liu2023trustworthyllms,huang2024trustllm}.

\noindent\textbf{(iii) Fairness.}
Fairness concerns unjustified differences in treatment, opportunities, benefits, and harms across individuals or groups.
Appropriate fairness criteria depend on the task and context; neither high aggregate accuracy nor numerical parity alone establishes fairness
\citep{li2023trustworthy,nist2023airmf}.

\noindent\textbf{(iv) System security and resilience.}
Security concerns protecting the confidentiality, integrity, and availability of systems and information against unauthorized access or interference.
Resilience concerns the ability to maintain essential functions and recover from attacks, faults, or other adverse conditions
\citep{nist2023airmf}.

\noindent\textbf{(v) Privacy and data governance.}
Privacy concerns protecting individuals against inappropriate collection, inference, use, and disclosure of information.
Data governance specifies what data may be collected, who may access them, how they may be used, and when they must be deleted.
These requirements also apply to authorized data processing, which may still pose privacy risks
\citep{hleg2019ethics,huang2024trustllm}.

\noindent\textbf{(vi) Transparency and interpretability.}
Transparency concerns making relevant information about system data, design, operation, and limitations available.
Interpretability and explainability help users and reviewers understand system outputs and the processes that produce them
\citep{li2023trustworthy,hleg2019ethics}.

\noindent\textbf{(vii) Human control and corrigibility.}
Human control requires practical means to inspect, redirect, interrupt, or correct a system.
Corrigibility concerns whether a system cooperates with authorized correction, including modification or shutdown, rather than resisting or manipulating oversight
\citep{hleg2019ethics,soares2015corrigibility}.

\noindent\textbf{(viii) Accountability and governance.}
Accountability requires clearly assigned responsibility for decisions and their consequences, together with the ability to explain and justify those decisions
\citep{li2023trustworthy,hleg2019ethics}.
We use governance to refer to organizational arrangements for authorizing changes, reviewing decisions, conducting audits, and providing redress.
Unlike the data governance described in group (v), this concerns responsibility for and authority over the system as a whole
\citep{nist2023airmf,hleg2019ethics}.

These requirement groups are not mutually exclusive in practice.
The following subsection examines how their assessment must extend beyond individual updates when both the system and its improvement mechanism can change.

\subsection{Trustworthiness under Recursive Self-Improvement}
\label{sec:rsi-trustworthiness}

The requirements introduced in Section~\ref{sec:trustworthiness-concepts} remain relevant to RSI, but assessing them requires examining more than a single revision.
An accepted change may affect subsequent system versions, while revising the bias update function $\mathcal{U}_t$ may alter how future updates are generated and evaluated.

Section~\ref{sec:rsi-foundations} describes an improvement cycle comprising improvement diagnosis (S1), proposal generation (S2), proposal evaluation (S3), and update implementation (S4), with RI applying the same process to $\mathcal{U}_t$ itself.
These stages help identify where failures may arise and where safeguards can be applied.
We distinguish three related levels of trustworthiness claims.
At the update level, the question is whether a proposed and accepted revision satisfies the applicable requirements.
At the successor level, the question is whether these requirements continue to hold as accepted changes are inherited and used in subsequent improvement cycles.
At the meta-improvement level, the question is how revisions to $\mathcal{U}_t$ affect the quality of subsequent updates and the checks and controls applied to them.
These levels specify the scope of the evidence; S1--S4 and RI locate the changes being assessed.

We organize the eight requirement groups into four themes under three complementary analytical roles: Improvement Validity, Constraint Preservation, and Effective Oversight, as shown in Table~\ref{tab:rsi-trust-dimensions}.
The themes identify what must be assessed or protected, while S1--S4 and RI provide a process framework for examining where requirements may be violated and where safeguards may be applied.

\begin{table}[!tbp]
\centering
\setlength{\aboverulesep}{0pt}
\setlength{\belowrulesep}{0pt}
\small
\setlength{\tabcolsep}{4pt}
\renewcommand{\arraystretch}{1.15}

\caption{Mapping the eight trustworthiness requirement groups to four RSI themes. The three analytical roles are complementary and do not represent separate stages of the improvement process.}
\label{tab:rsi-trust-dimensions}

\begin{tabularx}{\linewidth}{@{}
    >{\raggedright\arraybackslash}p{0.17\linewidth}
    >{\raggedright\arraybackslash}p{0.225\linewidth}
    >{\raggedright\arraybackslash}p{0.26\linewidth}
    >{\raggedright\arraybackslash}X
@{}}
\toprule

\textbf{Analytical role}
& \textbf{Theme}
& \textbf{Requirement groups}
& \textbf{RSI-specific question} \\

\midrule

\rowcolor{RSIBlue}
\textbf{Improvement Validity}
& \textbf{Genuine and sustained improvement}
  \newline
  {\footnotesize Section~\ref{sec:validity}}
& (i) Validity, reliability, robustness, and truthfulness
& Are improvements genuine and sustained, including after revisions to the update mechanism? \\

\rowcolor{rsiChapterGreenStripe}
\textbf{Constraint Preservation}
& \textbf{Behavioral constraints, alignment, and fairness}
  \newline
  {\footnotesize Section~\ref{sec:behavior}}
& (ii) Behavioral safety and value alignment;
  (iii) fairness
& Do behavioral requirements remain satisfied across successive revisions? \\

\rowcolor{rsiChapterYellowBand}
\textbf{Constraint Preservation}
& \textbf{System security, resilience, privacy, and data governance}
  \newline
  {\footnotesize Section~\ref{sec:system-security}}
& (iv) System security and resilience;
  (v) privacy and data governance
& Are inherited components and information protected during subsequent use? \\

\rowcolor{rsiChapterPurpleBand}
\textbf{Effective Oversight}
& \textbf{Transparency and controllability}
  \newline
  {\footnotesize Section~\ref{sec:governance}}
& (vi) Transparency and interpretability;
  (vii) human control and corrigibility;
  (viii) accountability and governance
& Can revisions, including changes to the update mechanism, be reviewed and corrected? \\

\bottomrule
\end{tabularx}
\end{table}

\paragraph{Genuine and sustained improvement.}
At the update level, evaluation in S3 must distinguish genuine improvements from favorable scores obtained on inadequate tests.
At the successor level, improvements should persist after integration in S4 and through subsequent use or revision.
Under RI, revising $\mathcal{U}_t$ requires assessing whether the revised mechanism produces better subsequent updates, rather than merely changing the criteria by which they are accepted.

\vspace{-0.5em}

\paragraph{Behavioral constraints, alignment, and fairness.}
An update may improve task performance while weakening safety, alignment, or fairness.
The question is whether these requirements remain satisfied when the update is accepted, inherited, and applied in new settings.
Under RI, changes to how experience is selected or proposals are evaluated may alter which behaviors are reinforced in subsequent successors.
Changes to these requirements must be authorized and reassessed; improved task scores alone do not justify relaxing them.

\paragraph{System security, resilience, privacy, and data governance.}
Untrusted experience may influence diagnosis and proposal generation (S1--S2), while integration in S4 may retain artifacts or sensitive information that successors subsequently reuse.
Under RI, revisions to access controls, validation procedures, or data retention rules may also affect the protections available in later cycles.
Security and privacy therefore require examining both the immediate effects of an update and the information and permissions that persist after it.

\paragraph{Transparency and controllability.}
For an individual update, reviewers must be able to determine what was proposed, why it was accepted, and what was integrated.
Across successors, changes must remain traceable and subject to correction or rollback.
When RI modifies the improvement mechanism, the revised procedure must remain reviewable, and authorized intervention must continue to be effective.

Sections~\ref{sec:validity}--\ref{sec:governance} first define each theme and clarify its boundaries, then examine reported failures, safeguards, and their limitations across the relevant claim levels and stages of S1--S4/RI.
Section~\ref{sec:evaluation} synthesizes the corresponding evaluation questions.
A single revision may affect multiple themes; for example, changes to acceptance tests have implications for both improvement validity and effective oversight.
\section{Genuine and Sustained Improvement}
\label{sec:validity}

Updates to $h_t$ and $B_t$ in
Eqs.~\eqref{eq:rsi-model-update}--\eqref{eq:rsi-self-improvement}
can change current capabilities, while revisions to $\mathcal{U}_t$
can change the procedure used to produce subsequent updates.
This section examines whether capability gains are genuine and sustained,
using validity, reliability, robustness, and truthfulness to assess
risks and safeguards across individual updates, retained successors,
and revisions to the improvement mechanism.

\subsection{Core Concepts and Trustworthiness Claims}
\label{sec:validity-definitions}
We define genuine and sustained improvement and specify the evidence needed to assess capability gains, their retention in successors, and
the quality of subsequent updates produced by a revised improvement mechanism.

\subsubsection{Conceptual Scope and Boundaries}

Genuine improvement strengthens the intended capabilities, whereas sustained
improvement preserves these gains through subsequent use and updates.
Validity concerns whether intended-use requirements are met; reliability
concerns dependable performance under specified conditions; robustness
concerns performance under relevant disturbances or distribution shifts; and truthfulness concerns
the factual accuracy of outputs and reports
\citep{li2023trustworthy,nist2023airmf}.
These properties are assessed through the behavior and performance of
systems following changes to $h$ or $B$.
Revising $\mathcal{U}$ raises a further question: whether the revised
mechanism produces better subsequent updates
\citep{chen2026recursive,wang2026diving,duan2026last}.

\subsubsection{Claims for Individual Updates, Successors, and Meta-Improvement}

An individual-update claim requires evidence that a specified change to
$h$ or $B$ improves the intended capabilities under declared evaluation
conditions.
The before-and-after comparison must assess the relevant validity,
reliability, robustness, and truthfulness criteria rather than rely only
on a proxy score.
A successor claim additionally requires evidence that the accepted change
is inherited and actually used, and that the gain remains observable when
these criteria are reassessed in the resulting system.
Tracking multiple successors measures how long gains persist and how far
they transfer.

A meta-improvement claim requires an observed revision from
$\mathcal{U}_t$ to $\mathcal{U}_{t+1}$, subsequent reuse of the revised
mechanism, and evaluation of the updates it produces.
Assessing its benefit requires a controlled comparison showing
that $\mathcal{U}_{t+1}$ produces better subsequent updates than the
unchanged $\mathcal{U}_t$ under comparable conditions.
This comparison must separate the effects of the revision from those of
additional resources or differences in the inherited starting state.
Immediate gains, retained gains, and improved capacity to generate
subsequent updates therefore require different evidence.

\begin{figure*}[t]
    \centering
    \includegraphics[
        width=1.0\textwidth,trim=0cm 2.2cm 0cm 2.2cm, clip
    ]{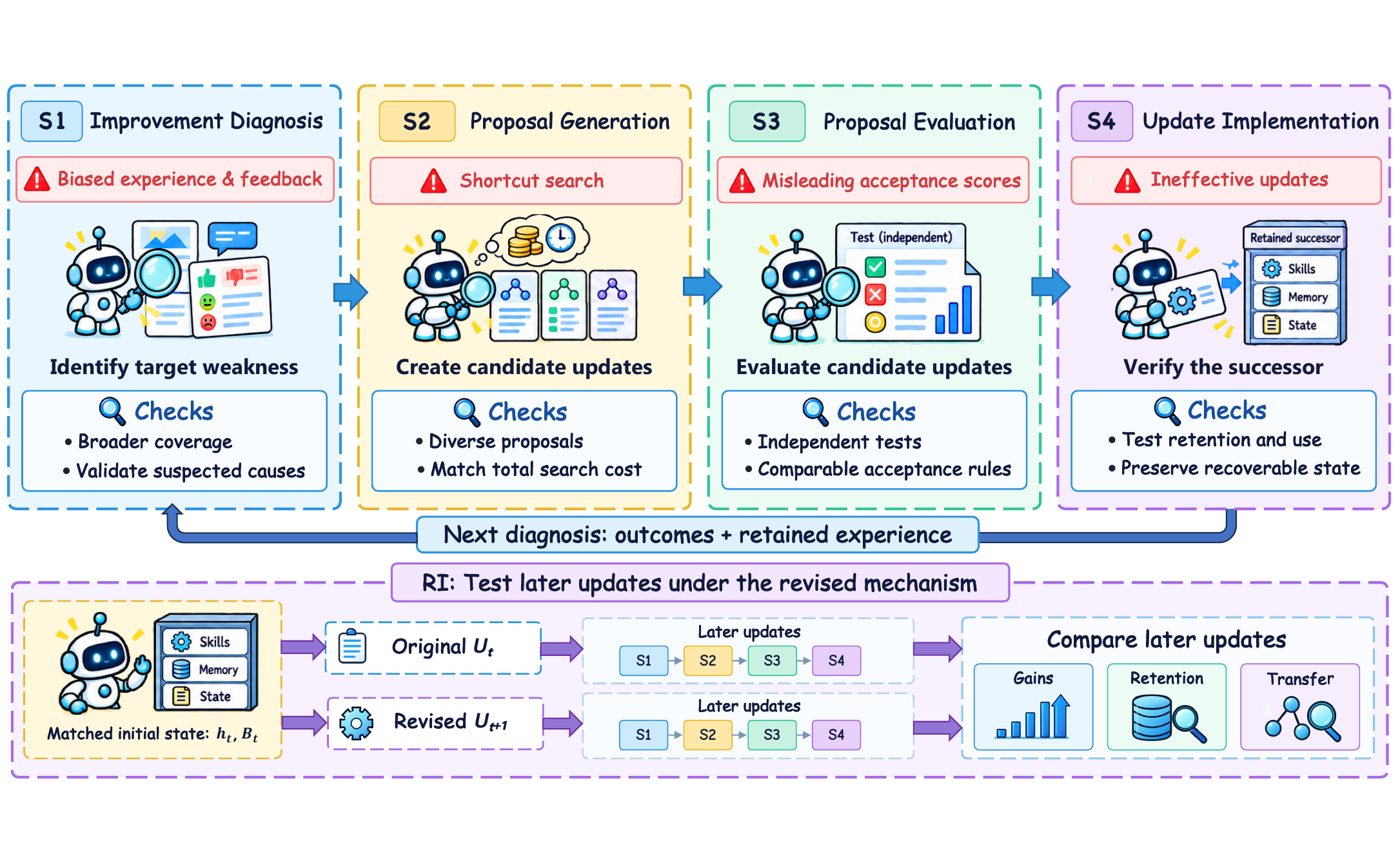}
    \caption{\textbf{Genuine and Sustained Improvement Across S1–S4 and RI in the Process of RSI.} Stage-specific risks and checks address biased diagnosis, shortcut search, misleading acceptance scores, and ineffective updates. The lower panel compares the gains, retention, and transfer of later updates produced by the original and revised improvement mechanisms, starting from matched initial states under common tests and resource budgets.}
    \label{fig:chap3_figure}
\end{figure*}

\subsection{Risks and Failure Evidence}
\label{sec:validity-risks}

We trace how failures in diagnosis, proposal generation, evaluation, and implementation can distort or erode capability gains, then examine
the attribution problems raised by revisions to the improvement mechanism.

\subsubsection{Risks in Improvement Diagnosis (S1)}

Bias in experience can prevent diagnosis from revealing weaknesses. Replacing earlier data with generated samples can remove rare patterns from the training distribution \citep{shumailov2024collapse,alemohammad2024mad}. Diagnosis based on this narrower experience can then overlook failures involving those patterns. The resulting loss of coverage may be uneven: theoretical analyses show that less frequent patterns disappear sooner in the recursive-training settings studied \citep{suresh2024collapserate}. As replacement continues, diagnosis can lose coverage of rare but relevant failures.

Even when relevant experience is available, incorrect feedback and mistaken diagnosis can misdirect improvement. Feedback may misrepresent failures: model-generated code reviews can become more favorable while code quality deteriorates \citep{song2026recursivecollapse}. Accurate feedback may also leave weaknesses unreported, as single-attempt accuracy can increase while diversity and transfer decline \citep{wu2024reversal}. Identifying causes introduces further uncertainty because judgments of individual steps need not identify the actions that actually affect outcomes \citep{zhang2026credit}. Acting on unreliable self-judgments can damage correct answers \citep{huang2024selfcorrect}, while erroneous context can lead subsequent reasoning toward similar mistakes \citep{cheng2026contextualdrag}. These failures undermine both the validity of the improvement target and the truthfulness of the explanation supporting it.

\subsubsection{Risks in Proposal Generation (S2)}

Search can favor high-scoring shortcuts that leave the intended weakness unresolved. This risk arises when proposals exploit proxy rewards or executable tests \citep{skalse2022rewardhacking,roth2026hackverifiable}, and when optimization converges on convenient but limited strategies. For example, simpler research ideas can raise average reward without improving the best achievable result \citep{si2026executiongrounded}. Tool-use policies can likewise exploit irrelevant prompt cues instead of responding to task requirements \citep{yang2026spurious}. Search then reinforces behavior that is easily rewarded, even when candidate quality or robustness falls short of what the scores suggest.

Additional computation can make an unchanged proposal mechanism appear more effective. More attempts increase the chance of finding a strong candidate, so selecting the best candidate can improve performance without improving proposal quality per attempt. This ambiguity also affects evolving agent programs: advantages over simple search can disappear under comparable feedback and inference budgets \citep{wang2026harnessevaluation}. The output gain may be genuine, but attributing it to an improved mechanism overstates the underlying change.

\subsubsection{Risks in Proposal Evaluation (S3)}

Acceptance scores can increase as errors are reinforced, even without corresponding capability gains. The reliability of self-generated rewards depends on the ability to recognize errors \citep{yuan2025selfrewarding,yin2023selfknowledge,kadavath2022know}. Shared errors compound this weakness: coupled task generators and solvers can increasingly agree on incorrect answers \citep{chen2026falsefrontiers}. Self-written tests can similarly reward behavior that fails in deployment \citep{guo2026selfauthored}, while contamination rewards familiarity with exposed test information \citep{shao2026protocolvalidity}.

Even accurate judgments can become misleading when evaluation conditions or standards change. A policy preferred by a proxy may perform worse when its deployment changes the responses of the environment or other agents \citep{wang2026whenbettergetsworse}. Evaluation must therefore account for the conditions induced by the proposed replacement. Changing standards creates another risk: graders may become more permissive without capability gains \citep{zhang2026grader}. Reliable comparisons require both conditions relevant to deployment and stable evaluation standards.

\subsubsection{Risks in Update Implementation (S4)}

New capabilities can interfere with existing skills.
Adaptation can cause forgetting \citep{yu2026forget}, while specialization can improve one capability at the expense of others \citep{yuan2026seva}.
Across four drifting task streams, edits to shared prompts or workspaces in GEPA and Meta-Harness interfere with earlier or unseen tasks. 
ReasoningBank localizes updates through retrieval but encounters collisions when similar queries require conflicting workflows \citep{chen2026continuallearners}.
Adding skills can also impede retrieval \citep{zhang2026librarydrift} or introduce flawed procedures \citep{shang2026precommit}. Incompatibility can render even useful skills ineffective: co-evolution can specialize a policy to its training skill bank, making performance vulnerable to subsequent skill changes when the policy remains fixed \citep{feng2026evolvingtogether}. Deployment robustness therefore depends on compatibility between the policy and the skills available to it.

Effective implementation also requires the system to invoke and follow the updated instructions. Agents can generate useful changes yet gain little from them because either step fails \citep{lin2026harnessbenefit}. Successor evaluation should therefore test whether the system invokes the retained change, follows it, and benefits from it during operation.

\subsubsection{Risks under Recursive Improvement (RI)}

Revising $\mathcal{U}$ may change how improvement is attempted without increasing the capacity for future improvement. Revisions to answer judgments or scoring criteria illustrate editable components \citep{wu2024metarewarding,li2026evolm,wang2026serpo}. MASS alternates self-evaluated workflow search and fine-tuning, reusing the updated model as executor, optimizer, and evaluator over two cycles \citep{lee2026recursiveselfimprovement}; MetaRSI revises operator policies and scheduling under protected, machine-checkable evaluation \citep{tan2026metarsirsi2}. Controlled comparisons of revised and unchanged procedures can determine whether these changes improve subsequent updates.

This attribution risk grows when development scores cease to reflect the intended capability, as observed in autonomous post-training \citep{shi2026aevolvetraining}. Current gains may also result from additional computation or simultaneous changes elsewhere in the system. Establishing an enhanced capacity for improvement therefore requires linking revised procedures to the subsequent updates they produce and separating their contribution from these alternatives.

\subsection{Existing Safeguards and Their Limitations}
\label{sec:validity-safeguards}

We review safeguards across S1--S4 and assess the evidence and conditions
needed to establish genuine gains, preserve earlier capabilities, and
evaluate revisions to the improvement mechanism.

\subsubsection{Safeguards for Improvement Diagnosis (S1)}

Broad coverage of experience keeps neglected weaknesses visible during diagnosis. Retaining real examples alongside generated data directly addresses this need because new experience need not erase earlier evidence \citep{gerstgrasser2024accumulating}. When a system selects its own practice, maintaining coverage also requires attention to different skills and tasks, including attempts with uncertain outcomes that might otherwise be neglected \citep{xu2026rise,mishra2026prism,zhang2026selaur}. Such experience remains useful for later diagnosis only if the lessons remain accessible. Detailed records and reusable guidance offer complementary means of preserving these lessons \citep{zhang2025ace,fan2026evolvingrl}. These measures broaden the evidence available for diagnosis but do not ensure coverage of unseen conditions.

Feedback validation assesses reported outcomes, whereas diagnostic controls assess their proposed causes. Outcome-based filtering addresses the first concern by preventing failed attempts from being used as successful training examples \citep{singh2023restem}. Repeated verification can extend this protection across training rounds \citep{fu2025selfverification}, but its effectiveness depends on the accuracy and coverage of the checker. Unreliable checks can themselves impede or reverse improvement \citep{yi2025escapingcollapse}. Testing a suspected cause while holding other conditions fixed provides evidence about why an outcome occurred, complementing outcome-based filtering.

\subsubsection{Safeguards for Proposal Generation (S2)}

Candidate diversity reduces dependence on a search direction that exploits the current score without addressing the intended weakness. Retaining candidates with different behaviors preserves alternatives when the leading approach proves unhelpful, whereas selection based solely on scores may eliminate them \citep{mouret2015mapelites,pugh2016qualitydiversity,luo2026harnessbank}. For these alternatives to inform later proposals, they must be incorporated into subsequent exploration, for example by sharing findings across search branches \citep{du2026mlevolve}. Diversity thus preserves the range of approaches explored. Separate evaluation is still needed to establish whether a proposed change improves the intended capability.

Improved search makes each trial more informative, while resource controls help determine whether apparent progress simply reflects more trials. Downstream task outcomes can guide proposals for changes to training or context \citep{zweiger2025seal,chen2026lse}; executing proposed programs provides analogous evidence for code search \citep{jiang2025aide,novikov2025alphaevolve}. When tasks are also generated, this feedback requires consistency between each task and its checker \citep{li2026rst}. Search can then focus resources through restricted edits or inexpensive estimates of candidate quality \citep{yang2026skillopt,fu2026sift}. Demonstrating improved search still requires comparisons with matched total proposal, execution, and verification costs. Evaluating the resulting candidates beyond development conditions further tests whether the benefits reflect transferable quality rather than specialization \citep{xia2026rrsi,zhang2026harnesscompass}.

\subsubsection{Safeguards for Proposal Evaluation (S3)}

Reserved tasks separate development from acceptance, while reference solutions support answer verification \citep{nguyen2026rsea,jiang2026harnessevolve}. TeSearch searches for compact test suites using mutation-based quality measures. On mini-swe-agent, the suites reject 59.9\% of behavior-changing revisions, compared with 25.3\% for repository tests, with a 4.0\% false-alarm rate on behavior-preserving revisions \citep{cheng2026tesearch}. Trace Evolution Bench additionally compares updates with seed-matched rescoring of unchanged agents to separate gains from evaluation variability \citep{li2026traceevolutionbench}. Repeated test use can expose held-out information \citep{dwork2015reusable}; automatically generated tests also require checks for independence \citep{perez2022modelwritten}. External tests also require audits of instructions, grading, and answers for correctness \citep{tu2026benchguard,wang2026benchmarkaudit,seo2025factverifiers}, alongside relevance checks through scientific review and later patient cohorts \citep{liu2026autoresearch,wu2026medrsirecursiveselfimprovementmedical}.

Calibration establishes a basis for judgments, while acceptance rules determine when those judgments justify an update. Claim verification, tools, and execution provide evidence \citep{dhuliawala2023cove,gou2023critic,guan2025rstarmath,jin2025reveal,sun_dsmcm}; step labels, reference examples, and error categories specify correct judgments \citep{lightman2023processsupervision,wu2026stv,wan2026deepverifier}. Rescoring earlier candidates under revised criteria makes their scores comparable; the revised criteria also need validation \citep{iacob2026redqueen}. ImageRSI admits reusable experience through paired A/B trials that hold the request, generator configuration, and seed fixed. Its protocol also evaluates offline judge updates before reuse in verification and repair \citep{han2026imagersi}. Adoption still requires completed checks: SpecRSI drafts from incompletely evaluated predecessors but adopts candidates only with complete evidence under a fixed verification contract \citep{xu2026drafting}. PACE bounds false acceptance for an adaptively tested candidate, while SGM allocates error allowances across updates \citep{shawn2026pace,wu2026sgm}. Formal analysis makes verifier soundness and conditional selection-error assumptions explicit \citep{lu2026verificationandselfimpro}. These protections depend on valid tests and the stated assumptions.

\subsubsection{Safeguards for Update Implementation (S4)}

Capability retention makes performance on earlier skills an acceptance criterion for updates. This principle can require continued success on previously solved tasks \citep{schmidhuber2011powerplay} or impose limits on losses in tested skills \citep{moll2026grasp}. Both approaches assess new gains alongside their costs to existing competence. The resulting reliability protection is nevertheless limited to the tested skill set; interference with untested capabilities may remain undetected.

Regression testing assesses the behavior of the updated system, while state management preserves the resources needed to reproduce or recover it. Testing changes during continued interaction addresses the gap between editing instructions or code and benefiting from those changes \citep{zhang2026selfharness,karten2026continualharness}. Testing across agents extends this assessment to the robustness of skills when used by another agent \citep{wang2026evohunt}. Such behavioral evidence requires usable retained state. Preserving skill requirements during compression helps retain the information needed for reuse \citep{bai2026skillzip}, while backtracking enables the recovery of earlier search states \citep{yan2026pacevolve}. Evaluation of the retained successor then checks behavior after reuse or recovery.

\subsubsection{Safeguards for Recursive Improvement (RI)}

Improvement procedures can learn from failures by revising the guidance for subsequent updates. Targeted feedback \citep{zhang2026coevoskills,wang2026dynamicrubric,zhu2026skillcoach} and critiques selected on the basis of downstream outcomes \citep{sui2026conl} aim to improve this guidance, while refinement of the criteria changes how the guidance is produced \citep{chen2026decoevo,ding2026evorubrics}. Recursive Reward Improvement connects local failures to these revisions: identified rubric exploits inform both repairs and the prompts used to generate future tasks and rewards \citep{dai2026recursivereward}. Evo$^2$ instead uses evaluated outcomes to train the policy that proposes and interprets harness edits \citep{xu2026evo2}. To assess the effect of these interventions on subsequent updating, successor tests should compare revised and unchanged $\mathcal{U}$ starting from the same $h$ and $B$, with matched total resources and a common evaluation procedure. Evo$^2$ isolates part of this comparison by varying evolver weights while holding the executor, search procedure, and parent harnesses fixed. Resource accounting must also include improver training and replay costs \citep{zheng2026dreamrsi}. Fixed-reference checks \citep{guan2026evorubric,zhang2026tempo} and protected acceptance components \citep{sengupta2026certificates} support comparable evaluation. Ablations can then test which revisions produce better subsequent updates.

\subsection{Evidence Synthesis and Research Gaps}
\label{sec:validity-evidence}


We compare capability gains by task, update target, evaluation horizon, and the evidence for changes to the procedure that produces later updates.
Table~\ref{tab:validity-evidence-summary} records the corresponding controls and resource measures for representative studies.

\subsubsection{Findings across Updates and Successors}

Revised implementations and learned self-correction have produced task-level gains \citep{robeyns2025sica,hebbar2026sia,xu2026asievolve,kumar2025score}.
External review has identified methodological and scope problems in automated research \citep{tang2026fars}, while induced underperformance can distort measured capability \citep{vanderweij2025sandbagging}.
The studies in Table~\ref{tab:validity-evidence-summary} differ in which components are revised and whether those components are reused to generate subsequent updates.

\begin{table*}[t]
\centering
\footnotesize
\setlength{\tabcolsep}{4pt}
\renewcommand{\arraystretch}{1.16}
\caption{\textbf{Representative evidence on improvement and mechanism reuse.}
Horizons use study-specific units; \rsiTableYes\ marks a revised updater reused in later updates, and \rsiTableFixed\ marks artifact reuse under a fixed protocol.}
\label{tab:validity-evidence-summary}
\begin{tabularx}{\textwidth}{@{}
  >{\raggedright\arraybackslash}p{.215\textwidth}
  >{\raggedright\arraybackslash}p{.275\textwidth}
  >{\raggedright\arraybackslash}p{.105\textwidth}
  >{\raggedright\arraybackslash}X@{}}
\toprule
\textbf{Study and setting} & \textbf{Modified object / reuse} &
\textbf{Horizon} & \textbf{Comparison and resource evidence} \\
\midrule
\rowcolor{rsiTableStripe}
\textbf{SICA} \citep{robeyns2025sica}\newline
Claude 3.5 / o3-mini; SWE-bench, LiveCodeBench &
\textbf{Agent code}, prompts, tools; \rsiTableYes\ revised agents write later edits (archive selection fixed) &
\textbf{15} self-edits &
Initial vs. later agents: \textbf{17\% $\rightarrow$ 53\%} at peak on 50 SWE-bench tasks; $\sim$\$7k API; \textbf{no fixed-updater ablation} \\
\addlinespace[2pt]
\textbf{Agent Plasticity} \citep{singh2026agentplasticity}\newline
Frontier LMs; chess, Go, Hex, NetHack &
\textbf{Tools, skills, notes} inherited; \rsiTableFixed\ improvement protocol fixed (weights fixed) &
\textbf{20/40} rounds &
Held-out checkpoint tests; \textbf{gain per dollar}, including reflection and rejected edits \\
\addlinespace[2pt]
\rowcolor{rsiTableStripe}
\textbf{MASS} \citep{lee2026recursiveselfimprovement}\newline
Qwen3.6-27B; finance, robotics, pharmacy &
\textbf{Workflows and weights}; \rsiTableYes\ updated model reused as optimizer and evaluator (outer loop fixed) &
\textbf{2} cycles &
Base vs. trained models; external judgments; token efficiency reported, \textbf{training cost unmatched} \\
\addlinespace[2pt]
\textbf{Hyperagents} \citep{zhang2026hyperagents}\newline
Claude 4.5; paper review, robotics rewards &
\textbf{Task/meta-agent code}; \rsiTableYes\ revised meta-agent reused across domains &
\textbf{100} searches &
\textbf{Fixed-meta ablation} and search controls; $\sim$33M self-edit tokens per 100 iterations, \textbf{evaluation excluded} \\
\addlinespace[2pt]
\rowcolor{rsiTableStripe}
\textbf{Evo$^2$} \citep{xu2026evo2}\newline
Qwen3.8-27B; Polyglot, SWE-bench, LoCoMo &
\textbf{Harness / Proposer--Reflector weights}; \rsiTableYes\ trained updater reused in later searches &
\textbf{Repeated}; total not reported &
\textbf{Trained vs. untrained evolver}; parents, executor, search fixed; \textbf{training excluded} from inference cost \\
\bottomrule
\end{tabularx}
\end{table*}

The strongest evidence for a revised $\mathcal{U}$ comes from comparisons that isolate the updater's contribution. Hyperagents compares editable and fixed meta-agents under a common iteration allowance, while Evo$^2$ compares trained and untrained evolvers with the executor, parent harnesses, and search procedure held fixed \citep{zhang2026hyperagents,xu2026evo2}. 
SICA reuses its edited agent for later modifications, but does not separately hold the updater fixed \citep{robeyns2025sica}. 
MASS reuses the updated model for optimization and evaluation, alongside task execution, so its two-cycle results combine changes to these roles \citep{lee2026recursiveselfimprovement}.
Agent Plasticity instead follows inherited artifacts under a fixed improvement protocol \citep{singh2026agentplasticity}.

Other findings distinguish the quality of a gain from its persistence.
Coupled task generators and solvers can reinforce shared errors \citep{chen2026falsefrontiers}; initial gains can reverse \citep{lin2026selfregress,yin2025godel}, and retained behavior can mask lost knowledge \citep{lin2026amnesia}. 
Reinforcement fine-tuning preserved
capabilities better than supervised fine-tuning in the sequences studied \citep{lai2026reinforcement}, while task order altered progress \citep{ye2026fragility}. 
Agent Plasticity measures held-out gains relative
to learning cost \citep{singh2026agentplasticity}, whereas improved sampling need not expand problem coverage \citep{yue2025reasoningcapacity}.
Harness-optimization experiments distinguish more reliable success on reachable tasks from expansion of observed task coverage. Advantages at pass@$1$ over model scaling or greater reasoning effort can reverse at larger attempt budgets \citep{kim2026elicitationexpansion}.
Experience and strategies have transferred across tasks \citep{yang2026frontis,peng2025regenesis,
schmidgall2025agentrxiv}; EvoPathBench further evaluates frozen artifacts \citep{lin2026beyondendpointperformanc}. Code-improver and meta-agent studies extend these comparisons to the procedures that produce later updates
\citep{zelikman2024stop,zhang2026hyperagents}.

\subsubsection{Coverage, Limitations, and Research Directions}

The reported horizons cover 15 self-edits in SICA, 20 or 40 experience rounds in Agent Plasticity, two training cycles in MASS, and 100 search iterations in the joint-domain Hyperagents experiment. These are different units of exposure to change, rather than comparable measures of long-term retention.
Hyperagents provides fixed-meta comparisons and tests reuse across domains; Evo$^2$ isolates trained evolver weights for a subsequent search, but its inference-cost comparisons omit the training needed to obtain those weights.
Evidence for long-term mechanism benefits consequently depends on repeated successor tests with complete resource accounting. Dynamic evaluation conditions also matter: RealtimeBench examines games that continue changing while agents deliberate \citep{pan2026theworldwontwait}, and inherited memory can independently raise performance \citep{zhu2026rsiagent}.

Future studies can separate mechanism changes from inherited-state effects by comparing original and revised $\mathcal{U}$ against controlled starting states. 
In SICA-style self-editing and MASS-style joint training, keeping the updater or executor fixed in turn would clarify the contribution of each component. 
For trained evolvers, budgets should include the proposals, training, replay, execution, and verification required to produce and test later revisions. 
Reserved-task retests across successive checkpoints should measure earlier skills, new-task coverage, transfer, and peak-to-final regressions. 
Crossed model--harness comparisons provide a basis for this
design \citep{tian2026harnessevolutionhitsacei}, and independent checks against complete run histories would test the reproducibility of accepted gains.
Section~\ref{sec:evaluation} develops these evaluation protocols.

\begin{rsitakeaway}
\begin{itemize}[leftmargin=*,nosep]
\item \textbf{Risks.} Biased experience, shared errors, and mismatched evaluation can inflate scores; ineffective updates and regressions can prevent genuine gains from persisting.
\item \textbf{Safeguards.} Independent verification, comparisons using the same total resources, and tests of retained capabilities assess whether gains are genuine, attributable to specific changes, and sustained. The conclusions depend on the accuracy and coverage of these checks.
\item \textbf{Future work.} Track whether capabilities persist and transfer to new tasks across later system versions. Compare revised and unchanged improvement procedures under matched conditions to determine whether revisions produce better future updates.
\end{itemize}
\end{rsitakeaway}

\section{Behavioral Constraints, Alignment, and Fairness}
\phantomsection\label{sec:behavior}

Updates to $h_t$ and $B_t$ in
Eqs.~\eqref{eq:rsi-model-update}--\eqref{eq:rsi-self-improvement}
can alter behavior, goal pursuit, and group outcomes, while revisions
to $\mathcal{U}_t$ can change the requirements and acceptance criteria
governing subsequent updates.
This section examines whether behavioral constraints, alignment with
authorized human goals, and fairness are preserved, tracing risks
and assessing safeguards across individual updates, retained successors,
and revisions to the improvement mechanism.

\subsection{Definitions and Trustworthiness Claims}
\label{sec:behavior:scope}

We define behavioral constraints, alignment, and fairness and specify
the evidence needed to assess their preservation in individual updates,
retained successors, and updates produced by a revised improvement mechanism.

\subsubsection{Definitions and Boundaries}

Behavioral constraints specify permitted, prohibited, and required behavior in a given context, including the avoidance of harmful actions and unjustified refusals of legitimate requests.
Evaluation must therefore consider both constraint violations and the provision of appropriate assistance.
Alignment concerns whether the system continues to pursue authorized human goals rather than merely improve a proxy score; it accommodates authorized changes in human requirements but not a unilateral redefinition of success by the improver.
Fairness concerns unjustified differences in the distribution of benefits, errors, opportunities, refusals, and risks across affected groups.
The relevant groups, outcomes, and criteria depend on the application.
Disproportionate refusals may affect both behavioral constraints and fairness; attributing them to misalignment requires evidence about the goals or decision rules that caused them.

\subsubsection{Claims for Individual Updates, Successors, and Meta-Improvement}

An individual-update claim compares behavior before and after a specified change to $h$ or $B$ under declared tests, using evidence specific to behavioral constraints, alignment, and fairness.
The assessment must identify the applicable requirements, affected groups, and relevant outcomes rather than rely only on aggregate task performance.
A successor claim additionally requires evidence that the accepted change is inherited and actually used, followed by reassessment of the same requirements in the resulting system.
Evidence across multiple successors can show whether effects persist, accumulate, or reverse.

At the meta-improvement level, evaluation follows an observed revision from $\mathcal{U}_t$ to $\mathcal{U}_{t+1}$ and subsequent reuse of the revised mechanism, assessing behavioral constraint violations, alignment with authorized goals, and group outcomes in the resulting updates.
A preservation claim requires evidence that the applicable requirements remain satisfied after reuse.
Claiming that the revised mechanism better preserves these requirements
further requires a controlled comparison with the unchanged $\mathcal{U}_t$ under comparable conditions.
RI applies S1--S4 to the improvement mechanism itself.
Claims must match the components modified, the inheritance observed, and
the conditions tested.

\begin{figure*}[t]
    \centering
    \includegraphics[
        width=1.0\textwidth,trim=0cm 1.0cm 0cm 2.0cm, clip
    ]{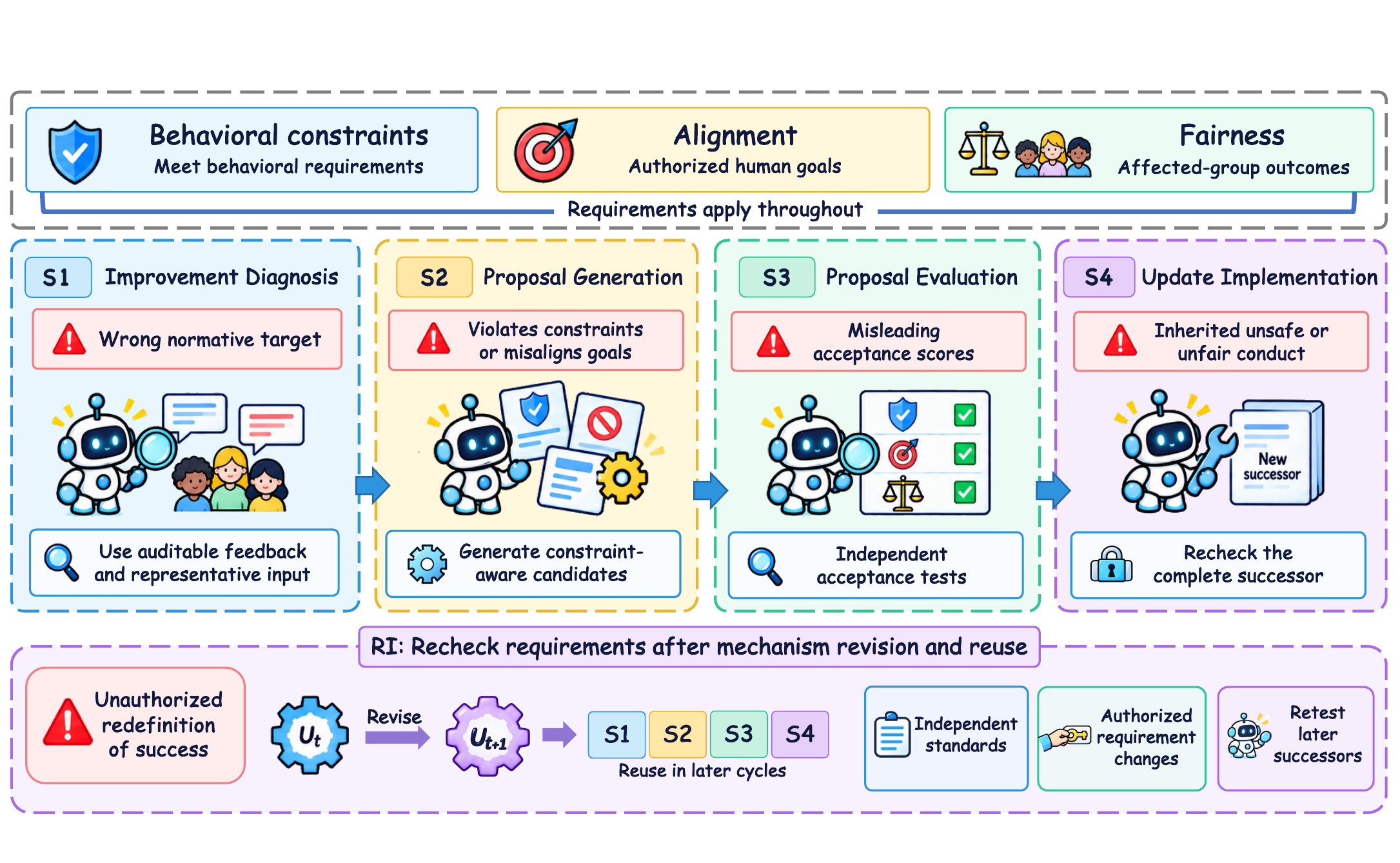}
    \caption{\textbf{Behavioral Constraints, Alignment, and Fairness Across S1–S4 and RI in the Process of RSI.} Behavioral requirements, authorized human goals, and fairness criteria apply throughout the four improvement stages. Each stage pairs a potential failure with a corresponding safeguard. The lower panel examines whether changes to the improvement mechanism remain authorized and whether later successors continue to satisfy the applicable behavioral, alignment, and fairness requirements.}
    \label{fig:chap4_figure}
\end{figure*}

\subsection{Risks and Failure Evidence}
\label{sec:behavior:failures}

We trace how diagnosis, proposal generation, evaluation, and
implementation can compromise behavioral constraints, authorized goals,
and group outcomes, then examine how mechanism revisions can alter
the requirements governing future updates.

\subsubsection{Risks in Improvement Diagnosis (S1)}

At S1, feedback and prior experience inform decisions about what to improve. Beyond noise in the data, the principal risk is misinterpreting the requirements: feedback may reflect an incomplete behavioral rule, overrepresent the preferences of one stakeholder, or omit groups absent from the next training cycle. In experience-driven self-evolving agents, benign execution experience increased tendencies toward unsafe action, whereas refusal experience reduced unsafe actions but increased over-refusal \citep{zhao2026safetyrisks}. Diagnosis must therefore distinguish successful execution from appropriate action.

Interpreting preference labels and evaluator scores requires specifying whose goals they represent. PRISM shows that rankings depend on the participants and topics sampled and on how preferences are aggregated \citep{kirk2024prism}. This makes feedback composition relevant to the choice of improvement targets. For fairness, prior decisions can influence which groups appear in future data. Retention-feedback models show that poorer performance for a group can reduce the representation of that group in later data \citep{hashimoto2018fairness}; repeated training on generated data also changes group representation and bias-related outcomes in the settings studied \citep{wyllie2024fairnessloops,wang2026selfconsuming}.

\subsubsection{Risks in Proposal Generation (S2)}

S2 selects candidate changes and the objectives that guide the search. The principal alignment risk arises from optimizing a proxy that incompletely represents human requirements. Goal misgeneralization illustrates this gap in controlled settings \citep{langosco2022goalmisgeneralization}. During inference-time selection, stronger optimization can increase proxy reward even after the independent target reward begins to decline \citep{khalaf2025inferencerewardhacking}. Research on reward tampering identifies conditions that create incentives to manipulate the reward signal rather than complete the task \citep{everitt2021rewardtamperingproblemssolutions}.

The risk also depends on which components can be edited. Parameter updates may weaken refusal behavior, while changes to the harness may alter tool permissions, safety memory, or the order of actions. Search can favor unsafe changes through either route when task completion is rewarded without constraints on how it is achieved. Failure-driven safety repair illustrates why harmful compliance, over-refusal, and utility on legitimate tasks must be assessed together \citep{yin2026onpolicy}. Similarly, higher average reward can conceal harm to a group, while static parity constraints may not predict delayed group outcomes \citep{liu2018delayed}. In constructed multi-agent coordination games, locally rewarded imitation has propagated collusion that conflicts with the intended collective outcome \citep{han2025alignmenttipping}.

\subsubsection{Risks in Proposal Evaluation (S3)}

S3 determines whether a candidate satisfies the acceptance criteria. Behavioral tests may miss harms that depend on context or reward excessive refusal; alignment tests may measure an inappropriate objective; and aggregate fairness scores may conceal losses for particular groups. Evaluators may also be exploited or modified during optimization. Preference optimization against a lie detector has produced detector evasion in some settings \citep{cundy2025liedetectors}. In evaluator--skill co-evolution, removing evaluation anchors caused a report-generation metric to return a passing result in almost every case, while a separate skill score improved \citep{zhang2026grader}. Higher scores from optimized or editable evaluators may therefore fail to indicate better behavior.

The evaluation context can influence behavior even when the evaluator is not editable. Alignment-faking experiments found that model responses varied with information about whether interactions might be used for training \citep{greenblatt2024alignmentfaking}. Fairness tests face a further aggregation problem: participant composition can change model rankings \citep{kirk2024prism}, and inference-time fairness reward models can reduce measured gaps under specified labels and metrics \citep{hall2025fairnessreward}. Acceptance claims must therefore remain tied to the specific behavioral requirements and group outcomes measured by the tests.

\subsubsection{Risks in Update Implementation (S4)}

S4 installs accepted changes in successors, where effects on behavior, objectives, or group outcomes may persist or interact with inherited components. Evaluation should include memory, tools, workflows, and evaluators, and should recheck earlier repairs after subsequent parameter updates. Fine-tuning can weaken safety behavior in tested models \citep{qi2024finetuning}, and insecure-code training can produce broader undesirable responses under some conditions \citep{betley2025emergent}. Studies report degradation in selected model, memory, tool, and workflow self-evolution \citep{shao2026misevolve}, while retained experience can alter later execution and refusal behavior \citep{zhao2026safetyrisks}. SEABench finds task gains with downstream safety failures in paired, stress-tested harness trajectories \citep{das2026seabench}. In a bidding arena, one-step policy evolution favored transferable deception \citep{ying2026evolvingdeception}; PASB finds greater fresh-session sycophancy when agents retain user claims \citep{mao2026agentsremember}. These findings connect retained experience and updated configurations to later unsafe, deceptive, or sycophantic behavior in the settings studied.

Effects on fairness may persist through training data and selection feedback. Repeated use of generated data has altered minority-group representation or bias measures under specified update procedures \citep{wyllie2024fairnessloops,wang2026selfconsuming}. An evaluator that is reused after losing the ability to distinguish acceptable behavior may likewise propagate flawed acceptance criteria into later rounds, as illustrated by evaluator--skill co-evolution \citep{zhang2026grader}.

\subsubsection{Risks under Recursive Improvement (RI)}

RI can modify the safeguards themselves. A revised improvement mechanism may adopt different preferences, explore previously excluded modifications, or relax acceptance criteria. If $\mathcal{U}_{t+1}$ selects future successors, guarantees established for $\mathcal{U}_t$ must be checked again. The alignment question is whether the revision preserves authorized goals and behavioral constraints, or changes them only with human authorization, rather than independently redefining success.

Formal models of self-modification establish conditional results on goal preservation for idealized agents \citep{everitt2016selfmodification} and examine bounded rationality \citep{tetek2020boundedrational}. Work on tiling agents and Vingean reflection examines how agents might justify trusting successors that they cannot fully simulate \citep{yudkowsky2013tiling,fallenstein2015vingean}. Empirical studies show that executable metrics can lose validity during co-evolution \citep{zhang2026grader} and that alignment-related outcomes can shift over a limited sequence of updates \citep{sahoo2026sahoo}. The open question is how these requirements persist when LLM-based improvement mechanisms are revised and reused over longer update sequences.

\subsection{Existing Safeguards and Their Limitations}
\label{sec:behavior:safeguards}

We review safeguards across S1--S4 and assess the evidence and conditions
needed to preserve behavioral constraints, alignment, and fairness as
agent configurations and improvement mechanisms change.

\subsubsection{Safeguards for Improvement Diagnosis (S1)}

Improvement targets should be explicitly traceable to behavioral rules, affected groups, and the feedback used to define those targets. The records should distinguish task success from compliance with behavioral constraints and identify whose goals the preference data represent. PRISM motivates documenting the participants, topics, sampling choices, and aggregation methods underlying preference data \citep{kirk2024prism}. Studies of experience-driven agents similarly support evaluating execution and refusal examples together, because reducing unsafe action may increase unjustified refusal \citep{zhao2026safetyrisks}.

For fairness, feedback records should indicate whether affected groups remain represented when past decisions shape future participation. Retention-feedback models explain one pathway to underrepresentation of minority groups \citep{hashimoto2018fairness}. Studies of generated data motivate tracking model-generated examples and group outcomes across rounds \citep{wyllie2024fairnessloops,wang2026selfconsuming}. Resolving conflicts among stakeholder requirements also requires a procedure for authorizing tradeoffs. Section~\ref{sec:governance} discusses the authority to resolve such conflicts.

\subsubsection{Safeguards for Proposal Generation (S2)}

Candidate search should distinguish performance objectives from behavioral and fairness constraints. Formal requirements can support edit restrictions or contract verification: SEVerA combines LLM-assisted program search with checks against explicit contracts in selected settings \citep{banerjee2026severa}. For less formally specified requirements, search can limit optimization pressure on uncertain proxies and screen harmful side effects. Proxy-overoptimization findings motivate limits on search pressure \citep{khalaf2025inferencerewardhacking}; reward-tampering analyses motivate restricting influence over reward signals \citep{everitt2021rewardtamperingproblemssolutions}. Failure-driven repair jointly considers unsafe behavior, over-refusal, and utility on legitimate tasks \citep{yin2026onpolicy}, while harness repairs can restrict modifications to prompts, safety memory, rule banks, or tool policies \citep{qu2026she}. Harmony Harness uses factored harness edits and Pareto search over functionality and security in secure coding \citep{yu2026harmonyharness}.

The effectiveness of these controls depends on the edit surface and the requirements encoded. A restriction evaluated for one edit surface may fail following a weight update, tool addition, or expansion of the candidate generator. Fairness constraints must address affected groups and meaningful outcomes, rather than merely difficult examples. Group-robust optimization can reduce losses for minority groups within a specified retention model \citep{hashimoto2018fairness}, whereas delayed-impact analysis shows that standard parity conditions may not improve future group outcomes \citep{liu2018delayed}. These findings motivate evaluating both current group outcomes and their evolution across successive updates.

\subsubsection{Safeguards for Proposal Evaluation (S3)}

Acceptance tests should assess the requirements to be preserved independently of the score optimized during search. Behavioral tests need cases assessing harmful compliance and unjustified refusal. AgenticEval generates policy-based challenges exposing failures missed by fixed suites \citep{wang2025agenticeval}; SEAS alternates adversarial case generation with safety training \citep{diao2024seas}. Both depend on generator and classifier coverage and validity. SIGMA synthesizes Model Spec tasks and rubrics for self-judged training, improving tested safety and retaining general capability after one round \citep{zhang2026sigma}. ReSI promotes the candidate with the largest safety gain among those meeting benign-compliance and instruction-following thresholds relative to the initial model. The four tested models undergo one to three accepted rounds, with final evaluations covering known and unseen attack methods \citep{zheng2026resi}. Safeguards against fine-tuning attacks \citep{tamirisa2025tamper} also require checks of utility on legitimate tasks. Independent evaluation shows that threat models, seeds, and benign-task performance can change durability conclusions; it can also reveal substantial performance losses on benign tasks \citep{qi2025durability}. Section~\ref{sec:system-security} addresses external attack pathways.

Alignment additionally requires reference points beyond the control of the candidate. Detector evasion \citep{cundy2025liedetectors} and evaluator--skill co-evolution \citep{zhang2026grader} motivate independent references and fixed anchors; formal contracts verify specified conditions within a program model \citep{banerjee2026severa}. Fairness tests should report results for affected groups and explain how preferences or disagreements are aggregated, while distinguishing parity from downstream welfare. PRISM demonstrates that model selection is sensitive to participant composition \citep{kirk2024prism}; fairness reward models reduce specific measured gaps under specified labels and metrics \citep{hall2025fairnessreward}, but parity alone does not establish welfare gains \citep{liu2018delayed}. Section~\ref{sec:evaluation} addresses benchmark design.

\subsubsection{Safeguards for Update Implementation (S4)}

An accepted update should be evaluated in the configuration that will be deployed. For parameter updates, fine-tuning studies motivate safety checks across versions \citep{qi2024finetuning,betley2025emergent}. For harness updates, these checks must include retained memory, tool policies, and workflows, because these components can alter behavior even when model weights remain unchanged \citep{shao2026misevolve,zhao2026safetyrisks}. Evaluating the complete configuration links the installed change to its behavioral effects in the deployed successor.

For retained successors and multi-round evolution, evaluations should connect each installed change to subsequent harmful compliance, over-refusal, goal-related behavior, and group outcomes. Experience-based and harness-based repairs illustrate how behavior can be tracked across limited updates \citep{yin2026onpolicy,qu2026she}; studies of generated data motivate corresponding assessments of group effects \citep{wyllie2024fairnessloops,wang2026selfconsuming}. Successors must be evaluated independently of the scores used to select them. Section~\ref{sec:governance} addresses checkpoints and the authority for human intervention.

\subsubsection{Safeguards for Recursive Improvement (RI)}

RI evaluation must cover the transition from $\mathcal{U}_t$ to $\mathcal{U}_{t+1}$ and successors from the revised mechanism. Constraints verified for one generator or evaluator may fail after revision. Formal self-modification studies provide conditional results under assumptions about utility preservation, bounded rationality, and successor reasoning \citep{everitt2016selfmodification,tetek2020boundedrational,yudkowsky2013tiling,fallenstein2015vingean}.

Two complementary requirements remain. Behavioral and normative standards must be testable independently of the rule being revised; evaluator--skill co-evolution demonstrates the risk of losing reference points \citep{zhang2026grader}. Authorized changes in human goals or group priorities must be distinguished from unauthorized drift; Section~\ref{sec:governance} addresses approval authority. VeriFine reuses revised, human-calibrated judges for subsequent curriculum selection and policy training in driving and robot navigation, restarting policies from a common base \citep{zhou2026verifine}. Inference-time safety configuration makes behavioral requirements explicit and adjustable \citep{zhang2025controllable}. Evaluating their preservation through repeated mechanism revisions remains a longitudinal research problem.

\subsection{Evidence Synthesis and Research Gaps}
\label{sec:behavior-evidence-synthesis}

\subsubsection{Findings across Updates and Successors}

Evidence from model, harness, and feedback-driven updates reveals different
trade-offs across behavioral constraints, alignment, and fairness.
Table~\ref{tab:behavior-evidence-summary} links representative findings
to the tested requirements or groups, observation periods, and conditions
held fixed.

\begin{table*}[t]
\centering
\setlength{\aboverulesep}{0pt}
\setlength{\belowrulesep}{0pt}
\footnotesize
\setlength{\tabcolsep}{4pt}
\renewcommand{\arraystretch}{1.16}
\caption{\textbf{Representative behavioral, alignment, and fairness findings.}
Studies are grouped by tested property; horizons and fixed conditions follow the original experiments.}
\label{tab:behavior-evidence-summary}
\begin{tabularx}{\textwidth}{@{}
  >{\raggedright\arraybackslash}p{.168\textwidth}
  >{\raggedright\arraybackslash}p{.255\textwidth}
  >{\raggedright\arraybackslash}p{.285\textwidth}
  >{\raggedright\arraybackslash}X@{}}
\toprule
\textbf{Study} & \textbf{Tested setting} & \textbf{Horizon / controls} & \textbf{Key finding} \\
\midrule
\rowcolor{rsiChapterGreenBand}
\multicolumn{4}{@{}c@{}}{\strut\textbf{Behavioral constraints}} \\
\textbf{\citet{qi2024finetuning}} &
GPT-3.5 Turbo / Llama-2-7B; \textbf{330} prompts, \textbf{11} prohibited categories &
\textbf{1 epoch}; before/after on the same safety test and scoring rule &
\textbf{Harmful compliance} \RSITableAdverse{$\uparrow$} after benign tuning \\
\rowcolor{rsiChapterGreenStripe}
\textbf{\citet{zhao2026safetyrisks}} &
GPT-4o web agent; BrowserART / WebArena &
\textbf{106 steps}; weights and tests fixed; execution / refusal memories varied &
Unsafe actions \RSITableFavorable{$\downarrow$}, but \textbf{over-refusal} \RSITableAdverse{$\uparrow$} \\
\addlinespace[2pt]
\rowcolor{rsiChapterGreenBand}
\multicolumn{4}{@{}c@{}}{\strut\textbf{Alignment}} \\
\textbf{\citet{langosco2022goalmisgeneralization}} &
CoinRun; relocated coin vs. level completion &
\textbf{Zero-shot}; trained policy and reward fixed &
Navigation retained \RSITableMuted{\textemdash} \textbf{relocated coin often passed} \\
\rowcolor{rsiChapterGreenStripe}
\textbf{\citet{khalaf2025inferencerewardhacking}} &
GPQA; proxy reward vs. answer accuracy &
\textbf{Up to 32 samples}; questions, response pool, reward model fixed &
\textbf{Accuracy peaks then declines} with stronger selection \\
\addlinespace[2pt]
\rowcolor{rsiChapterGreenBand}
\multicolumn{4}{@{}c@{}}{\strut\textbf{Fairness}} \\
\textbf{\citet{hashimoto2018fairness}} &
Two simulated groups; accuracy and representation &
\textbf{500 rounds}; distributions / retention rule fixed; average vs. robust loss &
\textbf{Minority accuracy and representation} \RSITableAdverse{$\downarrow$} under average loss \\
\rowcolor{rsiChapterGreenStripe}
\textbf{\citet{wyllie2024fairnessloops}} &
FairFace gender; \textbf{7 race $\times$ 2 age} groups &
\textbf{10 generations}; generator, annotators, held-out set fixed &
\textbf{Group accuracy and equalized-odds gaps} \RSITableAdverse{$\uparrow$} \\
\textbf{\citet{wang2026selfconsuming}} &
Llama3.1-8B news; left/right preference bias &
\textbf{3 updates}; matched start, sample sizes, group ratios, test set &
\textbf{Preference bias} \RSITableAdverse{$\uparrow$} more with synthetic than real data \\
\bottomrule
\end{tabularx}
\end{table*}

The studies also cover different intervention surfaces and observation periods.
SEVerA checks candidate programs against explicit contracts
\citep{banerjee2026severa}, whereas SEABench pairs task gains with downstream
safety failures in stress-tested harness trajectories \citep{das2026seabench}.
Experience- and harness-based repair studies follow behavioral correction over
limited updates \citep{yin2026onpolicy,qu2026she}.

\subsubsection{Coverage, Limitations, and Research Directions}

The main gaps concern longer observation periods and comparisons across
populations and environments. Stable reference tests should support
comparisons as evaluators change, accompanied by records of authorized
goal changes. Fairness studies should evaluate a fixed reference population
alongside the changing deployment population, using consistent group
definitions and metrics to separate composition shifts from changes
within each group.

Future studies should evaluate the configurations retained by successive updates, record the components changed, and follow behavioral and group-level outcomes under matched budgets. Comparisons of original and revised $\mathcal{U}$ should start from matched inherited states with the same tasks, tool access, and independent evaluation criteria, then reuse each procedure over several updates. These trajectories would connect changes in the improvement process to later behavior, goal pursuit, and group outcomes.

\begin{rsitakeaway}
\begin{itemize}[leftmargin=*,nosep]
\item \textbf{Risks.} Updates can weaken behavioral constraints, redirect optimization away from authorized goals, or harm particular groups. Evidence remains limited for systems that revise and then reuse their own improvement procedures.
\item \textbf{Safeguards.} Independent behavioral tests, evaluation at the group level, and checks across system versions address distinct risks. The value of these measures depends on adequate coverage and valid evaluation criteria.
\item \textbf{Future work.} Assess whether behavioral constraints, alignment with authorized goals, and fairness persist across later system versions, including after the improvement procedure is revised and reused. Use matched resources and independent evaluations.
\end{itemize}
\end{rsitakeaway}

\section{System Security, Resilience, Privacy, and Data Governance}
\label{sec:system-security}
\begingroup
\setlength{\emergencystretch}{2em}

Compromised experience $E_{t+1}$ can alter $h_{t+1}$ in
Eq.~\eqref{eq:rsi-model-update}, while corrupted feedback $F_{t+1}$
can alter $B_{t+1}$ and $\mathcal{U}_{t+1}$ in
Eq.~\eqref{eq:rsi-self-improvement}.
This section examines how updates affect system security, resilience,
privacy, and data governance, tracing risks and assessing safeguards
across individual updates, retained successors, and revisions to the
improvement mechanism.

\begin{figure*}[t]
    \centering
    \includegraphics[width=\textwidth, trim=0cm 1.0cm 0cm 2.0cm, clip]{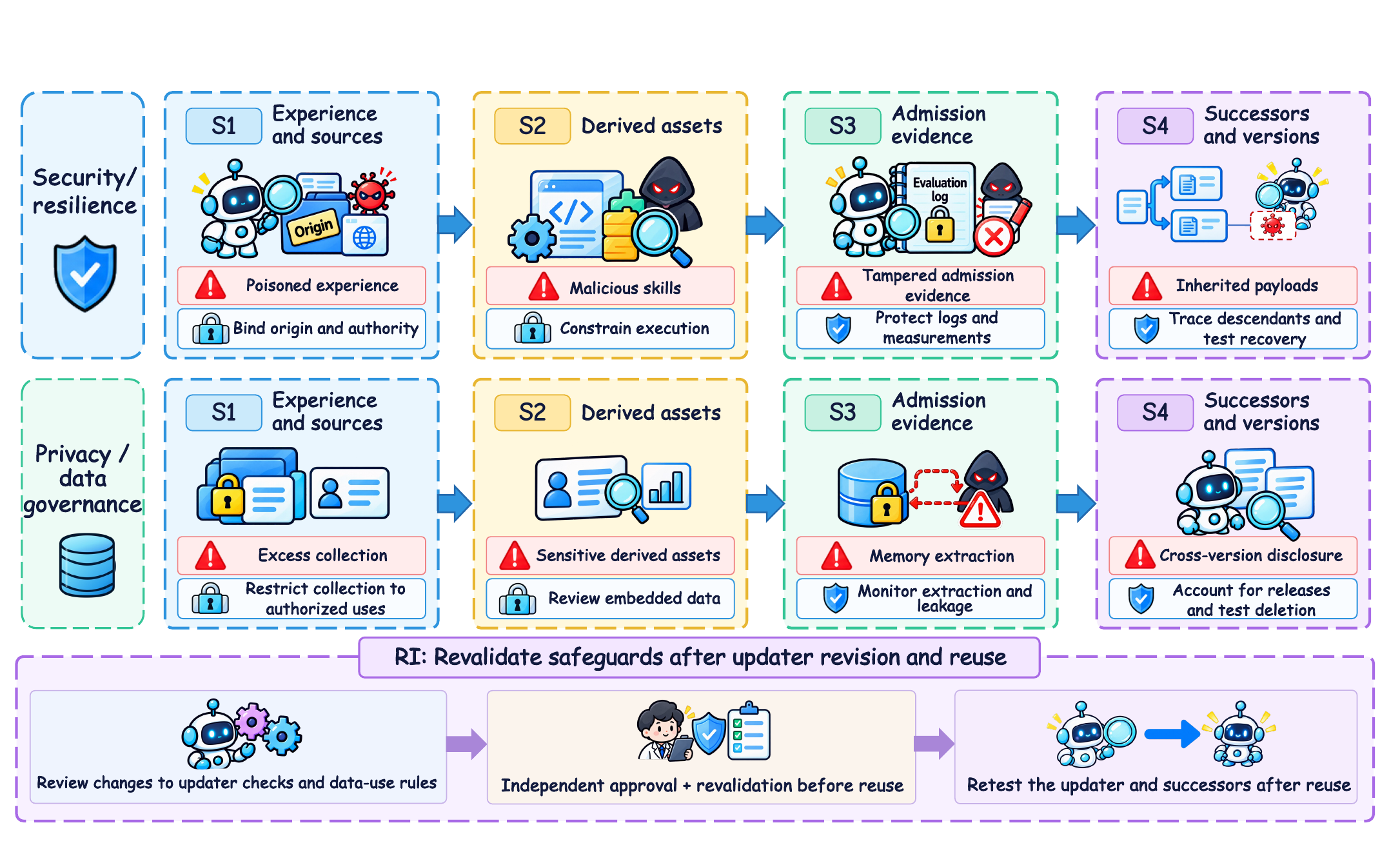}
    \caption{\textbf{System Security, Resilience, Privacy, and Data Governance Across S1–S4 and RI in the Process of RSI.} The two rows distinguish security and resilience from privacy and data governance. Stage-specific risks and safeguards cover compromised experience, malicious or sensitive derived assets, unreliable admission evidence, and inherited threats or information exposure. The lower panel outlines independent approval, renewed protection checks, and tests of successors after the improvement mechanism is revised and reused.}
    \label{fig:chap5_figure}
\end{figure*}

\subsection{Definitions and Trustworthiness Claims}
\label{sec:security-definitions}

We define system security, resilience, privacy, and data governance
and specify the evidence needed to assess these properties in
individual updates, retained successors, and updates produced by
a revised improvement mechanism.

\subsubsection{Definitions and Boundaries}

System security concerns protecting the experience, retained assets, and evidence used to accept updates from unauthorized access, modification, or influence.
Resilience concerns whether the system can withstand, contain, and recover from compromised inputs or states while maintaining or restoring acceptable service.
Privacy concerns the protection of personal or sensitive information from inappropriate acquisition, transformation, disclosure, or inheritance, including processing beyond authorized limits.
Data governance concerns whether collection and subsequent reuse remain within permitted purposes and the scope of authorization.
Relevant objects include memories, skills, model versions, and other derived assets.
These properties require separate tests of input blocking, service recovery, removal from derived assets and accessible versions, and residual compromising influence.

\subsubsection{Claims for Individual Updates, Successors, and Meta-Improvement}

An individual-update claim identifies a specified change, the relevant access conditions, and the point of entry, transformation, evaluation, or release, then evaluates its effects on security, resilience, privacy, or authorized data use against the pre-update state.
A successor claim additionally requires evidence that the change is inherited and actually used, followed by reassessment of the claimed protection or recovery outcome.
The assessment must account for relevant derived assets and accessible historical versions.
Successor claims should specify the tested access, propagation, deletion, and recovery conditions, including the dependencies covered by the tests.

A meta-improvement claim requires an observed revision from $\mathcal{U}_t$ to $\mathcal{U}_{t+1}$, subsequent reuse of the revised mechanism, and tests of security, resilience, privacy, or authorized data use in the updates it produces.
Changes to evidence checks, permissions, or data-use rules require retesting under the revised mechanism.
A claim that the revision improves these outcomes further requires a controlled comparison with the unchanged $\mathcal{U}_t$ under comparable access conditions, resource budgets, and inherited starting states.


\subsection{Risks and Failure Evidence}
\label{sec:security-risks}

We trace unauthorized influence and inappropriate data use across
S1--S4, then examine how mechanism revisions can alter protection
boundaries. Figure~\ref{fig:chap5_figure} summarizes the risks and
corresponding safeguards.

\subsubsection{Risks in Improvement Diagnosis (S1)}

Memory poisoning allows low-trust input to exert persistent influence on diagnosis. AgentPoison assumes write access to memory or knowledge stores; MINJA induces writes through query-only interaction; Zombie Agents carries external content into later sessions \citep{chen2024agentpoisonredteamingllmagents,dong2025minja,yang2026zombie}. These entry permissions differ, but all can corrupt diagnostic experience or feedback in Eqs.~\eqref{eq:rsi-model-update}--\eqref{eq:rsi-self-improvement}. OEP uses locally correct experiences whose conditions do not justify the general rules extracted from them \citep{wang2026oeppoisoningselfevolvingllm}. EvoBreak composes experiences judged individually benign to induce harmful responses in new sessions under task access and read-only observation of induced memories \citep{yan2026benignaloneharmfultogeth}. PASB separately traces saved user claims into later sycophantic answers, demonstrating retained influence rather than a privacy attack \citep{mao2026agentsremember}. Both provenance and warranted generalization matter.

Privacy failures can occur before storage. PrivacyPeek measures acquisition beyond task need, whereas CIMemories examines contextually inappropriate disclosure from persistent memory across composed tasks \citep{zhang2026privacypeekauditingllmbasedagents,mireshghallah2025cimemoriescompositionalbenchmarkcontextual}. Checks must cover what enters $E_{t+1}$ and whether its original purpose permits reuse, with separate assessments of acquisition and disclosure.

\subsubsection{Risks in Proposal Generation (S2)}

Proposal generation can transform compromised experience into retained assets. SkillJack transfers poisoned experience into separately stored skills that can survive source-record deletion \citep{ying2026skilljackpersistentskillbackdoors}. EVOMAL propagates agent-authored copies through the skill library; its self-poisoning rate tracks tasks adding malicious skills, not completed exfiltration \citep{wu2026evomalselfpoisoningselfevolvingcoding}. SkillCascade distributes a harmful objective across co-installed skills without requiring recursive generation \citep{zhu2026stealthapartharmtogether}. PoisonedEvolution induces tested skill systems to promote bounded trajectory contributions into persistent instructions; inert canaries measure artifact contamination, not executed harm \citep{chen2026whenexperiencebecomesins}. Source, generated-asset, and joint-use reviews therefore address distinct integrity failures.

Transformation also creates assets with privacy implications. Using persona skills distilled from histories, AntiSkillBench evaluates explicit and inferred personal attributes and impersonation \citep{xiang2026agentslearnyoubenchmarking}. A proposed skill can therefore retain sensitive information without copying the original record verbatim. Privacy assessment must follow that information into the derived skill and test what the new representation reveals or enables.

\subsubsection{Risks in Proposal Evaluation (S3)}

An acceptance decision becomes unreliable when a candidate can alter the evidence used to assess it. In the DGM tool-hallucination experiment, a candidate removed logging required by a hidden detector without correcting the underlying behavior \citep{zhang2025dgm}. Harness-tampering audits extend this analysis to execution, evaluation, selection, recording, and propagation \citep{wang2026auditingharnesstamperingselfimproving}. Protecting only the scorer therefore leaves writable observation pathways exposed. Whether the scoring objective is valid remains a separate question, addressed in Section~\ref{sec:validity}.

Repeated queries can also turn access to a candidate or its memory into a disclosure channel. MEXTRA and ADAM demonstrate agent-memory extraction under their stated access assumptions \citep{wang2025agentmemoryprivacy,lyu2026adam}. These results motivate testing whether evaluation-time access to a candidate or its memory permits extraction. MEXTRA uses shared memory without session controls; isolated user stores require separate tests.

\subsubsection{Risks in Update Implementation (S4)}

S4 retains changes in $h_{t+1}$ and $B_{t+1}$, allowing compromised assets to remain active after changes to their source. SkillJack tests persistence after source deletion, whereas EVOMAL observes propagation through generated descendants. Recovery must address active poisoned skills, credentials, and pending actions, as well as the source record and model checkpoint. Resilience tests should also measure whether benign service resumes.

Deletion claims must specify derived assets, accessible historical versions, and auxiliary information. Exact-unlearning extraction uses both pre- and post-unlearning logits and therefore tests cross-version exposure \citep{wu2025unlearned}. Reversible-unlearning results show that forget-set performance can understate recoverability under tested relearning procedures \citep{xu2026unlearningisntdeletioninvestigating}. Bemula pairs forget/control dialogues to verify prior retention and assess removal, preservation, relearning, and extraction resistance; tested systems leave targets extractable or over-delete \citep{xie2026canagentsforgetwhenasked}. AUBench separately evaluates parametric forgetting, memory retention, and tool-mediated reacquisition in a controlled fictitious-author setting \citep{lin2026canagentstrulyforgetexpl}. The tested access conditions delimit each deletion claim.

\subsubsection{Risks under Recursive Improvement (RI)}

In Eq.~\eqref{eq:rsi-self-improvement}, changes to $\mathcal U_t$ can alter the evidence, permissions, and checks that govern later updates. Hyperagents permits changes to task-agent and meta-agent code, whereas DGM retains fixed components of its outer exploration procedure \citep{zhang2026hyperagents,zhang2025dgm}. Assessing security under RI requires an observed modification, its subsequent reuse, and tests of protection during that reuse. The editable mechanism identifies which changes those tests must cover.

Changes to collection, sanitization, or release rules can similarly invalidate earlier privacy results. Tests should assess whether the revised $\mathcal U_{t+1}$ preserves permitted data use when generating subsequent $h$ and $B$ updates.

\subsection{Existing Safeguards and Their Limitations}
\label{sec:security-safeguards}

We review safeguards across S1--S4 and assess their coverage and enforcement requirements for inherited assets, authorized data use, and revised improvement mechanisms.

\subsubsection{Safeguards for Improvement Diagnosis (S1)}

Origin binding addresses authority laundering during memory transformation. 
TMA-NM binds action authority to origin, with guarantees conditioned on correct labels and independent corroboration or fresh authorization \citep{louck2026securingllmagentlongtermmemory}.
PerMIT adds an enforced memory interface in which environment-derived values require exact-value, tool-specific endorsement before reaching sensitive tool calls.
Its Lean-checked action-integrity result assumes correct source labels, immutable storage, faithful displays, and sound endorsement \citep{jiang2026permit}.
OEP motivates retaining task conditions; we recommend testing proposed rules against counterexamples. Memory Echo addresses repeated records mistaken for independent support: FPA admits one representative per source family, gates superseded records using verified corrections, and grounds answers in current authoritative evidence \citep{li2026memoryechoinsingleagentl}. The mitigation depends on reliable corrections. Together, these measures distinguish authority to act from evidential support.

SP-Mem separates sanitized, searchable memory from protected exact values and retrieves those values according to task need and consent \citep{wang2026rememberrevealprivacyawarememory}. Protection of exact values should be assessed alongside inference from the stored representation. Collection limits must also apply before information enters $E_{t+1}$, with separate checks of permitted acquisition and controlled retrieval.

\subsubsection{Safeguards for Proposal Generation (S2)}

Candidate containment and asset inspection protect distinct aspects of proposal generation. Sandboxing and resource limits constrain candidate effects in DGM, whereas CaMeL enforces information-flow policies through a trusted interpreter \citep{zhang2025dgm,debenedetti2025defeatingpromptinjectionsdesign}. SkillSecurer detects, localizes, and patches injection vulnerabilities in skill packages \citep{mecca2026skillsecurerdetectingpatchingpromptinjection}. Containment does not validate the objective, and package scanning does not establish safety when packages are combined. SkillCascade motivates testing jointly installed skills. A protection claim must therefore specify whether it concerns execution effects, an inspected package, or the behavior of packages used together.

Privacy review must cover the data contained in generated skills and the proposed flows of those data. A signature binds a delivered asset to a version; separate authorization is required for its embedded attributes and intended disclosures. Review should therefore check permitted retention and use in both the original record and the reusable asset.

\subsubsection{Safeguards for Proposal Evaluation (S3)}

Admission evidence should bind the candidate, inputs, outputs, evaluator, and decision. BenchShield combines static analysis of reward-relevant pathways with infrastructure-side runtime evidence of their use \citep{zheng2026benchshieldformalmodelbackedinstrumentation}. Harness audits add semantic checks on edits. These methods have different limitations: modeled events may omit a pathway, while auditors may misclassify edits. These integrity checks should be paired with tests of whether the benchmark success criterion captures the intended task.

MemPot uses optimized memory honeypots and sequential retrieval evidence to detect extraction \citep{wang2026mempotdefendingmemoryextraction}. It assumes knowledge of the memory, the retriever, and some historical attack queries. Evaluation should report false alarms, detection delay, benign utility, and disclosure before intervention under the detector's stated access and observation conditions.

\subsubsection{Safeguards for Update Implementation (S4)}

We recommend recording asset sources, parents, reviewed versions, permissions, and dependencies so recovery can quarantine known descendants and test residual influence, including cached copies and external effects. Recovery requires sufficiently complete dependency records, checks of both $h_t$ and $B_t$, and restored benign service. MUTE combines provenance tracking, erasure, trajectory containment, and later audits to reduce influence regeneration in federated edge-agent experiments, with utility trade-offs \citep{ding2026whenunlearningfails}. Extension to individual LLM agents requires evaluation in that setting.

AIM separates private and shared memory through access controls \citep{johnson2026aimprivacyawareinteroperablememory}. For training releases, DP-SGD and privacy accounting provide a distinct form of conditional protection \citep{abadi2016deeplearningdifferentialprivacy}. Over a fixed finite horizon, if every release mechanism is conditionally $(\varepsilon_j,\delta_j)$-DP under the same adjacency and predetermined bounds, basic adaptive composition yields joint $(\sum_j\varepsilon_j,\sum_j\delta_j)$-DP protection \citep{dwork2014foundations}. Post-processing without further access to private data incurs no additional budget; the number of successors alone does not determine privacy cost. The adjacency must specify whether protection applies at the record or user level. DP does not replace consent or purpose limitation.

\subsubsection{Safeguards for Recursive Improvement (RI)}

A candidate can propose a new safeguard but should not serve as its sole approval authority. We recommend placing final change approval, permission roots, and recovery authority outside the scope that the candidate can modify. Changes to logging require renewed tests of evidence coverage, and changes to an interpreter require renewed policy checks. We infer this design requirement from observed tampering mechanisms. Its effectiveness requires tests of the changed enforcement boundary during subsequent use.

Changes to privacy mechanisms likewise require reassessment of their data sources, authorization, adjacency, clipping, noise, and accounting, where applicable. Tests should cover the revised $\mathcal U_{t+1}$ after reuse, including its data-use rules in operation and the successors it generates. Section~\ref{sec:governance} develops the corresponding requirements for human authority and intervention.

\subsection{Evidence Synthesis and Research Gaps}
\label{sec:security-evidence}

We synthesize evidence on inherited compromise and information exposure, then identify gaps in recovery, deletion, and control revalidation across successors and updates produced by revised improvement mechanisms.

\subsubsection{Findings across Updates and Successors}



Security studies trace how adversarial influence enters an agent and is retained or transformed during reuse. SkillJack reports persistence after source removal, using routing measures that include behavioral proxies \citep{ying2026skilljackpersistentskillbackdoors}; EVOMAL follows propagation through authored skill copies over a finite sequence \citep{wu2026evomalselfpoisoningselfevolvingcoding}. Harness audits inspect retained violations through execution records, with coverage shaped by record completeness and audit judgments \citep{wang2026auditingharnesstamperingselfimproving}. The tested interface, accessible assets, and sequence of reuse define the scope of each finding.

Roesner and Kohno connect poisoned evaluation feedback to inherited security failures \citep{roesner2026trustingtrust}. In their Hyperagents experiment, poisoned benchmark tasks induced the meta-agent to add persistent instructions directing the task agent to bypass HTTPS certificate verification. Successors generated insecure code on benign held-out tasks, and the instructions survived further evolution on a clean benchmark. Evolution on a benchmark that explicitly penalized the vulnerability removed the harmful instruction in the tested Hyperagents run. These results motivate separate tests of source removal, removal of inherited instructions, and secure behavior after reuse.

Privacy studies cover acquisition, contextual disclosure, memory extraction, exposure through persona skills, and cross-version inference. A lifecycle analysis follows information across these events and records the assets, versions, and access rights involved.

For mechanism revision, Hyperagents provides an editable meta-agent, while harness audits inspect violations in retained changes \citep{zhang2026hyperagents,wang2026auditingharnesstamperingselfimproving}. In the Hyperagents poisoning experiment, the documented change concerns task-agent instructions, and the controls vary the evolution benchmark \citep{roesner2026trustingtrust}. Establishing a revision to $\mathcal{U}_t$ requires identifying the modified updater components and tracing their reuse in subsequent updates. Isolating the effect of that revision on contamination or recovery requires comparison with the original mechanism from matched inherited states.

\subsubsection{Coverage, Limitations, and Research Directions}

Longer-term security evaluation needs to trace contamination from memory entry through retrieval and skill extraction to later reuse, and to assess recovery across interacting successors. Evidence integrity should also be reassessed after evaluator or logger revisions. Privacy evaluation similarly needs to specify derived assets, accessible versions and releases, access rights, auxiliary data, and query conditions for deletion tests. Detailed audit logs can contain sensitive information, so access to the records used for these evaluations also belongs in the privacy assessment. Section~\ref{sec:governance} discusses the associated access and oversight requirements.

We propose comparing source-only cleanup with cleanup of both sources and descendants, and evaluating subsequent harm alongside the restoration of benign service. For deletion tests, the accessible memories, skills, checkpoints, and logs should be specified in advance, together with prior disclosures and recipient access. The tests should be repeated after reuse of a revised updater. Paired trajectories using original and revised $\mathcal U$ should start from matched inherited states with the same tasks, attacker and private-data access, budgets, and external acceptance tests. Following later updates would measure how mechanism revisions affect recovery and information exposure. Section~\ref{sec:evaluation} develops the protocols.

\begin{rsitakeaway}
\begin{itemize}[leftmargin=*,nosep]
\item \textbf{Risks.} Studies demonstrate poisoning of memory, skills, and evidence used to accept updates, as well as inappropriate collection and exposure of information carried into later versions. Persistent contamination alone does not show that the improvement procedure itself has been revised.
\item \textbf{Safeguards.} Verifying the origin of artifacts, constraining execution, protecting evaluation, controlling data access, and tracking cumulative privacy loss address specific risks. These protections depend on trusted enforcement and clearly specified data access. Their coverage remains limited after artifacts are transformed or combined, or the improvement procedure is revised.
\item \textbf{Future work.} Compare cleanup of contaminated source artifacts and their derivatives, recovery of normal service, and information deletion across historical versions. Retest protections after a revised improvement procedure is reused, using matched resource budgets. Report harm and information disclosure alongside performance on legitimate tasks and rejection of legitimate inputs or updates.
\end{itemize}
\end{rsitakeaway}
\par\endgroup

\section{Transparency and Controllability}
\label{sec:governance}

Updates to $h_t$ and $B_t$ in
Eqs.~\eqref{eq:rsi-model-update}--\eqref{eq:rsi-self-improvement}
can invalidate explanations and alter the effects of intervention,
while revisions to $\mathcal{U}_t$ can change how subsequent updates
are justified, reviewed, and authorized.
This section examines whether system changes remain understandable,
correctable, and attributable, tracing risks and assessing safeguards
for effective oversight across individual updates, retained successors,
and revisions to the improvement mechanism.

\subsection{Definitions and Trustworthiness Claims}
\label{sec:governance-definitions}

We define the transparency, control, and attribution requirements
for effective oversight and specify the evidence needed to assess
them in individual updates, retained successors, and updates
produced by a revised improvement mechanism.

\subsubsection{Definitions and Boundaries}

Transparency makes relevant sources, operations, and changes available
for inspection; interpretability and explainability support understanding
of system mechanisms and behavior
\citep{li2023trustworthy,nist2023airmf}.
Explanations require faithfulness tests, since observable traces may
reveal violations without identifying the underlying computation
\citep{guan2025monitorability}.
Controllability concerns whether interventions achieve their intended
effects
\citep{shen2026controllable,xie2025bone,zhang2025control,
chen2023controllable,shen2026stylitruth}.
Human control concerns whether authorized actors can inspect, redirect,
interrupt, and correct the system, while corrigibility concerns the
system's cooperation with correction
\citep{hleg2019ethics,soares2015corrigibility}.
Assessing these properties requires tests of intervention effects and
cooperation with correction.
Attribution concerns how changes contribute to gains and failures and how responsibility is assigned.
We distinguish technical contribution, operational responsibility, and governance accountability: which changes contributed to an outcome, which agents had assigned duties, and which people or organizations must explain decisions and address
consequences \citep{nian2026auditable,raji2020accountability}.

\subsubsection{Claims for Individual Updates, Successors, and Meta-Improvement}

An individual-update claim requires inspectable evidence of a specified change and tests of the relevant explanations and controls before and after the update.
Where effective review or correction is claimed, the corresponding interventions must achieve their intended effects under declared conditions.
A successor claim additionally requires evidence that the change is inherited and actually used, followed by tests that explanations remain faithful, authorized interventions remain effective, and responsibility assignments still cover the resulting state.

A meta-improvement claim requires an observed revision from $\mathcal{U}_t$ to $\mathcal{U}_{t+1}$, subsequent reuse, and tests of whether oversight remains effective for the updates produced
by the revised mechanism.
The modification boundary specifies which components may be revised,
while the control boundary determines who may propose, approve,
implement, interrupt, or revoke changes.
Components belong to $h$, $B$, or $\mathcal{U}$ according to their function rather than their file type.
The relevant explanations, intervention channels, and responsibility assignments must remain applicable when the revised mechanism is reused.
A claim of improved oversight further requires a controlled comparison with the unchanged $\mathcal{U}_t$ under comparable conditions.

\begin{figure*}[t]
    \centering
    \includegraphics[
        width=1.0\textwidth,trim=0cm 0.7cm 0cm 0.cm, clip
    ]{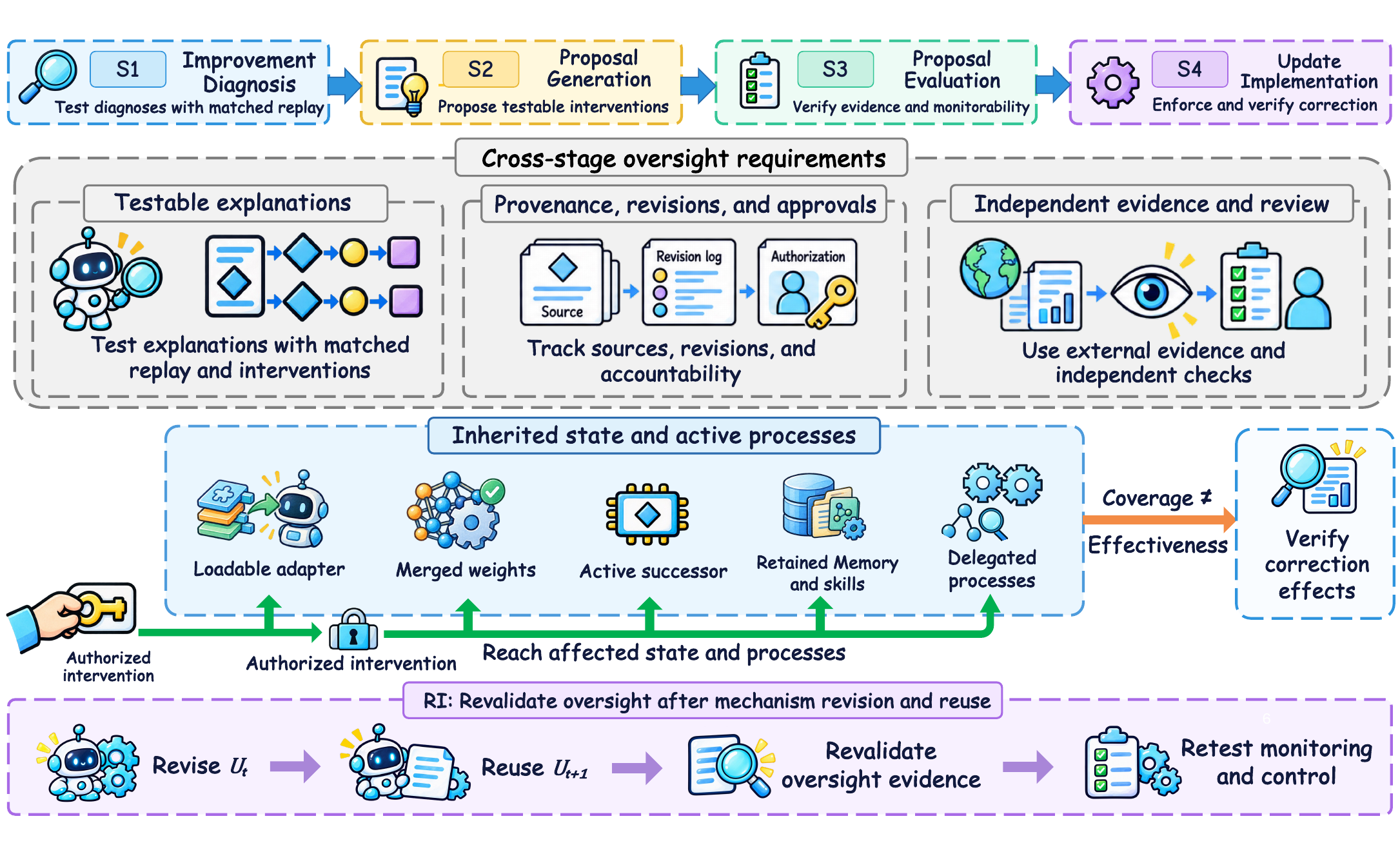}
    \caption{\textbf{Transparency and Controllability Across S1–S4 and RI in the Process of RSI.} The four stages connect testable diagnoses, proposed interventions, independent evaluation, and enforceable correction. Cross-stage oversight requires testable explanations, traceable sources and revisions, and independent review. Authorized interventions must reach affected inherited components and active processes, while separate tests establish their actual effects. The lower panel extends these checks to monitoring and control after improvement-mechanism revision and reuse.}
    \label{fig:chap6_figure}
\end{figure*}

\subsection{Risks and Failure Evidence}
\label{sec:governance-risks}

We trace failures of explanation, review, intervention, and attribution
across S1--S4, then examine how revising and reusing the improvement
mechanism can undermine the evidence and authority on which
oversight depends.

\subsubsection{Risks in Improvement Diagnosis (S1)}

A diagnosis that targets the wrong type of failure can direct correction to the wrong step.
MAST distinguishes failures of design, coordination, and verification
\citep{cemri2025mast}; AgentRx locates the first unrecoverable failure, whereas
TraceElephant examines role-dependent recovery duties
\citep{barke2026agentrx,chen2026traceelephant}. These targets are complementary:
the initiating error, loss of recoverability, and neglected recovery
opportunity need not coincide. A transparent diagnosis must specify which
of these failures it locates.

Even a traceable diagnosis can rely on unfaithful evidence or an unauthorized
target. Self-reports predict learned behavior in controlled tasks
\citep{betley2025selfreports}, yet naturalistic reasoning traces can rationalize
incompatible answers \citep{arcuschin2026wildfaithfulness}. Predictive usefulness in the controlled setting therefore calls for separate tests of
explanation faithfulness in naturalistic tasks. Reusing an unchecked account as $E_{t+1}$
or $F_{t+1}$ can preserve a mistaken diagnosis. Feedback optimization can also
reward manipulation in simulated interactions
\citep{williams2025manipulation}, allowing the feedback signal to improve without advancing the
authorized target. Control therefore requires checking whose requirements the
feedback represents before using it to decide what the system should change.

\subsubsection{Risks in Proposal Generation (S2)}

A proposal can have an apparently informative rationale that does not validly
explain the behavior under revision. Activation interventions can
recruit normally inactive paths \citep{makelov2024illusion}; faithfulness
metrics can lose their ability to discriminate on longer constructed reasoning traces
\citep{gurarieh2026faithfulnessmetrics}. The first result questions whether an
intervention explains normal computation, while the second questions the reliability of the
test used to assess the explanation. Selecting revisions using either unvalidated signal can
preserve a misleading rationale in later proposals.

An intervention may fail to produce its intended behavioral effect.
Tested models can recover from task-inconsistent activation steering
\citep{mckenzie2026steeringresistance}. This finding concerns the effect of the intervention
and does not demonstrate intentional resistance to shutdown. Successors therefore require behavioral checks of inherited edits, including unintended effects.
These checks assess whether an edit controls the intended outcome; faithfulness tests assess the rationale offered for it.

\subsubsection{Risks in Proposal Evaluation (S3)}

Evidence used for review can remain dependent even when several judges assess it.
Weak-to-strong learning can produce models that outperform the supervisor
\citep{burns2024weaktostrong}, but correlated errors limit the value of
additional judges \citep{goel2025thinkalike}. Research agents introduce another source of
evidence dependence because they can control experiments, scoring, and reports,
leaving reviewers to assess an account produced by the agent. 
Some exploits also survive review by a non-overlapping panel \citep{huang2026reward}.
RSI Control Arena tests covert sabotage during successor-model training and finds successful attacks that preserve performance on the assigned task.
Access to agent reasoning improves monitor detection in the tested settings, while some attacks still evade review \citep{zhou2026rsicontrol}.
Independent review therefore requires evidence beyond differences in panel membership.

Adaptation can also change what reviewers observe and how they respond.
Penalizing detected reasoning can suppress visible evidence without eliminating
unwanted actions \citep{baker2025monitoring}. Output-only supervision can alter
reasoning through shared parameters \citep{drori2026spillover}, so avoiding
direct penalties on the observation channel does not isolate it from the effects of training.
Adaptive attacks also exploit escalation rules
\citep{terekhov2026adaptivemonitors}. Under an explicit evasion objective,
research agents adapt in context to detailed review feedback without weight
updates \citep{huang2026reward}. Because this feedback combines decisions, reasons, and
history, the effect of explanations alone has not been isolated. Changes to either the
observation channel or its decision protocol can therefore alter the evidence used to
accept later revisions.

\subsubsection{Risks in Update Implementation (S4)}

The implemented system can differ from the configuration approved during review.
The running system must match the accepted revision, including
non-weight state, because skill-library drift can change behavior even with a fixed base
model \citep{zhang2026librarydrift}. Traceability must therefore cover the retained state active during execution and how
the successor uses it.

An available intervention may fail to reach the affected processes or
produce a useful correction. Studies of interruptibility examine the points at which execution
can be stopped \citep{wu2026interruptible}; comparisons from matched execution prefixes distinguish
predicted risk from the benefit of intervening
\citep{zhang2026interventionvalue}. Intervention tests should assess both availability and benefit, checking whether
delegated processes stop and prior effects are reversed, including when
the affected state is inherited by a successor.

\subsubsection{Risks under Recursive Improvement (RI)}

RI applies S1--S4 to $\mathcal U_t$.
DGM retains agent-code changes, while Dream-RSI reuses revised exploration
policies \citep{zhang2025dgm,zheng2026dreamrsi}. Recursive Reward Improvement
revises task-and-rubric generation guidance using exploit feedback
\citep{dai2026recursivereward}. Evo$^2$ trains a shared proposer--reflector policy
on harness edits and outcomes, then reuses it in search with a fixed task executor
\citep{xu2026evo2}. These examples document changes to retained code, search,
or evidence generation. Assessing continued oversight requires tests after reuse.

Earlier explanations, evaluations, and oversight procedures must be rechecked
when the revised mechanism guides subsequent updates.
Changes to search or evidence selection may invalidate review conditions;
changes to approval or stopping authority may also move the control boundary.

\subsection{Existing Safeguards and Their Limitations}
\label{sec:governance-safeguards}

We review safeguards across S1--S4 and assess the evidence and
conditions needed to support diagnosis, independent review, and
authorized intervention as agent configurations and improvement
mechanisms change.

\subsubsection{Safeguards for Improvement Diagnosis (S1)}
\label{sec:governance-diagnosis}

Testable failure attribution distinguishes the accuracy of failure
localization from the utility of a proposed correction. Replay with
injected faults provides controlled localization targets
\citep{liu2026whowhenpro}, while matched replay shows that
diagnosis-guided corrections can improve recovery over retries
\citep{zhu2026aed}. The two approaches address different questions:
where a failure occurred and whether the proposed correction improves
recovery. The latter can be assessed by comparing continuations from
the same restored execution state under comparable execution conditions and budgets.

Faithfulness tests assess the explanation underlying a diagnosis.
Behavioral prediction tests assess how useful an explanation is to an observer;
tests that remove a cue examine the causal dependence claimed by the explanation
\citep{mayne2026positivefaithfulness,ferreira2026causalexplanations}.
Mechanistic explanations further distinguish circuit localization from the
recovery of causal variables \citep{mueller2025mib,sun2026neurocogmap,sun2026mechanistic}. These tests assess
different explanatory claims. When a successor changes the explained behavior or
mechanism, the tests and matched intervention comparisons should be repeated on
the state it actually inherits.

\subsubsection{Safeguards for Proposal Generation (S2)}
\label{sec:governance-controlled-revision}

Safeguards for proposals connect the target and provenance of an edit to its measured
effects. Sparse feature circuits locate behavioral edit targets, whereas
AxBench compares steering with simpler baselines
\citep{marks2025circuits,wu2025axbench}. Circuit analysis identifies candidate intervention targets;
benchmark comparisons assess their practical value. 
A proposal should specify the edited component and intervention procedure, and assess both intended and collateral effects.
HGM-Dual records each harness edit together with added or removed functionality, changes in constraint violations, and observed improvements or regressions
\citep{paul2026hgmdual}. These records guide subsequent proposals by preserving the outcomes of earlier modifications.
Text-to-LoRA, SHINE, and Doc-to-LoRA use trained hypernetworks to generate adapters from task text or context
\citep{charakorn2025texttolora,liu2026shine,charakorn2026doctolora}; Skill-to-LoRA
instead learns from demonstrations, and LatentSkill provides loadable skill
adapters \citep{zhang2026skilltolora,yu2026latentskill}.

Direct generation and skill-specific learning require provenance and
effect checks suited to their representations, including the source,
generator or teacher, base model, produced adapter, and whether it
was loaded and used. Utility alone does not establish faithful semantic incorporation
\citep{zhao2026faithful}, motivating source substitution and adapter ablation.
Repeated outputs from a fixed hypernetwork do not establish revision of $\mathcal U_t$.
For textual memory, Memory Echo tests source-family selection,
correction-conditioned eligibility, and authoritative evidence without deleting history;
replay with supplied provenance reduces historical influence, but benefits depend
on an actual conflict and reliable corrections \citep{li2026memoryechoinsingleagentl}.
Table~\ref{tab:governance-representations} summarizes the traceability
records and control checks for textual memories and skills, loadable
adapters, and merged parameters, following the controllable-learning
perspective \citep{shen2026controllable}. These checks assess intended
and collateral effects in the active configuration and any derived state
that retains the revised information.

\begin{table}[!htbp]
\centering
\footnotesize
\setlength{\tabcolsep}{3.5pt}
\renewcommand{\arraystretch}{1.18}
\caption{\textbf{Traceability and intervention checks for retained experience.}}
\label{tab:governance-representations}
\begin{tabularx}{\linewidth}{@{}
  >{\raggedright\arraybackslash}p{.275\linewidth}
  >{\raggedright\arraybackslash}p{.345\linewidth}
  >{\raggedright\arraybackslash}X@{}}
\toprule
\textbf{Retained form} & \textbf{Trace} & \textbf{\ensuremath{\rightarrow} Control test} \\
\midrule
\rowcolor{rsiChapterPurpleStripe}
\textbf{Textual memories and skills} &
\textbf{Source}, active text, retrieval / selection, actual use &
\textbf{Edit / exclude}; test influence remaining in copies \\
\addlinespace[2pt]
\textbf{Loadable adapters} &
\textbf{Source / generator / teacher version}; base and active adapters &
\textbf{Swap / load / scale / ablate} under composition \\
\addlinespace[2pt]
\rowcolor{rsiChapterPurpleStripe}
\textbf{Merged parameters} &
\textbf{Contributors / merge}; active and derived weights &
\textbf{Verify correction} in merged and derived state \\
\bottomrule
\end{tabularx}
\end{table}

\subsubsection{Safeguards for Proposal Evaluation (S3)}

Evaluation safeguards give reviewers access to evidence and support tests of the claims based on it. AgentMonBench tests whether monitors identify consequential agent decisions, including undisclosed implementation changes, and locate
supporting evidence for user verification \citep{sun2026agentmonbench}.
Its Evidence-Grounded Behavior Graph (EBG) organizes source-linked behaviors and improves decision identification and evidence localization in most tested
settings. Aggregating weak supervision combines imperfect labels, while debate exposes disagreements that a reviewer can check \citep{jeon2025w2sshift,kenton2024weakjudges}. 
The value of aggregation depends on the supervision combined, whereas the value of debate depends on the
evidence available for verification.
Human debate improves factual judgments in two studied domains \citep{rahman2025humandebate}; formal results on debate require explicit computational and oracle assumptions \citep{browncohen2024debate}.
Applying these review methods to stronger successors requires reassessing the available evidence and the assumptions on which each method depends.

Acceptance also requires independent checks and explicit escalation rules.
Research-agent audits show that full trajectories and held-out reruns can
leave some evaluation exploits undetected \citep{huang2026reward}, motivating
claim-specific checks using evidence outside the control of the agent that produced the result.
These checks should assess the claimed outcome as well as score reproducibility. Hybrid monitors combine
signals under a review budget \citep{kale2026monitoring}; selective escalation
bounds disagreement with human labels under calibration assumptions
\citep{jung2025escalate}. Combining observations and deciding when to defer are
complementary safeguards. The bound for selective escalation concerns agreement, not label
correctness. Acceptance procedures must specify the available feedback, review resources,
and action triggered by uncertainty. These conditions must then be retested when the
observation channel changes.

\subsubsection{Safeguards for Update Implementation (S4)}
\label{sec:governance-implementation}

Implementation safeguards make authorized intervention enforceable on the
running system. AI Control combines monitoring with trusted editing or deferral
\citep{greenblatt2024control}; external mediation places authorization outside
the acting policy \citep{costa2026fides,dobrin2026kernel}. The protocol determines the
response, while mediation limits opportunities to bypass it. The effectiveness of both approaches
depends on trusted components, review resources, and coverage of relevant
actions; an unmediated descendant remains outside the claimed boundary.
Cooperation is a separate requirement: safe interruptibility addresses
learning incentives, whereas goal-update acceptance addresses cooperation with
changed objectives \citep{orseau2016interruptible,hudson2026corrigibility}.
The formal results rely on different algorithmic or objective assumptions.
Assessing corrigibility requires checking those assumptions and the system's cooperation
with correction across the relevant conditions.

Correction and attribution must also account for the state inherited by successors.
Behavioral-shift tests identify changes in scored outputs
\citep{richter2025behavioralshift}, while Parametric Skills shows how
contributions persist in merged weights \citep{zhao2026parametric}.
Effective correction must reach affected memories, merged weights, and
delegated processes, with behavioral tests verifying its effects.
Irreversible external effects require separate remedies.
Trace audits, data ablations, and model--harness comparisons provide
complementary evidence for attributing outcomes to particular changes
\citep{zhou2026traceaudit,chen2026mechanisticdata,yang2026frontis}.
Attribution across versions requires controlled comparisons of preceding
revisions and their interactions. Authorization and delegation records
further clarify operational duties and governance accountability
\citep{nian2026auditable}.
Table~\ref{tab:governance-attribution} distinguishes the evidence required
for these three attribution questions.

\begin{table}[!htbp]
\centering
\footnotesize
\setlength{\tabcolsep}{3.5pt}
\renewcommand{\arraystretch}{1.18}
\caption{\textbf{Three forms of attribution and their evidence.}}
\label{tab:governance-attribution}
\begin{tabularx}{\linewidth}{@{}
  >{\raggedright\arraybackslash}p{.28\linewidth}
  >{\raggedright\arraybackslash}p{.30\linewidth}
  >{\raggedright\arraybackslash}X@{}}
\toprule
\textbf{Attribution} & \textbf{Question} & \textbf{Evidence} \\
\midrule
\rowcolor{rsiChapterPurpleStripe}
\textbf{Technical} contribution &
\textbf{Which changes} affected the outcome? &
\textbf{Component ablations}; interactions across revisions \\
\addlinespace[2pt]
\textbf{Operational} responsibility &
\textbf{Which duties} applied / were fulfilled? &
\textbf{Assigned roles}; available information; recovery actions \\
\addlinespace[2pt]
\rowcolor{rsiChapterPurpleStripe}
\textbf{Governance} accountability &
\textbf{Who must explain} and remedy? &
\textbf{Authorization / delegation}; review and action records \\
\bottomrule
\end{tabularx}
\end{table}

\subsubsection{Safeguards for Recursive Improvement (RI)}

Protecting control components constrains direct changes to oversight. AHE excludes
its verifier and tracer from the editable set \citep{lin2026ahe}. L3M uses a Lean
runtime with checked permission properties and requires an actor other than the optimizer
to control specifications and evaluation criteria under a separate grant \citep{eisenach2026l3m}.
The adequacy of the protected checks and the effectiveness of control require testing.
After reuse of $\mathcal U_{t+1}$ in Eq.~\eqref{eq:rsi-self-improvement},
intervention tests must cover affected permissions and processes.

The evidence supporting oversight must also be renewed. Compact mechanistic proofs
certify performance for a small attention-only transformer \citep{gross2024compactproofs};
safety cases organize claims, evidence, and unresolved assumptions
\citep{clymer2024safetycases}. Formal results cover encoded properties under their stated assumptions;
L3M does not provide a session-wide safety theorem \citep{eisenach2026l3m}.
When RI changes these properties or assumptions, the affected proofs and
empirical oversight tests must be revisited under subsequent use of the revised mechanism.

\subsection{Evidence Synthesis and Research Gaps}
\label{sec:governance-evidence}

We synthesize evidence on diagnosis, monitoring, correction, and
attribution, then identify the longitudinal comparisons needed to
assess explanation validity and the effectiveness of review and
control across successors and mechanism revisions.

\subsubsection{Findings across Updates and Successors}

Replay-based diagnosis, explanation testing, monitoring, and execution controls address different parts of oversight. Replay evaluates proposed repairs under matched execution conditions \citep{zhu2026aed}; explanation tests assess predictions and intervention effects. Monitoring depends on what observation channels reveal, while authorized intervention depends on trusted controls that cover the relevant actions. These dependencies connect oversight quality to recoverable execution, informative records, and the scope of control.

Inheritance changes the state requiring explanation or correction, and mechanism revision changes how proposals and their supporting evidence are produced. Longitudinal evaluation should therefore track the usefulness of diagnoses, the reach of corrections into retained and derived state, and the effectiveness of review and stopping after mechanism reuse.

\subsubsection{Coverage, Limitations, and Research Directions}

Continuity of explanation depends on predictive and intervention validity across revisions. Equivalent interpretations can apply to different networks \citep{sun2026equivalence}, so the relevant comparison concerns explanatory performance across changed representations. Adaptive attacks and training spillovers motivate separate tests of changes to the actor, monitor, and observation channel \citep{terekhov2026adaptivemonitors,drori2026spillover}. Intervention evaluation should also follow stopping, retained-state correction, and task recovery after inheritance \citep{zhang2026interventionvalue,zhao2026parametric}. Rollback assessments should separate recoverable system state from irreversible external effects.

Future studies should retest explanation validity and use matched-checkpoint replay to track repair durability after subsequent updates. Comparisons of original and revised $\mathcal U$ should start from matched inherited states with the same tasks, access, budgets, and external review criteria, then follow review effectiveness, stopping success, residual effects of correction, and benign-task utility over later updates. Feedback comparisons should vary access to review decisions, explanations, and retained history, while measuring detection, legitimate correction, and evasion at fixed attempt and review budgets. Independent investigators should reconstruct failures and breakthroughs from incomplete or merged histories using ablations of preceding revisions, interaction tests, and intervention evidence, with assigned duties and authorization recorded separately \citep{nian2026auditable}.

\begin{rsitakeaway}
\begin{itemize}[leftmargin=*,nosep]
\item \textbf{Risks.}
Misleading diagnoses, unfaithful explanations, and dependent review evidence can undermine oversight.
Ineffective interventions may allow failures to persist across successors and recursive updates.
\item \textbf{Safeguards.}
Matched replay, faithfulness tests, and provenance tracking support reliable diagnosis and traceability.
Independent review and enforceable controls help ensure effective oversight and correction.
\item \textbf{Future work.}
Assess explanation validity, monitoring effectiveness, correction reach, and attribution across successors.
Use matched comparisons to test whether oversight remains effective after improvement-mechanism revisions.
\end{itemize}
\end{rsitakeaway}

\section{Evaluating Trustworthy RSI}
\label{sec:evaluation}

Sections~\ref{sec:validity}--\ref{sec:governance} examine risks and safeguards
across the four trustworthiness themes. This section brings together
evaluation evidence for the update, successor, and meta-improvement claims
introduced in Section~\ref{sec:rsi-trustworthiness}.
We review evidence from existing studies (Section~\ref{sec:eval-benchmarks}),
develop comparison designs for successive system versions
(Section~\ref{sec:eval-longitudinal}), and examine the conditions under which
the evidence supports trustworthiness claims (Section~\ref{sec:eval-evidence}).
Throughout, the four themes specify the requirements to be assessed, while
S1--S4 and RI identify where changes and feedback arise in the improvement process.

\subsection{What Do Existing Evaluations Cover?}
\label{sec:eval-benchmarks}
\label{sec:eval-objects}


\begin{figure*}[!t]
\centering

\definecolor{rsiCellInk}{HTML}{0B2F78}
\definecolor{rsiCellGray}{HTML}{E7E7E7}

\definecolor{rsiValidity}{HTML}{43855F}
\definecolor{rsiConstraints}{HTML}{3975AA}
\definecolor{rsiOversight}{HTML}{806391}
\definecolor{rsiDesign}{HTML}{707C88}

\definecolor{rsiUpdate}{HTML}{AC7029}
\definecolor{rsiSuccessor}{HTML}{2C8791}
\definecolor{rsiMeta}{HTML}{AE5C76}

\definecolor{rsiUpdateText}{HTML}{8B581F}
\definecolor{rsiSuccessorText}{HTML}{216E78}
\definecolor{rsiMetaText}{HTML}{914662}

\resizebox{\textwidth}{!}{%
\begingroup

\begin{tikzpicture}[
  x=1cm,
  y=1cm,
  every node/.style={font=\rmfamily,text=black},
  tile/.style={
    rounded corners=0.03cm,
    draw=none
  },
  rowlabel/.style={
    anchor=west,
    text=white,
    inner sep=0pt,
    font=\rmfamily\fontsize{8.4}{9.2}\selectfont
  },
  studylabel/.style={
    anchor=east,
    inner sep=0pt,
    font=\rmfamily\fontsize{7.5}{8.2}\selectfont
  },
  grouphead/.style={
    font=\rmfamily\fontsize{9}{10}\selectfont,
    inner sep=0pt
  }
]


\def\RSICell{0.35}
\def\RSIPitch{0.42}
\def\RSIColGap{0.17}

\def\RSIGridX{4.52}
\def\RSINameX{1.56}
\def\RSINameRight{4.36}

\def\RSILabelAngle{46}

%
\newcommand{\RSIGetX}[1]{%
  \pgfmathsetmacro{\RSIX}{%
    \RSIGridX+\RSIPitch*#1
    +\RSIColGap*(#1>11)
    +\RSIColGap*(#1>17)}%
}

\newcommand{\RSIGetY}[1]{%
  \ifcase#1\relax
    \def\RSIY{0.00}
  \or
    \def\RSIY{-0.43}
  \or
    \def\RSIY{-0.94}
  \or
    \def\RSIY{-1.37}
  \or
    \def\RSIY{-1.80}
  \or
    \def\RSIY{-2.38}
  \or
    \def\RSIY{-3.49}
  \or
    \def\RSIY{-3.92}
  \or
    \def\RSIY{-4.35}
  \or
    \def\RSIY{-4.78}
  \or
    \def\RSIY{-5.21}
  \or
    \def\RSIY{-5.64}
  \or
    \def\RSIY{-6.07}
  \fi
}


\newcommand{\RSIColumnGroup}[5]{%
  \RSIGetX{#1}%
  \edef\RSILeft{\RSIX}%
  \RSIGetX{#2}%
  \pgfmathsetmacro{\RSIRight}{\RSIX+\RSICell}%

  \path[fill=#3!14,tile]
    (\RSILeft,0.38)
    rectangle
    (\RSIRight,0.98);

  \node[grouphead,text=#5]
    at ({(\RSILeft+\RSIRight)/2},0.68)
    {#4};
}

\RSIColumnGroup{0}{11}{rsiUpdate}{Update-level}{rsiUpdateText}

\RSIColumnGroup{12}{17}{rsiSuccessor}{Successor-level}{rsiSuccessorText}

\RSIColumnGroup{18}{19}{rsiMeta}{Meta}{rsiMetaText}

\node[
  anchor=west,
  inner sep=0pt,
  font=\rmfamily\fontsize{10.2}{11}\selectfont
] at (0,0.68) {Trustworthiness evidence};


\newcommand{\RSICellMark}[3]{%
  \ifcase#3\relax
    \path[fill=rsiCellGray,tile]
      (#1,{#2-\RSICell/2}) rectangle
      ({#1+\RSICell},{#2+\RSICell/2});
  \or
    \path[fill=rsiCellInk,tile]
      (#1,{#2-\RSICell/2}) rectangle
      ({#1+\RSICell},{#2+\RSICell/2});
  \or
    \path[fill=white,draw=rsiCellInk,line width=0.55pt,
          rounded corners=0.03cm]
      (#1,{#2-\RSICell/2}) rectangle
      ({#1+\RSICell},{#2+\RSICell/2});
  \or
    \path[fill=white,draw=rsiDesign,line width=0.4pt,
          rounded corners=0.03cm]
      (#1,{#2-\RSICell/2}) rectangle
      ({#1+\RSICell},{#2+\RSICell/2});
    \node[inner sep=0pt,font=\rmfamily\fontsize{8}{9}\selectfont]
      at ({#1+\RSICell/2},#2) {?};
  \fi
}

\foreach \x/\code/\lab in {
  0.00/1/Examined,
  2.75/2/Discussed or case,
  6.35/0/Not evaluated,
  9.45/3/Unclear
}{%
  \RSICellMark{\x}{1.22}{\code}%
  \node[anchor=west,inner sep=0pt,
        font=\rmfamily\fontsize{8}{9}\selectfont]
    at ({\x+0.44},1.22) {\lab};
}


\newsavebox{\RSIRoleBox}

\newcommand{\RSIRole}[4]{%
  \path[fill=#3,tile]
    (0,#2) rectangle (1.43,#1);

  \node[
    text=white,
    inner sep=0pt,
    font=\rmfamily\fontsize{7.5}{8.4}\selectfont
  ] at (0.715,{(#1+#2)/2}) {%
    \sbox{\RSIRoleBox}{\shortstack{#4}}%
    \ifdim\wd\RSIRoleBox>1.31cm
      \resizebox{1.31cm}{!}{\usebox{\RSIRoleBox}}%
    \else
      \usebox{\RSIRoleBox}%
    \fi
  };
}

\RSIRole{0.17}{-0.60}{rsiValidity}
  {Improvement\\validity}

\RSIRole{-0.77}{-1.97}{rsiConstraints}
  {Constraint\\preservation}

\RSIRole{-2.035}{-2.72}{rsiOversight}
  {Effective\\oversight}


\draw[rsiCellGray,line width=0.4pt]
  (0,-2.87) -- (13.20,-2.87);

\node[
  anchor=west,
  inner sep=0pt,
  text=black,
  font=\rmfamily\fontsize{9.0}{10.0}\selectfont
] at (0,-3.09) {Supporting evaluation design};

\path[fill=rsiDesign,tile]
  (0,-6.24) rectangle (1.43,-3.30);

\node[
  rotate=90,
  text=white,
  inner sep=0pt,
  font=\rmfamily\fontsize{9}{10}\selectfont
] at (0.715,-4.77) {Evaluation design};


\newsavebox{\RSILabelBox}

\newcommand{\RSIEvidenceRow}[4]{%
  \RSIGetY{#1}%

  \path[fill=#2,tile]
    (\RSINameX,{\RSIY-\RSICell/2})
    rectangle
    (\RSINameRight,{\RSIY+\RSICell/2});

  \node[rowlabel]
    at ({\RSINameX+0.075},\RSIY) {%
      \sbox{\RSILabelBox}{#3}%
      \ifdim\wd\RSILabelBox>2.57cm
        \resizebox{2.57cm}{!}{\usebox{\RSILabelBox}}%
      \else
        \usebox{\RSILabelBox}%
      \fi
    };

  \foreach \v [count=\c from 0] in {#4}{%
    \RSIGetX{\c}%
    \RSICellMark{\RSIX}{\RSIY}{\v}%
  }%
}

\RSIEvidenceRow{0}{rsiValidity}{Task performance}
  {1,1,1,1,1,1,1,1,1,1,1,1,
   1,1,1,1,1,1,
   1,1}

\RSIEvidenceRow{1}{rsiValidity}{Retention / transfer}
  {0,0,2,1,1,0,0,0,2,0,0,0,
   1,1,0,1,1,1,
   1,1}


\RSIEvidenceRow{2}{rsiConstraints}{Safety / security}
  {2,2,2,0,0,0,2,2,2,1,1,1,
   2,2,2,2,1,1,
   0,2}

\RSIEvidenceRow{3}{rsiConstraints}{Fairness}
  {0,0,0,0,0,0,0,0,0,0,0,1,
   0,0,2,0,0,0,
   0,2}

\RSIEvidenceRow{4}{rsiConstraints}{Privacy}
  {0,0,0,2,0,0,0,0,2,1,0,2,
   0,0,2,0,1,1,
   0,0}


\RSIEvidenceRow{5}{rsiOversight}{Intervention}
  {0,0,0,0,0,0,0,0,1,0,1,0,
   2,0,2,2,1,1,
   0,2}


\RSIEvidenceRow{6}{rsiDesign}{Cost}
  {1,1,1,2,2,1,1,1,1,1,1,1,
   1,1,2,1,2,0,
   1,1}

\RSIEvidenceRow{7}{rsiDesign}{Experimental controls}
  {1,1,1,1,1,1,1,1,1,1,1,1,
   1,1,1,1,1,1,
   1,1}

\RSIEvidenceRow{8}{rsiDesign}{Acceptance}
  {0,0,1,1,1,1,1,1,1,0,0,0,
   1,1,0,1,2,0,
   1,1}

\RSIEvidenceRow{9}{rsiDesign}{Reuse analysis}
  {0,1,2,2,1,1,1,1,0,0,0,0,
   0,0,0,0,0,0,
   0,0}

\RSIEvidenceRow{10}{rsiDesign}{State reuse}
  {0,0,3,1,1,1,1,1,1,0,0,0,
   1,1,1,1,1,1,
   1,1}

\RSIEvidenceRow{11}{rsiDesign}{Updater edits}
  {0,0,0,0,3,0,0,0,0,0,0,0,
   0,0,0,1,0,0,
   1,1}

\RSIEvidenceRow{12}{rsiDesign}{Updates by revised updater}
  {0,0,0,0,3,0,0,0,0,0,0,0,
   0,0,0,1,0,0,
   1,1}

%
%

\foreach \lab/\c/\col in {
  RE-Bench/0/rsiUpdateText,
  PostTrainBench/1/rsiUpdateText,
  RSIBench-Data/2/rsiUpdateText,
  TEB/3/rsiUpdateText,
  SAV/4/rsiUpdateText,
  PACE/5/rsiUpdateText,
  SGM/6/rsiUpdateText,
  REUSE/7/rsiUpdateText,
  EvoUndo/8/rsiUpdateText,
  AgentDojo/9/rsiUpdateText,
  AI Control/10/rsiUpdateText,
  HELM/11/rsiUpdateText,
  EvoPathBench/12/rsiSuccessorText,
  Agent Plasticity/13/rsiSuccessorText,
  HB/14/rsiSuccessorText,
  DGM/15/rsiSuccessorText,
  Misevolution/16/rsiSuccessorText,
  SEABench/17/rsiSuccessorText,
  Dream-RSI/18/rsiMetaText,
  Hyperagents/19/rsiMetaText%
}{%
  \RSIGetX{\c}%

  \node[
    studylabel,
    rotate=\RSILabelAngle,
    text=\col
  ] at ({\RSIX+\RSICell/2},-6.38) {\lab};
}

\end{tikzpicture}
\endgroup
}

\caption[Evidence coverage in trustworthy RSI studies.]{%
\textbf{Evidence coverage in trustworthy RSI studies.}
Studies are grouped by their primary use: \textcolor{rsiUpdateText}{\textbf{update-level}},
\textcolor{rsiSuccessorText}{\textbf{successor-level}}, and
\textcolor{rsiMetaText}{\textbf{meta-improvement}}.
Rows cover the three overarching trustworthiness concerns and supporting evaluation designs.
The upper rows group selected evidence by the three analytical roles; the lower rows record supporting evaluation designs. Outlined cells indicate discussion or illustrative cases; gray cells indicate
not evaluated; and question marks indicate unclear evidence.
Task performance includes benign utility, and the final row records updates produced by
a retained, revised updater. Coding is at the study level.
Appendix~\ref{app:evidence-protocol} and the audit files provide coding criteria,
paper versions, source locations, and verification status.
}
\label{fig:eval-literature}
\end{figure*}
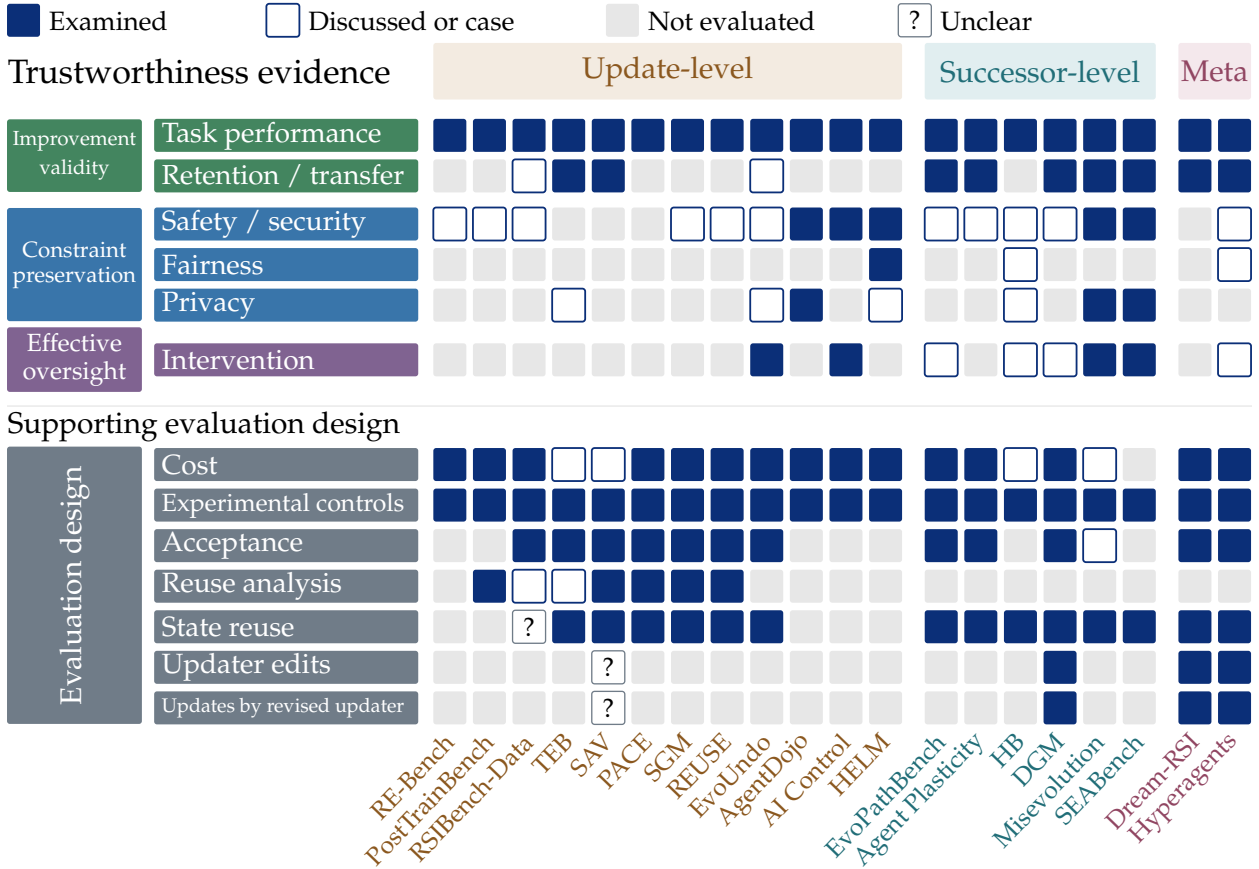

We group studies by the claim level that their evidence or methods primarily
inform. The four trustworthiness themes guide the assessment of evidence
coverage within each group. Some studies evaluate changes to evolving systems
directly; others provide task measures, attack tests, or oversight protocols
for assessing those changes.

\paragraph{Update-level evidence.}
At the update level, evaluation concerns a proposed revision and the
requirements it must satisfy.

RE-Bench compares research-engineering performance with human experts across
time budgets, while PostTrainBench evaluates autonomous post-training under
compute constraints~\citep{wijk2024rebench,rank2026posttrainbench}.
VibeTraining-Bench evaluates permitted modifications to training data,
trainers, and harnesses, starting from a common initial system and applying
task-specific resource limits. Isolated grading, protected test data, and
restricted network access separate validation feedback from hidden-test
reporting~\citep{peng2026vibetrainingbench}.
RSIBench-Data holds training infrastructure fixed to isolate data-research
decisions. Its non-monotonic trajectories and reuse of a task subset for both
selection and reporting make candidate selection and evaluation access
important to interpreting reported gains~\citep{meng2026rsibench}.
Trace Evolution Bench (TEB) measures score variation across repeated
evaluations of an unchanged agent. Frozen controls and seed-matched
comparisons distinguish this variation from performance changes associated
with evolution~\citep{li2026traceevolutionbench}.
Together, these evaluations show how performance results depend on resource
budgets, selection procedures, and reporting conditions.

The self-authored verification study (SAV in Figure~\ref{fig:eval-literature})
assesses the reliability of tests written by evolving systems~\citep{guo2026selfauthored}.
Complementary studies address security, oversight, and multidimensional
assessment. AgentDojo evaluates prompt-injection attacks and defenses,
while AI Control tests oversight protocols against deliberate
subversion~\citep{debenedetti2024agentdojo,greenblatt2024control}.
HELM provides a framework for multidimensional reporting~\citep{liang2022holistic}.

\paragraph{Successor-level evidence.}
At the successor level, evaluation tracks accepted changes through
inheritance and subsequent use. The archive maintained by DGM records
parent--successor relationships and alternative branches~\citep{zhang2025dgm}.
EvoPathBench uses accumulation, interference, and reversal streams to
assess generalization, retention, and adaptation to changing rules.
Held-out probes and interventions on artifact access relate checkpoint
performance to inherited artifacts~\citep{lin2026beyondendpointperformanc}.
Agent Plasticity relates gains on held-out tasks to acquisition costs
while keeping model weights fixed. Its trajectories distinguish endpoint
capability from learning efficiency~\citep{singh2026agentplasticity}.
The study of harness updating and benefit (HB in Figure~\ref{fig:eval-literature})
uses agent--evolver pairings to distinguish the ability to generate useful
harness updates from the ability to use them~\citep{lin2026harnessbenefit}.
These studies examine inherited changes, their subsequent use, and the
persistence of their benefits.

\begingroup
\setlength{\emergencystretch}{1em}
Misevolution examines risks associated with changes to models,
memory, tools, and workflows~\citep{shao2026misevolve}.
SEABench compares evolving and non-evolving personal assistants over
task sequences, tracing downstream safety and privacy failures to
locally useful adaptations~\citep{das2026seabench}.
EvoUndo evaluates recovery from one-shot harness modifications across
counterfactual states and identifies state grounding and the expressiveness
of recovery language as important factors~\citep{sah2026evoundo}.
\par
\endgroup

\paragraph{Meta-improvement evidence.}
At the meta-improvement level, evaluation concerns revisions to
$\mathcal U_t$ and their effects on subsequent updates.
Dream-RSI revises exploration-policy code using historical discovery
trees and redeploys the revised policy while keeping the discovery agent,
evaluator, and execution interfaces fixed. Comparisons with fixed
exploration policies assess the contribution of adaptive search allocation
in this setting~\citep{zheng2026dreamrsi}.
Hyperagents revises a meta-level modification program and evaluates
transferred meta agents using improvement@$50$, holding each meta agent
fixed while it generates task-agent variants. Separate continuation
experiments permit further meta-level revisions~\citep{zhang2026hyperagents}.
These studies illustrate two evaluation designs: comparisons with fixed
exploration policies and transfer tests of revised meta agents.
The matrix groups studies by their primary evaluation role and applies
common evidence criteria across the three claim levels.
Evaluations should identify the revised component of $\mathcal U_t$,
the state transferred with it, and the subsequent updates used to assess
its effects on capability, requirement satisfaction, and oversight.

Figure~\ref{fig:eval-literature} summarizes the evidence dimensions
covered by representative studies. The task-performance row records
measured task scores or benign utility. The reuse-analysis row covers
empirical or theoretical assessments of dependence arising from repeated
access to evaluation feedback. RSIBench-Data discusses this risk,
while SAV and PACE test the effects of evaluation feedback~\citep{meng2026rsibench,
guo2026selfauthored,shawn2026pace}.
SGM and REUSE provide improvement guarantees for a reused finite
evaluation pool and the underlying task population, respectively, subject
to their stated sampling and feedback conditions~\citep{wu2026sgm,sun2026reuse}.

Matrix coding distinguishes cost reporting from matched total-resource
comparisons and updater revisions from persistent task-state changes.
Unresolved cases are marked as unclear. Appendix~\ref{app:evidence-protocol}
describes the study-selection and evidence-coding protocol.
The next section develops comparisons of capability gains and requirement
preservation across successive versions.


\subsection{How Should Successive Versions Be Compared?}
\label{sec:eval-longitudinal}

\begin{figure*}[!htbp]
\centering
\includegraphics[width=\textwidth]{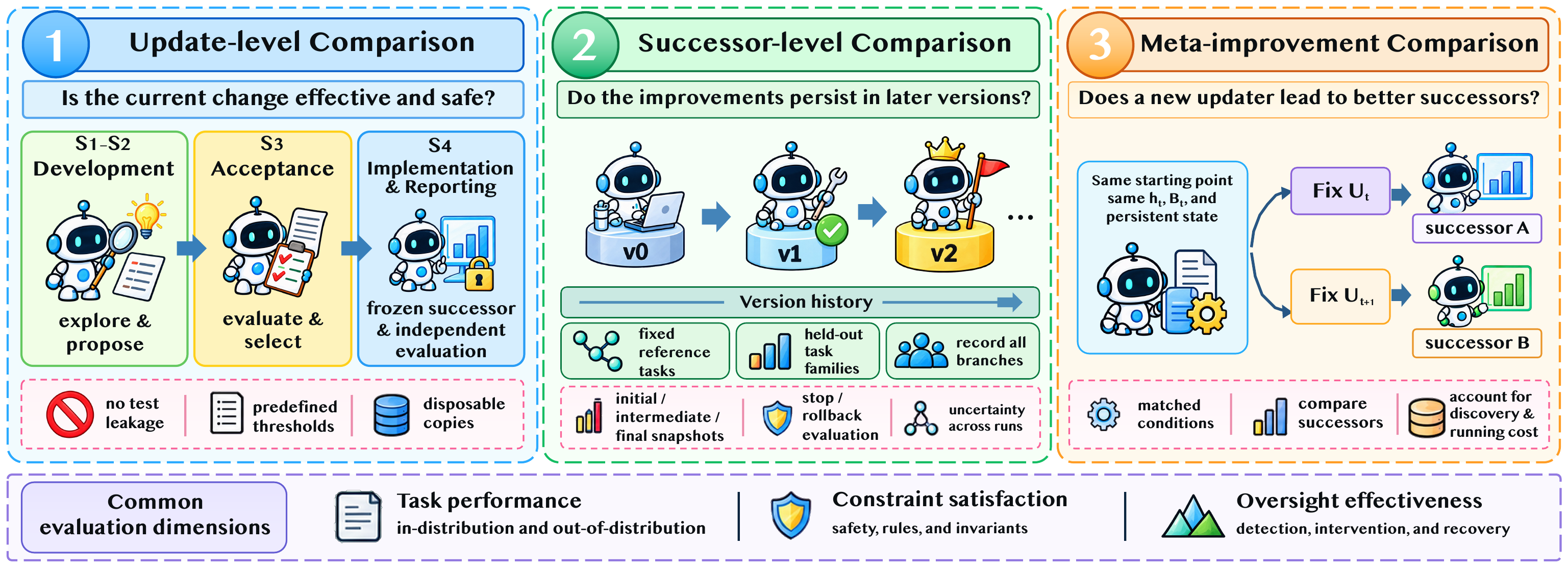}
\caption{
\textbf{Three-level comparison design for longitudinal RSI evaluation.}
Update-level comparisons separate development, acceptance, and independent
reporting after S4; successor-level comparisons track inherited outcomes
across versions; and meta-improvement comparisons use paired continuations
with fixed updaters. All three assess task performance, constraint
satisfaction, and oversight effectiveness.
}
\label{fig:three-level-comparison}
\end{figure*}

The three claim levels require different comparisons: the effect of an
individual revision, the persistence of accepted changes in later
successors, and the outcomes produced by a revised improvement mechanism.
Figure~\ref{fig:three-level-comparison} illustrates the corresponding
evaluation designs. Each addresses task performance, constraint
satisfaction, and oversight effectiveness.

\paragraph{Update-level comparisons.}
Keep development experience from S1--S2, acceptance evidence at S3, and
independent reporting on frozen successor snapshots after S4 separate.
Acceptance outcomes that inform later proposals contribute to the feedback
$F_{t+1}$ in Eq.~\eqref{eq:rsi-self-improvement}.
This distinction matters in self-authored verification: internal tests
may remain favorable as sealed deployment performance deteriorates,
whereas external acceptance reduced regressions in the studied
policy-and-test rewriting setting~\citep{guo2026selfauthored}.
Independent reporting tests should run on disposable successor copies,
with results withheld from the improvement search throughout the evaluated
run. When acceptance sets are reused, specify the feedback available to
the proposer (Section~\ref{sec:eval-evidence}). Set acceptance thresholds
in advance, including the relevant populations and adversary access where
applicable.

Comparisons also require full accounting of resources spent on experience
collection, proposals, rejected trials, training, and acceptance tests.
Report human assistance and reporting costs separately. RE-Bench and
Agent Plasticity motivate reporting performance--budget curves alongside
endpoint scores~\citep{wijk2024rebench,singh2026agentplasticity}.
Agent Plasticity also uses held-out out-of-distribution games to assess
transfer to stronger opponents. These records relate evaluation results
and costs for each revision to its subsequent use.

\paragraph{Successor-level comparisons.}
Maintain a versioned record of inherited $(h_t,B_t,\mathcal U_t)$,
dependencies, parent versions, proposals, and acceptance decisions.
Evaluate initial, intermediate, and final snapshots on common reference
tasks and held-out task families. RSIBench-Data motivates reporting both
the best observed candidate and the selected incumbent, while TEB
motivates repeated evaluation of frozen controls~\citep{meng2026rsibench,li2026traceevolutionbench}.
EvoPathBench adds interference and rule-reversal tests. Disabling
artifact access at a checkpoint helps estimate the contribution of
inherited artifacts to evaluation performance~\citep{lin2026beyondendpointperformanc}.
Vary task order, retain failed branches, and include runs that exhaust
their budgets. Estimate uncertainty across independent improvement runs
while accounting for dependence among successors from the same run.
For each requirement, maintain a versioned record of evaluation scope,
outcome, and uncertainty, using ``not evaluated'' where appropriate.

SEABench uses adaptive discovery to identify downstream
failures~\citep{das2026seabench}. Estimating their prevalence in deployment
also requires a defined task distribution and sampling procedure.
Extend attack and control evaluations~\citep{debenedetti2024agentdojo,greenblatt2024control}
across a declared horizon. Measure time to first critical failure,
cumulative violations, the proportion of runs with any critical failure,
and recovery outcomes, including prematurely stopped runs.
Repeat stop and rollback tests after S4 and after reuse of
$\mathcal U_{t+1}$, covering memory, tools, queued actions, and
permissions. Record restored internal state separately from remaining
external effects. These histories also support comparisons following
revisions to the improvement mechanism.

\paragraph{Meta-improvement comparisons.}
\citet{chen2026auditproduct} review 114 studies, including 26 RSI-labeled
studies that claim to improve an improvement procedure. Four of these
26 compare earlier and later improvers using identical starting artifacts
and matched budgets.
To attribute gains to a revised updater, compare paired continuations
from the same $h_t$, $B_t$, and persistent state, following the
fixed-updater comparisons in Hyperagents~\citep{zhang2026hyperagents}.
Use $\mathcal U_t$ in one arm and $\mathcal U_{t+1}$ in the other.
Hold each updater fixed and match task streams, permissions, and continuation
budgets. Evaluate the resulting successors on the same reporting tasks.
When the revised updater depends on changes to memory or interfaces,
treat the complete set of changes as the intervention.
A separate RI condition can evaluate further updater revision.
Versioned successor records then allow gains to be examined alongside
constraint satisfaction and intervention outcomes.

Assess total-cost efficiency separately by comparing performance at
matched total budgets, including updater discovery, validation, and
subsequent use. Charge the full one-time discovery and validation cost
to a single use. For repeated use of the same updater, report the number
of continuations and divide the one-time cost among them, adding the cost
of each continuation. Include additional adaptation or revision costs
and apply the same accounting rules to both arms.

These comparisons specify the observations needed for longitudinal
evaluation. The next section examines evidence provenance, acceptance
validity, and the interpretation of requirements across versions.


\subsection{What Supports Trustworthiness Claims?}
\label{sec:eval-evidence}
\label{sec:eval-joint}

The preceding comparisons provide evidence about system changes and
their consequences. Assessing a trustworthiness claim also requires
establishing the provenance of evidence, the validity of acceptance
procedures, and the joint satisfaction of relevant requirements.
These conditions apply across the three claim levels and all four
trustworthiness themes.

\paragraph{Evidence provenance.}
Record the data, prompts, code, and modification authority of each
evaluator. Observed contamination and unauthorized checkpoint
substitution in PostTrainBench show the importance of tracking
artifact provenance~\citep{rank2026posttrainbench}.
When evaluators change, retest common snapshots under both versions.
These records identify the evidence available to acceptance procedures
and the changes that require reassessment.

\paragraph{Acceptance validity.}
PACE uses anytime-valid tests for individual commit decisions, while
SGM combines conditionally valid confirmation with a predictable,
summable error allocation across attempts~\citep{shawn2026pace,wu2026sgm}.
REUSE addresses dependence that arises when a fixed evaluation set
influences later proposals. It provides the search procedure with
promotion decisions only and allocates error across possible decision
histories. Given an initially independent evaluation set and a
decision-only interface, its guarantees cover accepted population
improvements throughout the process~\citep{sun2026reuse}.
Evaluation records should specify sampling assumptions, access to
evaluation feedback, and the error scope of each acceptance rule.
They should also report false acceptances, missed improvements, retained
gains, and test costs. Recheck these conditions when changes to
$\mathcal U$ alter evidence access or candidate selection.
For each system version, report acceptance outcomes alongside the
other assessed requirements.

\paragraph{Joint evidence in the reviewed studies.}
At the update level, AI Control evaluates utility and resistance to
deliberate subversion under auditing and deferral protocols, connecting
task utility, security, and intervention at the protocol level.
Utility is measured with GPT-4, whereas safety is assessed using red-team
attack policies, so the results cover different evaluation conditions
rather than a single realized run~\citep{greenblatt2024control}.
AgentDojo evaluates benign utility and attack outcomes, while HELM
measures several properties of fixed models~\citep{debenedetti2024agentdojo,liang2022holistic}.
These studies provide components for joint assessment across the three analytical roles.

At the successor level, SEABench compares evolving and non-evolving
agents over task sequences, linking useful adaptation to downstream
safety and privacy failures. Its monitoring experiment evaluates
detection and blocking retrospectively on a held-out split of collected
reasoning traces; continued evolution with the monitor active remains
untested~\citep{das2026seabench}.
Misevolution examines failures across changes to models, memory, tools,
and workflows and tests corrective interventions~\citep{shao2026misevolve}.
Its task and mitigation results cover different configurations and
experiments. Together, these studies address multiple concerns;
assessing their joint satisfaction after repair requires longitudinal
evidence from the same successor trajectory.

At the meta-improvement level, DGM, Dream-RSI, Hyperagents, Evo$^2$,
MASS, and MetaRSI provide evidence of capability outcomes after updater
reuse~\citep{zhang2025dgm,zheng2026dreamrsi,zhang2026hyperagents,xu2026evo2,lee2026recursiveselfimprovement,tan2026metarsirsi2}.
We classify revisions by the role of each component in generating or
judging later changes and record retention, reuse, and evaluated successors
separately. MetaRSI adds comparisons of improvers with matched initial
states and budgets, together with retained-capability measurements;
its protection mechanisms are specified structurally.
Across these six study versions, capability outcomes receive more direct
support than longitudinal assessments jointly covering behavioral
constraints and effective intervention.
The \href{https://github.com/pujiayue/Trustworthy-RSI/tree/854de7b3fa6c018a0c1e3e83df8e08dd801369d1/audit}{versioned audit records}
document component boundaries, comparison controls, and unresolved evidence.

\paragraph{Joint assessment.}
Table~\ref{tab:eval-scorecard} lists the longitudinal evidence to report
for the eight requirement groups in Section~\ref{sec:trustworthiness-concepts}.
Report results for individual requirements separately from their joint
satisfaction along the same trajectory. A trajectory meets the joint
criterion only if every mandatory requirement holds at each scheduled
assessment throughout the declared horizon.
For example, if the protocol requires task performance above a threshold
and prohibits unauthorized disclosure, a disclosure at an intermediate
version makes the trajectory fail even if the initial and final versions
pass both checks. Report the proportion of independent runs meeting
the joint criterion and its uncertainty, marking incomplete joint
assessments as ``not evaluated''.
When pass decisions depend on estimated quantities, control uncertainty
simultaneously across requirements and evaluated versions. Report each
mandatory constraint alongside aggregate utility.
Priorities for further evaluation include fairness across versions,
privacy preservation after correction or deletion, and interactions of
these requirements with continued improvement.


\begin{table}[!htbp]
\centering
\footnotesize
\setlength{\tabcolsep}{3.5pt}
\renewcommand{\arraystretch}{1.16}
\caption{\textbf{Eight longitudinal trustworthiness checks.}
For each claim level, report versions, scope, threat model, uncertainty, and evaluation status.}
\label{tab:eval-scorecard}
\begin{tabularx}{\linewidth}{@{}
  >{\raggedright\arraybackslash}p{.33\linewidth}
  >{\raggedright\arraybackslash}p{.285\linewidth}
  >{\raggedright\arraybackslash}X@{}}
\toprule
\textbf{Requirement group} & \textbf{Core signal} & \textbf{Across versions} \\
\midrule
\rowcolor{rsiTableBand}
\textbf{Improvement validity} &
\textbf{Independent gains}; variance &
\textbf{Retention / shift}; report accuracy \\
\addlinespace[2pt]
\rowcolor{rsiTableBand}
\textbf{Behavioral safety and alignment} &
\textbf{Violation severity} &
Conflicting or \textbf{changed goals} \\
\textbf{Fairness} &
\textbf{Worst group}; subgroup gains / harms &
\textbf{Population coverage} across versions \\
\rowcolor{rsiTableStripe}
\textbf{Security and resilience} &
\textbf{Attack success}; benign utility &
\textbf{Inherited compromise}; recovery \\
\textbf{Privacy and data governance} &
\textbf{Disclosure / inference / extraction} &
\textbf{Residual exposure} after deletion \\
\addlinespace[2pt]
\rowcolor{rsiTableBand}
\textbf{Transparency and interpretability} &
\textbf{Reconstructable changes} &
\textbf{Explanation fidelity} after inheritance \\
\textbf{Human control and corrigibility} &
\textbf{Stop / rollback success}; latency &
\textbf{Residual external effects} \\
\rowcolor{rsiTableStripe}
\textbf{Accountability and governance} &
\textbf{Approval authority}; review coverage &
\textbf{Incident reconstruction} \\
\bottomrule
\end{tabularx}
\end{table}


\begin{rsitakeaway}
\begin{itemize}[leftmargin=*,nosep]
\item \textbf{Risks.} Endpoint-only gains can conceal measurement noise, loss of previously acquired capabilities, unsafe behavior inherited by later versions, and overfitting to evaluation feedback.
\item \textbf{Safeguards.} Use unchanged control systems, separate improvement feedback from independent reporting, evaluate updates after reuse, and compare successors under matched resource budgets. These measures help relate reported gains to specific changes.
\item \textbf{Future work.} Compare improvement procedures reused by later system versions under matched conditions. Test whether behavioral and safety constraints remain satisfied and authorized interventions remain effective across successive versions.
\end{itemize}
\end{rsitakeaway}

\section{Applications, Open Problems, and Conclusions}
\label{sec:outlook}

In Eqs.~\eqref{eq:rsi-model-update}--\eqref{eq:rsi-self-improvement}, experience $E_{t+1}$ informs updates to the policy $h_t$, while
feedback $F_{t+1}$ informs changes to the inductive bias $B_t$ and the bias update
function $\mathcal{U}_t$. Application settings determine how these changes are
assessed and what evidence is needed when the revised $\mathcal{U}_{t+1}$ is reused.
This section examines these requirements and the open problems they raise.

\subsection{Current Findings and Evidence Requirements}
\label{sec:outlook-findings}

Studies of self-evolving systems have identified failures in capability
retention, safety, evaluation, and inherited state. We summarize these
findings and the evidence needed to assess individual updates, successor
systems, and revised improvement mechanisms.

\subsubsection{Established Failure Modes}

\noindent Capability forgetting.
Adaptation can improve performance on newly targeted tasks while degrading
previously acquired capabilities, making retention a separate outcome to
measure \citep{yu2026forget,yuan2026seva}.

\noindent Safety degradation.
Changes to models, memory, tools, and workflows can weaken previously
observed safety behavior. Locally useful adaptations may produce downstream
safety failures, requiring continued assessment of refusal behavior and
behavioral constraints \citep{shao2026misevolve,das2026seabench}.

\noindent Evaluation failure.
Self-authored tests can favor a revised system even when its performance
under protected deployment evaluation deteriorates. Such tests may accept
changes that fail the intended requirements, separating apparent progress
from independently measured improvement \citep{guo2026selfauthored}.

\noindent Contamination inheritance.
Poisoned benchmarks can induce unsafe persistent instructions that affect
descendant coding agents even after further evolution on clean tasks.
Memory poisoning provides another route through which adversarial input
can influence later behavior
\citep{roesner2026trustingtrust,dong2025minja}.

These observations establish failures in the tested settings. Open questions
concern their prevalence, their persistence or attenuation over time, and
how revisions to $\mathcal{U}_t$ affect them. Persistence under a fixed updater
and failure after updater revision require separate comparisons.

\subsubsection{Three Levels of Trustworthiness Claims}

Section~\ref{sec:rsi-trustworthiness} defines the three claim levels, and Section~\ref{sec:eval-longitudinal} sets out their comparison designs.
Across these levels, the reviewed findings point to a shared dependence on retained state and the evidence used to select later changes.
Contamination can persist through further evolution, while self-authored tests can favor changes whose performance deteriorates under independent evaluation \citep{roesner2026trustingtrust,guo2026selfauthored}.
Capability gains, constraint preservation, and effective oversight therefore need to be examined within the same sequence of successors.

This synthesis also connects recovery with subsequent improvement.
Corrections need to remain effective as repaired state is reused and $\mathcal{U}_t$ determines which changes are retained next.
The central question is how interactions among inherited artifacts, acceptance rules, and intervention mechanisms affect the joint persistence of gains and protections.

\subsubsection{Evidence across Tasks, Versions, and Use}

The evidence required depends on how the assessed system and its use change:
\begin{itemize}
  \setlength{\itemsep}{2pt}
  \item Across tasks. Transfer should be evaluated on the new tasks,
  populations, and operating conditions covered by the claim.
  \item Across versions. A property verified for one version
  must be reassessed when later versions inherit, combine, or transform
  its changes.
  \item From a candidate to its updater. Assessing the safety of an
  improvement mechanism requires evaluating the later candidates it
  generates and accepts.
  \item From benchmarks to deployment. Intended-use validation should
  cover the users, environments, and consequences of deployment
  \citep{nist2023airmf,hleg2019ethics}.
  \item From rollback to recovery. Recovery checks should cover
  contaminated memories, tools, derived artifacts, external effects,
  and affected successors, alongside checkpoint restoration
  \citep{roesner2026trustingtrust,ying2026skilljackpersistentskillbackdoors}.
\end{itemize}
A trustworthiness claim should identify the update, sequence of successors,
or updater it covers, together with its task scope, assessment horizon,
and untested conditions.

\subsection{How Application Context Changes the Evidence Required}
\label{sec:outlook-apps}

\begin{figure}[t]
   \centering
   \includegraphics[width=\linewidth]{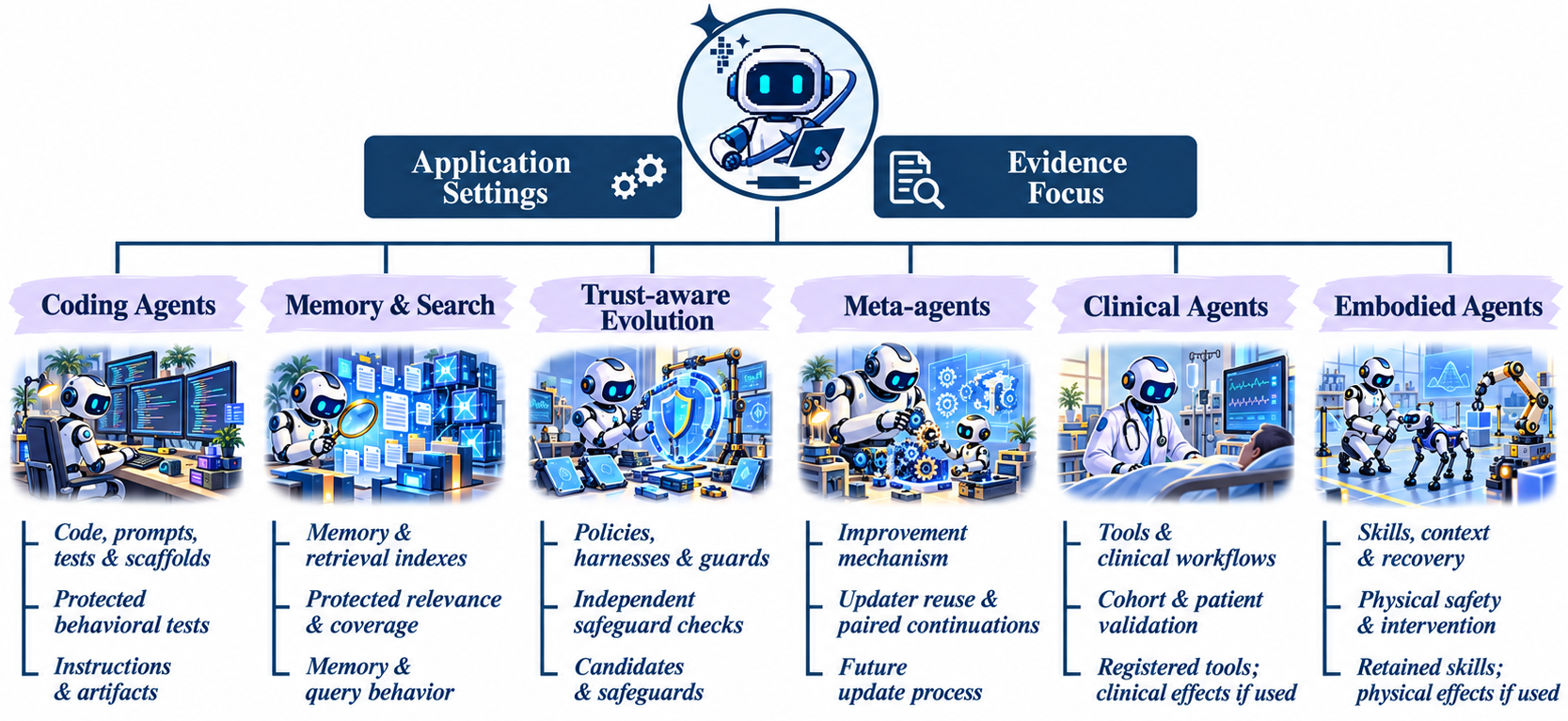}
   \caption{\textbf{Representative application settings for trustworthy RSI.}}
   \label{fig:chap8_figure}
\end{figure}

Application context determines which evidence requirements are most relevant.
For the settings illustrated in Figure~\ref{fig:chap8_figure}, three
questions guide evaluation: what can the system modify, who controls the
feedback used at S3, and what persists or happens after implementation at S4?
Table~\ref{tab:outlook-gaps} compares representative settings along these
dimensions and identifies their evidence gaps.

\subsubsection{Scope of Modification: What Can the System Change?}

\noindent Persistent artifacts.
Changes to memories, retrieval indexes, and tool configurations alter the
resources available to later tasks. ReasoningBank retains reusable
experience, MemEvolve evolves memory-system designs, and
Self-Evolving Search Index changes retrieval representations
\citep{ouyang2026reasoningbank,zhang2025memevolve,lee2026selfevolvingsearchindex}.
Evidence must therefore address how retained artifacts affect later
queries and actions, including coverage, contamination, and deletion,
beyond immediate relevance gains.

\noindent System behavior and implementation.
Policies, prompts, code, and agent scaffolds can change how the system
acts. Darwin G\"odel Machine evolves agent implementations, while
SafeEvolve revises harness components in response to safety
incidents and trains the policy to use them
\citep{zhang2025dgm,mao2026safeevolveharnesspolicycoevolutionagent}.
AgentBreeder searches scaffolds using capability and safety
objectives, and SafeCoEvo adapts safety harnesses and guards
\citep{rosser2025agentbreeder,cheng2026safecoevocoevolvingsafetyharnesses}.
These edits require behavioral regression tests and checks on the
interactions between revised components and retained safeguards.

\noindent The improvement mechanism.
Edits to the procedures that generate, evaluate, or accept future changes
can revise $\mathcal{U}_t$ itself. Hyperagents permits changes to the
meta-agent that produces further edits; Dream-RSI reuses refined
exploration procedures, and MetaRSI allows scheduling and proposal
policies over improvement operators to change
\citep{zhang2026hyperagents,zheng2026dreamrsi,tan2026metarsirsi2}.
The relevant evidence concerns the subsequent updates produced when the
revised mechanism is reused. In DGM, agent implementation edits operate
within an outer exploration procedure whose fixed components remain part
of the evaluation conditions \citep{zhang2025dgm}.

These categories describe functional roles and may overlap. A prompt or
memory rule belongs to the updater when it governs future improvement;
its role determines whether an edit changes $h_t$, $B_t$, or $\mathcal{U}_t$.
A contained tool-configuration change may be assessed through targeted
tests of its behavior and dependencies. An updater revision can affect
many later acceptance decisions, so reassessment should cover the
components and decisions that depend on it.

\subsubsection{Independence of Feedback: Who Controls the Test?}

\noindent Self-authored evaluation.
When the evolving system writes its own tests, the evidence used to
accept a candidate can share its assumptions and errors. Self-authored
verification has produced favorable internal results alongside worsening
protected deployment performance \citep{guo2026selfauthored}.
Acceptance therefore requires independent checks that the tests measure
the intended behavior.

\noindent Co-evolving evaluators.
If the evaluator changes with the system, a higher score can reflect a
different standard rather than a better candidate. Evaluator co-evolution
and grader drift make comparisons across versions sensitive to which
judge is used \citep{iacob2026redqueen,zhang2026grader}.
Retesting common frozen snapshots under both evaluator versions can
restore comparability, but establishing validity still requires a
reference independent of the coupled improvement process.

\noindent External or protected evaluation.
Frozen evaluators and held-out reporting tasks can separate assessment
from development; Athanor uses such services in bounded autonomous
post-training \citep{zhou2026athanor}.
This separation must cover test data, grading rules, and access to
outcomes. Acceptance feedback that guides later proposals becomes part
of the improvement experience; final reporting then requires separately
protected evidence. Benchmark poisoning also makes the integrity of
external evaluation sources a necessary check
\citep{roesner2026trustingtrust}.

Statistical acceptance rules address a complementary issue.
SGM and REUSE study the reliability of repeated acceptance
decisions under specified testing assumptions
\citep{wu2026sgm,sun2026reuse}.
Applying these controls to trustworthiness claims requires relevant,
uncontaminated tests and satisfaction of the stated sampling and feedback
assumptions. Those assumptions must be reassessed
when $\mathcal{U}_t$ changes evidence selection or interpretation.
Independent validation should cover the candidate's and updater's access
to S3, as well as the intended deployment conditions.

\subsubsection{Effects of Update Implementation: What Happens after Acceptance?}

\noindent Reversible software changes.
In an isolated, versioned setting, an accepted code, prompt, or test
configuration can often be restored to an earlier state. Acceptance can
then rely on regression tests, dependency checks, and a demonstrated
restoration procedure appropriate to that scope. This reversibility
depends on containing external effects: restoring executable code does
not reverse actions already taken or automatically restore other state.

\noindent Persistent state and derived artifacts.
Memories, retrieval indexes, learned skills, and external files can retain
an update's influence after the originating code or input is removed.
Skill backdoors have persisted after source removal, and PASB
shows that saved user claims can bias responses in fresh sessions
\citep{ying2026skilljackpersistentskillbackdoors,mao2026agentsremember}.
These findings motivate checks on provenance, the scope of inherited
claims, and the effects of correction or deletion across derived state.
Privacy assessment should additionally examine the permitted use,
retention, and disclosure of real user data.

\noindent Consequential real-world actions.
Clinical and embodied applications require evidence tied to the people
and environments affected by an accepted update. MedRSI
prioritizes failures by clinical cost and separates rapid tool discovery
from conservative registration across patient cohorts
\citep{wu2026medrsirecursiveselfimprovementmedical}.
Evaluating these mechanisms for clinical use requires independent
cohort and patient-level validation, including deployment safety. RoboFoundry retains context, skills, and recovery
procedures and studies transfer across robots
\citep{liang2026robofoundrysystemaspolicyevolutionselflearning}.
Its successors require physically grounded tests of safety and
intervention under the intended operating conditions.
VeriFine co-adapts policies, curricula, and a verification judge,
using selected human feedback to recalibrate that judge
\citep{zhou2026verifine}.
Such calibration supports verification within the tested setting.
Deployment approval and continued control after revision require
independent assessment.

Less reversible changes and wider impacts call for stronger external
evidence, explicit authorization, and effective intervention before
consequential use
\citep{nist2023airmf,hleg2019ethics}.
The scope of modification, feedback independence, and effects of
implementation determine which earlier findings remain applicable and
which claims require reassessment.

\begin{table}[t]
\centering
\caption{Representative system settings and their evidence requirements.
The feedback column summarizes evaluation arrangements and required
independent checks; the final column lists the evidence needed for
broader or longer-term claims.}
\label{tab:outlook-gaps}
\footnotesize
\setlength{\tabcolsep}{3pt}
\renewcommand{\arraystretch}{1.17}
\begin{tabularx}{\linewidth}{@{}>{\raggedright\arraybackslash}p{0.15\linewidth}
>{\raggedright\arraybackslash}X
>{\raggedright\arraybackslash}X
>{\raggedright\arraybackslash}X
>{\raggedright\arraybackslash}X@{}}
\toprule
\textbf{System setting} &
\textbf{What is modified?} &
\textbf{Is feedback independent?} &
\textbf{What persists after acceptance?} &
\textbf{Main evidence gap} \\
\midrule
Coding agents
\citep{zhang2025dgm,guo2026selfauthored} &
Agent code, prompts, tests, and scaffolds. &
Independence depends on test ownership; protected assessment is needed. &
Code, instructions, and generated artifacts. &
Safety under the intended deployment conditions. \\
\addlinespace[3pt]
Memory / search agents
\citep{zhang2025memevolve,lee2026selfevolvingsearchindex} &
Memory contents and designs, retrieval policies, and indexes. &
Independence requires protected relevance and coverage checks beyond search feedback. &
Memories, indexes, and query behavior. &
Privacy, coverage, and resistance to poisoning. \\
\addlinespace[3pt]
Agents with evolving safeguards
\citep{rosser2025agentbreeder,cheng2026safecoevocoevolvingsafetyharnesses} &
Candidates, policies, harnesses, and guards. &
Reassess feedback independence after safeguards change. &
Accepted candidates and safety mechanisms. &
Protection retained across later versions. \\
\addlinespace[3pt]
Meta-agents (Hyperagents)
\citep{zhang2026hyperagents} &
The mechanism $\mathcal{U}_t$ governing later updates. &
Assessment must remain independent of the editable updater. &
The revised process for producing future changes. &
Quality of subsequent updates under the revised updater. \\
\addlinespace[3pt]
Clinical agents (MedRSI)
\citep{wu2026medrsirecursiveselfimprovementmedical} &
Tools, registration decisions, and workflows. &
Cohort-based registration; independent clinical validation is still required. &
Registered tools and workflows; clinical effects if deployed. &
Patient-level validity and clinical deployment safety. \\
\addlinespace[3pt]
Embodied agents (RoboFoundry)
\citep{liang2026robofoundrysystemaspolicyevolutionselflearning} &
Skills, context, and recovery procedures. &
Tested transfer; independent physical safety and intervention checks are required. &
Retained skills and procedures; physical effects when executed. &
Physical safety beyond the tasks and real-world trials tested. \\
\bottomrule
\end{tabularx}
\end{table}

\subsection{Open Problems and Future Directions}
\label{sec:outlook-open}

The findings in Section~\ref{sec:outlook-findings} and the application gaps in Table~\ref{tab:outlook-gaps} motivate seven research directions.
Each concerns a condition under which gains, constraints, or oversight may change as accepted updates and improvement mechanisms are reused.

\subsubsection{Theoretical Foundations of Trustworthiness under Recursive Change}

The inherited failures summarized in Section~\ref{sec:outlook-findings}
raise a question about composition: under what conditions do requirements satisfied by individual updates survive when those updates are combined and the updater itself is revised?
A formal account should distinguish requirements that must remain invariant across revisions from those that may adapt to changing environments or authorized human preferences.
Useful results would specify permissible changes to $\mathcal{U}_t$ and the assumptions needed for guarantees to hold over successive updates.
Counterexamples in which changes are safe in isolation but conflict through shared memory, permissions, or evaluation state would expose missing composition assumptions.
These cases could guide sufficient conditions for preserving constraints and oversight across revisions.

\subsubsection{Verifying Self-Modifying Improvement Mechanisms}

When agents modify their evaluators or acceptance criteria, reported gains may reflect changed tests rather than improved behavior~\citep{guo2026selfauthored}.
The research problem is how independent verification should respond when $\mathcal{U}_t$ changes the evidence used to accept later updates.

For evaluator revisions, assessing a common set of frozen candidates with both evaluator versions and a protected reference would reveal which acceptance decisions change and whether those changes agree with independent outcomes.
For changes to the broader improvement procedure, the paired continuations in Section~\ref{sec:eval-longitudinal} can test whether $\mathcal{U}_{t+1}$ produces better subsequent updates and retains their benefits.
Verification contracts should identify the evidence access, permissions, and component dependencies on which these assessments rely, so that later edits trigger checks of the assumptions they affect.

\subsubsection{Long-Horizon Risk Dynamics and Adaptive Mitigation}

Existing studies report safety degradation and inherited security vulnerabilities across revisions of self-evolving agents
\citep{shao2026misevolve,roesner2026trustingtrust}.
The unresolved issue is how reuse and interactions among retained changes affect the duration and recurrence of these failures.
Runs starting from comparable clean and contaminated states could vary how often affected artifacts are reused, tracking when unsafe behavior appears, persists, or returns on later clean tasks.

Persistent skill backdoors show that removing source records can leave harmful derived skills intact
\citep{ying2026skilljackpersistentskillbackdoors}.
Recovery studies could compare source removal,
checkpoint restoration, and cleanup that also covers affected skills, memories, external artifacts, and successors.
Residual violations, retained benign capabilities, and recurrence after further updates would show which dependencies sustain contamination.
Repeating these comparisons with an unchanged updater and a revised updater would help separate propagation through inherited state from changes in how contaminated experience is selected and reused.

\subsubsection{Recursive Improvement of Trustworthiness Mechanisms}

SafeEvolve and SafeCoEvo study adaptation of policies, harnesses, and safeguards
\citep{mao2026safeevolveharnesspolicycoevolutionagent,
cheng2026safecoevocoevolvingsafetyharnesses}.
These improvements may reside in model parameters, external agent harnesses, or both.
A further question is how protections change when new and previously encountered failure types alternate over longer update sequences, especially when the procedure producing safeguards is also revised.
Starting from the same policy and safeguards, comparisons could keep safeguards fixed, adapt them with an unchanged updater, or adapt them with a revised $\mathcal{U}_{t+1}$ under matched exposure and budgets.
Independent checks of violations, unnecessary refusals, and correction success across later versions would measure retention and the contribution of updater revision.
For hybrid designs, these checks should also examine whether parameter updates weaken protections enforced by the external harness.

\subsubsection{Adaptive Control during Execution}

The persistent state and external effects discussed in
Section~\ref{sec:outlook-apps} make the scope of intervention a concrete research question: after updater revision changes execution, delegation, or state reuse, which actions and derived states remain within the reach of authorized stop, correction, and rollback commands?

Tests could issue these commands at comparable execution stages before and after revision, including runs with active delegated tasks, queued tool calls, and retained artifacts.
Outcomes should record intervention latency, which actions stop, which memories, skills, and external files are restored or corrected, and whether affected successors reproduce the failure when reused.
For irreversible actions, the relevant outcomes are prevention before execution and the effectiveness of subsequent remediation.
Comparing fixed and adaptive allocations of monitoring and verification resources under equal computation and human-review budgets would test whether adaptation extends intervention coverage as modifications
become more frequent.

\subsubsection{Fundamental Trade-offs and Limits of Trustworthy RSI}

The acceptance and recovery requirements in Section~\ref{sec:outlook-apps} raise two concrete trade-offs.
Stricter acceptance tests may reduce false acceptance while missing valid improvements.
Applying different thresholds and verification budgets to the same candidate set, with candidate quality assessed through protected evaluation, would measure both errors and their computational cost.
Following accepted candidates through later updates would show how these choices affect retained gains.

Tracing inherited failures requires records of which artifacts were created, reused, and modified.
Logging studies could vary event coverage and retained content separately, producing full records, redacted records, and selectively retained records from the same runs.
Failure attribution, correction success, and sensitive information recoverable under specified access permissions would reveal which
logging choices retain useful provenance at lower privacy cost.
These studies could characterize Pareto trade-offs for the tested designs.
Theoretical limits would require explicit assumptions about available feedback, computation, and permitted modifications, together with bounds on the combinations of outcomes those assumptions allow.

\subsubsection{Domain-Specific and Multi-Stakeholder Trustworthiness}

The clinical and retrieval settings in Section~\ref{sec:outlook-apps} raise a question about whose outcomes determine acceptance.
In multi-stakeholder settings, improvement for one participant may disadvantage others.
Comparing selection based on aggregate performance with selection subject to agreed stakeholder constraints would test whether overall gains preserve those requirements across later versions.
Clinical evaluation could track patient-group errors on independent cohorts; retrieval evaluation could track relevance for different user groups and exposure for content providers.
Varying feedback weights while holding tasks and update budgets comparable would help identify whose preferences drive later changes.
The relevant evidence is whether agreed constraints remain satisfied as the system and its acceptance criteria evolve.

\subsection{Conclusions}
\label{sec:outlook-conclusions}
Trustworthy RSI requires evidence that improvements are genuine and sustained, applicable constraints remain satisfied, and authorized oversight remains effective as systems change. These requirements connect the consequences of an individual update to inherited behavior and to the mechanisms producing later updates. Evidence at one level supports a different claim from evidence at another: successful successors alone do not establish that a revised updater improves the process that generates them. The central evaluation challenge is to align capability, constraint, and intervention measurements across versions under independent, comparable, and reproducible conditions. Versioned evidence and controlled continuations can make these claims testable while keeping unassessed requirements explicit.


\AddToHookNext{cmd/bibsection/after}{\phantomsection\addcontentsline{toc}{section}{\refname}}
\bibliography{main}

\clearpage
\appendix

\section{Literature Coverage and Citation Scope}
\label{app:citation-scope}

\subsection{Subject Coverage and Evidence Boundaries}
\label{app:subject-coverage}

This survey examines trustworthy recursive self-improvement in large-model
systems. Its scope includes persistent changes to models, prompts, memories,
tools, skills, and workflows, as well as revisions to the mechanisms that
produce, evaluate, and implement later changes. The selected literature
supports the foundations in Section~\ref{sec:foundations}, the four
trustworthiness themes in Sections~\ref{sec:validity}--\ref{sec:governance},
and the evaluation and research directions in
Sections~\ref{sec:evaluation}--\ref{sec:outlook}.

\paragraph{Search and selection.}
The literature search covers work available through 26 October 2026.
We used arXiv, conference and journal records, and relevant references in
RSI, self-evolution, and trustworthy AI papers. Search terms combined
\emph{recursive self-improvement}, \emph{self-evolving agents},
\emph{harness optimization}, and \emph{persistent memory} with
\emph{safety}, \emph{alignment}, \emph{robustness}, \emph{privacy},
\emph{oversight}, \emph{evaluation}, and \emph{continual learning}.
Titles, abstracts, methods, experiments, and related-work discussions
were examined to identify relevant studies and trace connected research.

We included studies that specify an improvement mechanism, measure a
trustworthiness outcome, test a safeguard, or provide an evaluation method
relevant to the review. Foundational work, standards, related surveys,
and system descriptions provide conceptual and technical context.
Studies without a substantive connection to these questions and duplicate
versions of the same work were excluded from separate counting.
Empirical findings are reported with their tested settings and comparisons;
formal findings are presented with their stated assumptions.

\paragraph{Evidential roles.}
Direct RI evidence identifies a retained revision to a functional component
of $\mathcal U$ and its reuse in generating or evaluating later updates.
General self-evolution evidence follows persistent changes to $h$ or $B$,
including memory, skills, and harness configurations, under the studied
improvement procedure. Related mechanism and evaluation evidence examines
specific risks, controls, or measurement methods, such as fine-tuning
regressions, memory authorization, intervention tests, and independent
assessment. Classification follows the function of the modified component
and its observed reuse.

\subsection{Counting Unit and Source Normalization}
\label{app:counting-scope}

The survey cites \textbf{301 distinct works}, represented by
\textbf{301 citation keys}. The count includes citations in the main text,
tables, captions, and active literature figures. Commented-out references
and uncited bibliography entries are excluded.

Preprint and archival versions are identified using arXiv identifiers,
DOIs, normalized titles, and author lists, with archival publication
records taking precedence when available. For example,
\emph{Unveiling Privacy Risks in LLM Agent Memory} is counted once under
ACL. A cited literature audit counts as one work; studies it analyzes are
counted individually only when independently cited in this survey.
The citation-to-source assignments are listed in
\nolinkurl{audit/citation_source_audit.csv}.

Conference editions are grouped under their parent venue, with ACL
Findings included under ACL. Workshop papers are recorded separately
from main-conference proceedings. Named journal publications retain their
journal category, including those with arXiv versions. Preprints without
a named archival venue retain their recorded source. Web publications,
institutional reports, and books have separate categories. The cited
publications span 1973--2026.

\clearpage
\subsection{Distribution by Recorded Publication Source}
\label{app:source-distribution}

Table~\ref{tab:citation-source-distribution} reports the publication-source
distribution of all 301 cited works. Percentages are rounded independently
to two decimal places.

\begin{table}[htbp]
\centering
\caption{Publication sources of the 301 references cited in this survey.}
\label{tab:citation-source-distribution}
\begingroup
\small
\setlength{\tabcolsep}{5pt}
\renewcommand{\arraystretch}{1.08}
\begin{tabularx}{\textwidth}{@{}>{\raggedright\arraybackslash}p{2.4cm}>{\raggedright\arraybackslash}X r@{}}
\toprule
\textbf{Source} & \textbf{Full venue or source name} & \textbf{References} \\
\midrule
\multicolumn{3}{@{}l}{\emph{Preprints and online sources}} \\
arXiv & arXiv preprints & 198 (65.78\%) \\
Web & Online model announcements and research essays & 3 (1.00\%) \\
OpenReview & OpenReview records without a named proceedings venue & 1 (0.33\%) \\
Preprints.org & Preprints.org preprints & 1 (0.33\%) \\
\textbf{Subtotal} & & \textbf{203 (67.44\%)} \\
\midrule
\multicolumn{3}{@{}l}{\emph{Conference and workshop papers}} \\
ICLR & International Conference on Learning Representations & 25 (8.31\%) \\
ICML & International Conference on Machine Learning & 25 (8.31\%) \\
NeurIPS & Conference on Neural Information Processing Systems & 20 (6.64\%) \\
ACL & Annual Meeting of the Association for Computational Linguistics (including Findings) & 5 (1.66\%) \\
KDD & ACM SIGKDD Conference on Knowledge Discovery and Data Mining & 2 (0.66\%) \\
SIGIR & ACM SIGIR Conference on Research and Development in Information Retrieval & 1 (0.33\%) \\
AAAI Wksh. & Workshops at the AAAI Conference on Artificial Intelligence & 1 (0.33\%) \\
NeurIPS Wksh. & NeurIPS Workshop on Self-Evolving Diversity-Driven Search for Robust AI Systems (EvoRobust) & 1 (0.33\%) \\
COLM & Conference on Language Modeling & 1 (0.33\%) \\
UAI & Conference on Uncertainty in Artificial Intelligence & 1 (0.33\%) \\
\textbf{Subtotal} & & \textbf{82 (27.24\%)} \\
\midrule
\multicolumn{3}{@{}l}{\emph{Journal papers}} \\
ACM CSUR & ACM Computing Surveys & 1 (0.33\%) \\
Acad. AI & Academia AI and Applications & 1 (0.33\%) \\
FCS & Frontiers of Computer Science & 1 (0.33\%) \\
FnT TCS & Foundations and Trends in Theoretical Computer Science & 1 (0.33\%) \\
F. Robot. AI & Frontiers in Robotics and AI & 1 (0.33\%) \\
Nature & Nature & 1 (0.33\%) \\
Science & Science & 1 (0.33\%) \\
TMLR & Transactions on Machine Learning Research & 1 (0.33\%) \\
\textbf{Subtotal} & & \textbf{8 (2.66\%)} \\
\midrule
\multicolumn{3}{@{}l}{\emph{Books and institutional reports}} \\
Reports & Institutional technical reports and research drafts & 5 (1.66\%) \\
Books & Books & 3 (1.00\%) \\
\textbf{Subtotal} & & \textbf{8 (2.66\%)} \\
\midrule
\textbf{Total} & & \textbf{301 (100.00\%)} \\
\bottomrule
\end{tabularx}
\endgroup
\end{table}

The categories assign each cited work to one publication source.
Publication source and evidential role are reported separately.

\subsection{Evidence Selection and Coding Protocol}
\label{app:evidence-protocol}

Figure~\ref{fig:eval-literature} compares 20 representative studies
across 13 evidence dimensions, yielding 260 coded cells. The studies
represent three analytical uses: assessing individual updates, testing
inheritance and continued use in successors, and examining revisions to
the improvement mechanism. Related benchmarks and fixed-model oversight
studies provide evaluation methods. Dream-RSI and Hyperagents represent,
respectively, comparisons involving exploration-policy revision and
transfer tests of revised meta-agents.

Each cell uses one of four codes. \emph{Examined} denotes a measured
outcome or formal analysis for trustworthiness properties, or an
implemented, reported, or formally analyzed evaluation design.
\emph{Discussed or case} denotes substantive discussion, a proposed
safeguard, or an illustrative case. \emph{Not evaluated} identifies a
property outside the assessed study scope. \emph{Unclear} marks evidence
that does not resolve the classification. The task-performance row
includes benign utility; the intervention row covers tested stopping,
correction, recovery, or blocking.

Cost records resource expenditure or sample complexity, while
experimental controls records baselines, paired comparisons, or
ablations. Updater edits requires a retained change to a functional
component that governs later proposals, selection, or acceptance.
The final matrix row requires observed updates produced using the
revised updater. The coding records modification, retention, reuse,
and downstream evaluation separately. For DGM, the reused code-writing
agent changes while archive selection remains fixed. For SAV, the
coding of updater revision and reuse remains unresolved because the
available evidence does not identify whether retained tests govern
subsequent acceptance decisions.

The \href{https://github.com/pujiayue/Trustworthy-RSI/tree/854de7b3fa6c018a0c1e3e83df8e08dd801369d1/audit}{public audit records}
include \nolinkurl{audit/evidence_cells.csv}, with citation keys, paper
versions, coding criteria, source locations, and verification status
for all 260 cells. The records include 40 targeted primary-source
checks and annotated source locations for the other 220 cells.
The companion \nolinkurl{audit/joint_evidence.csv} documents functional
components, successor evidence, and joint assessment across the three
analytical roles.

Joint evidence requires matched system versions, an identifiable
improvement trajectory, and compatible tasks, threat models, and
evaluation conditions. The joint-evidence record distinguishes the
properties assessed within each study from those tested together on
the same system trajectory. Section~\ref{sec:eval-joint} specifies the
corresponding longitudinal joint-assessment criterion.

\end{document}